\documentclass{article}

    \PassOptionsToPackage{numbers, compress}{natbib}

    \usepackage[preprint]{neurips_2023}

\usepackage{multirow}
\usepackage[utf8]{inputenc} 
\usepackage[T1]{fontenc}    
\usepackage{hyperref}       
\usepackage{url}            
\usepackage{booktabs}       
\usepackage{amsfonts}       
\usepackage{nicefrac}       
\usepackage{microtype}      
\usepackage{xcolor}         
\usepackage{graphicx}
\usepackage{amsmath}
\usepackage{amssymb}
\usepackage{mathtools}
\usepackage{amsthm}
\usepackage{wrapfig}

\renewcommand{\vec}[1]{\boldsymbol{#1}}
\title{Diffusion Models Generate Images Like Painters: \\an Analytical Theory of Outline First, Details Later}

\author{%
  Binxu Wang\\
  Kempner Institute\\
  Harvard University\\
  Boston, MA\\
  \texttt{binxu\_wang@hms.harvard.edu} \\
  \And
  John J. Vastola \\
  Department of Neurobiology\\
  Harvard Medical School\\
  Boston, MA\\
  \texttt{John\_Vastola@hms.harvard.edu} \\
}

\begin{document}

\maketitle


\begin{abstract}
How do diffusion generative models convert pure noise into meaningful images? In a variety of pretrained diffusion models (including conditional latent space models like Stable Diffusion), we observe that the reverse diffusion process that underlies image generation has the following properties: (i) individual trajectories tend to be low-dimensional and resemble 2D `rotations'; (ii) high-variance scene features like layout tend to emerge earlier, while low-variance details tend to emerge later; and (iii) early perturbations tend to have a greater impact on image content than later perturbations. To understand these phenomena, we derive and study a closed-form solution to the probability flow ODE for a Gaussian distribution, which shows that the reverse diffusion state rotates towards a gradually-specified target on the image manifold. It also shows that generation involves first committing to an outline, and then to finer and finer details. We find that this solution accurately describes the initial phase of image generation for pretrained models, and can in principle be used to make image generation more efficient by skipping reverse diffusion steps. Finally, we use our solution to characterize the image manifold in Stable Diffusion. Our viewpoint reveals an unexpected similarity between generation by GANs and diffusion and provides a conceptual link between diffusion and image retrieval.
\end{abstract}

\begin{figure*}[hb]
\vskip -0.1in
\begin{center}
\includegraphics[width=0.96\columnwidth]{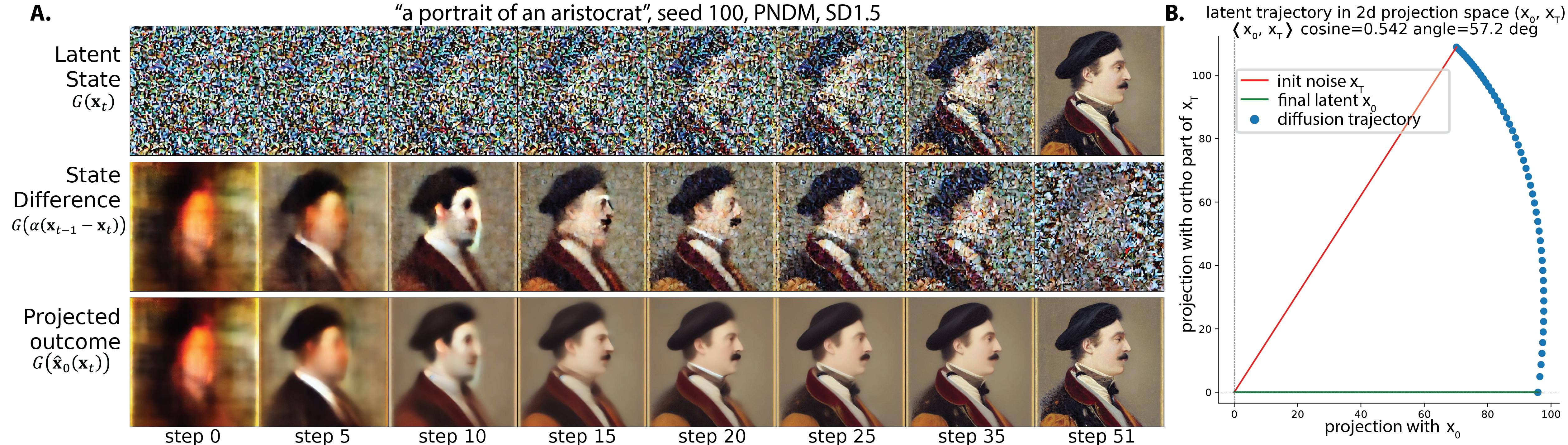}
\caption{\textbf{Characteristics of image generation by diffusion models}. \textbf{A}. Tracking latent states $G(\mathbf{x}_t)$ (top row), differences between nearby time steps $G(k (\mathbf{x}_{t-1} - \mathbf{x}_{t}))$ (middle row), and final image estimates $G(\hat{\mathbf{x}}_0(\mathbf{x}_t))$ (bottom row) suggests different measures of progress. \textbf{B}. Individual trajectories are effectively two-dimensional, with the transition from $\mathbf{x}_T$ to $\mathbf{x}_0$ being rotation-like.}
\label{fig:initial_obs}
\vspace{-15pt}
\end{center}
\end{figure*}


\section{Introduction}

Imagine an artist painting a picture of a natural landscape. We generally expect higher-level scene elements to appear first, and lower-level details later: the borders of the land and sky might be drawn, then the largest objects (like mountains and trees) might be placed, then minor objects (like rocks and small animals) might be placed, and finally fine details (like textures and shading) might be filled in. Do diffusion generative models \cite{sohl-dickstein2015noneq,song2019est,song2021scorebased}, which can generate natural landscapes like those of an artist, also construct images like this? If not, how do they work? Specifically, how do they `determine' what to generate from the noise?

By visualizing images throughout a reverse diffusion trajectory---from pure noise to the final image---the naive answer appears to be no. One gets the impression of an image emerging fully-formed from the noise; one is `uncovering' the image, or `opening one's eyes' to reveal an image that was always there. But visualizing an endpoint estimate of the reverse diffusion indicates that large-scale image features emerge before details (see e.g. Fig. 6 of \cite{ho2020DDPM} and Fig. 1 of \cite{karras2022elucidatingDesignSp}), which suggests that the naive view is misleading. Hertz et al. reach a similar conclusion from studying conditional diffusion models' cross-attention maps, finding that different parts of an image may be `attended to' at different times \cite{hertz2022crossAttnCtrl}, and that coarse features (e.g. of a bear, see their Figure 4) are attended to before details.

Our aim in this work is to explore the apparent outline-first, details-later behavior of reverse diffusion in quantitative detail, using a mix of simple theory and numerical experiments on pre-trained diffusion models. Our theory and experiments together support the following claims about diffusion model image generation: (i) individual reverse diffusion trajectories tend to be very low-dimensional; (ii) scene elements that vary more within training data tend to emerge earlier; and (iii) early perturbations substantially change image content more often than late perturbations. 

Our major contribution is to provide a closed-form solution to the sampling trajectory of probability flow ODE. This solution can qualitatively explain the generation behavior of pre-trained diffusion models and quantitatively predict their sampling trajectory in the early phase, just given knowledge of the mean and covariance of the training data. Practically, our result can be leveraged to accelerate sampling by skipping the early phase entirely, and it can also be used to characterize the image manifold embedded in the diffusion models. Finally, by deriving the sampling trajectory for the exact score of training data, we draw connections between diffusion models and image retrieval process.

Conceptually, the viewpoint we develop sheds light on the geometry of diffusion models, and in particular on the difficult-to-identify low-dimensional manifold that smoothly parameterizes generated images \cite{wenliang2022scorebased}. We identify some interesting parallels with the geometry of the analogous manifold for generative adversarial networks (GANs) \cite{goodfellow2014GANs}. 

\section{Diffusion generative modeling basics}
\label{sec:background}
There are several complementary theoretical frameworks of diffusion generative modeling \cite{sohl-dickstein2015noneq,ho2020DDPM,song2019est,yang2022review}. Guided by the unifying view of \cite{karras2022elucidatingDesignSp}, in this work, we focus on the continuous-time framework of Song et al. \cite{song2021scorebased} 
Diffusion generative models involve mapping a data distribution $p(\mathbf{x})$ to a simpler one $p(\mathbf{x}_T)$ via a stochastic process---typically pure diffusion or an Ornstein-Uhlenbeck (OU) process. This so-called `forward' process can be inverted via a `reverse' process, which is mathematically guaranteed to exist for reasonable choices of initial distribution and forward process \cite{anderson1982rdiffmath}. 
Thus, to generate new samples from $p(\mathbf{x})$, we can sample from the simpler distribution (e.g. a Gaussian) and run the reverse process.  

\paragraph{Forward/reverse diffusion.} We consider forward processes defined by the stochastic differential equation (SDE) 
\begin{equation} \label{eq:forward_special}
\dot{\vec{x}} = -\beta(t)\vec{x}+g(t)\vec{\eta}(t)
\end{equation}
where $\beta(t)$ controls the decay of signal, $g(t)$ is a time-dependent noise amplitude, $\vec{\eta}(t)$ is a vector of independent Gaussian white noise terms, and time runs from $t = 0$ to $T$. Its reverse process is
\begin{equation} \label{eq:rev_special}
    \dot{\mathbf{x}} = -\beta(t)\mathbf{x}- g(t)^2 \mathbf{s}(\mathbf{x}, t) + g(t)\mathbf{\eta}(t)
\end{equation}
where $\mathbf{s}(\mathbf{x}, t) := \nabla_{\mathbf{x}} \log p(\mathbf{x}, t)$ is the score function, and where we use the standard convention that time runs \textit{backward}, i.e. from $t = T$ to $0$. In this paper, we focus on one popular forward process: the variance-preserving SDE, which enforces the constraint $\beta(t)=\frac 12 g^2(t)$. The marginal probabilities of this process are
\begin{equation}
p(\mathbf{x}_t | \mathbf{x}_0) = \mathcal{N}(\mathbf{x}_t | \alpha_t \mathbf{x}_0, \sigma_t^2 \vec{I}) \hspace{0.3in} \alpha_t := e^{- \int_0^t \beta(t') dt'} \hspace{0.3in} \sigma_t^2 := 1 - e^{ - 2 \int_0^t \beta(t') dt' }
\end{equation}
where $\alpha_t$ and $\sigma_t$ represent the signal and noise scale, satisfying $\alpha_t^2 + \sigma_t^2 = 1$. Normally, as $t$ goes from $0\to T$, signal scale $\alpha_t$ monotonically decreases from $1\to 0$ and $\sigma$ increases from $0\to 1$. (Appendix \ref{apd:notation} relates our notation to others' notation.) 
Note that there exists a deterministic \textit{probability flow ODE} with the same marginal probabilities \cite{song2021scorebased} as the reverse SDE:
\begin{equation} \label{eq:rev_flow}
        \dot{\mathbf{x}} = -\beta(t)\mathbf{x} - \frac{1}{2}g(t)^2 \mathbf{s}(\mathbf{x},t)
\end{equation}
where time again runs backward from $t = 1$ to $t = 0$. In practice, instead of the SDE (Eq.\ref{eq:forward_special}), this deterministic process is often used to sample from the distribution \cite{karras2022elucidatingDesignSp}. The behavior of the probability flow ODE will be our main focus.  



\paragraph{Learning the score function.} The score function, which is required to reverse the forward process, can be learned via gradient descent on the denoising score-matching objective 
\begin{equation}\label{eq:denoising_score_obj}
    \mathbb E_{\mathbf{x}_0\sim p(\mathbf{x}_0),\vec{\epsilon}\sim\mathcal N(\vec{0},\vec{I})}\int_0^1 \gamma_t \|\vec{\epsilon}_\theta(\alpha_t \mathbf{x}_0+\sigma_t \vec{\epsilon},t)-\vec{\epsilon} \|_2^2 \ dt \hspace{0.4in} \vec{\epsilon}_\theta(\mathbf{x}_t,t)\approx -\sigma_t \nabla_{\mathbf{x}} \log p(\mathbf{x}_t)
\end{equation}
where $\vec{\epsilon}_\theta(\mathbf{x}_t,t)$ can be parameterized by a network, and $\gamma_t$ is a positive weighting function \cite{song2021scorebased}.


\paragraph{DDIM/PNDM samplers.} Sampling the reverse process is somewhat independent of score function learning \cite{karras2022elucidatingDesignSp}, enabling researchers to separately study its efficiency. Most samplers are equivalent to integrating the reverse SDE or ODE with some discretization. 
The original DDPM\cite{ho2020DDPM} is effectively the same as discretizing a reverse SDE (Eq. \ref{eq:rev_special}). The deterministic DDIM sampler \cite{song2020DDIM} is equivalent to solving the probability flow ODE (Eq. \ref{eq:rev_flow}) with an Euler method, which dramatically reduced the required number of steps. More advanced numerical methods have been used to integrate Eq. \ref{eq:rev_flow}; PNDM \cite{liu2022PNDM}, the default sampler for Stable Diffusion, utilizes an RK4 method. In this work, we focus our theory and analysis on the probability flow ODE and the corresponding DDIM/PNDM samplers. We comment on how other samplers affect our results in Sec. \ref{sec:effectCfgSampler}.

\section{Salient observations about image generation}
\label{sec:observations}

\paragraph{How should we measure generation progress?}

A common way to monitor image generation progress is to observe how $\mathbf{x}_t$ (or the decoded image $G(\mathbf{x}_t)$ in the case of latent diffusion \cite{rombach2022latentdiff}) changes over time. As previously mentioned, this approach tends to show a fully-formed image unveiled from noise (Fig. \ref{fig:initial_obs}A, top row). But is this what is `actually' happening? A simple but useful alternative is to observe (appropriately scaled) \textit{differences} $k(\mathbf{x}_{t-k}-\mathbf{x}_{t})$ between close time points---the next `layer of paint' that has been added to the canvas. These often appear to be like a sketch of the final outcome early in generation, and are increasingly contaminated by noise towards the end (Fig. \ref{fig:initial_obs}A, middle row). 

A more principled alternative, proposed by 
\cite{ho2020DDPM}, is to consider the sequence of \textit{endpoint estimates} $\hat{\mathbf{x}}_0$ of the reverse diffusion trajectory. In particular, the weighted combination of the state and score
\begin{equation} \label{eq:x0hat_def_general}
\hat{\mathbf{x}}_0(\mathbf{x}(t)) := \frac{\mathbf{x}(t)+\sigma_t^2 \vec{s}(\mathbf{x}(t),t)}{\alpha_t} 
\approx \frac{\mathbf{x}(t)-\sigma_t \vec{\epsilon}_\theta(\mathbf{x}(t),t)}{\alpha_t}  
\end{equation}
provides an endpoint estimate that improves over time. Technically, this is {the minimum mean squared error (MMSE) estimator of $\mathbf{x}_0$} given $\mathbf{x}_t$ and Gaussian noise \cite{kadkhodaieEero2021denoisPrior}, which has also been called ideal denoiser \cite{karras2022elucidatingDesignSp}. We found tracking this statistic throughout reverse diffusion provides substantial insight into generation: as early as the first time step, a rough outline is visible. As time goes on, one tends to see progressively finer details filled in (Fig. \ref{fig:initial_obs}A, bottom). We observed similar results for unconditional diffusion models (trained on MNIST, CIFAR-10, and CelebA-HQ; see SI Fig. \ref{fig:CelebA_inital_obs}-\ref{fig:MNIST_inital_obs}).



\paragraph{When do different image features tend to emerge?} According to the endpoint estimate $\hat{\mathbf{x}}_0(\mathbf{x}_t)$, high-level features tend to emerge before low-level ones \cite{ho2020DDPM,karras2022elucidatingDesignSp}.
For example, when generating ``a portrait of an aristocrat'' (Fig. \ref{fig:initial_obs}A), a generic face-like shape appears, and then is refined to include blobs that correspond to hat/hair and body. Coarse facial features and various image colors emerge, facial hair appears, and the blob above the face is `reinterpreted' into a hat. Finally, high-frequency details of the face and clothes are added. Unconditional models exhibit similar behavior (SI Fig. \ref{fig:CelebA_inital_obs},\ref{fig:Church_inital_obs}). 


\paragraph{What is the shape of individual trajectories?}\label{sec:obs_rot2d_approx} We also studied the geometry of reverse diffusion trajectories. Although the dimensionality of image/latent space is quite large, individual trajectories are effectively two-dimensional: the average variance explained by the top two principal components (PCs) is 99.98\% for our CelebA model, and 99.54\% for Stable Diffusion (see Table \ref{sec:traj_geom_stats} and Fig. \ref{fig:traj_geom_stats}). To good approximation, $\mathbf{x}_t$ always remains in the plane defined by the initial noise $\mathbf{x}_T$ and the final state $\mathbf{x}_0$: the variance explained by a projection onto this 2D plane is higher than 99.2\% for all models. Furthermore, the reverse diffusion trajectory is well-approximated by a rotation within this plane (Fig. \ref{fig:initial_obs}B), i.e.
\begin{equation}
\mathbf{x}_t \approx K_t \mathbf{x}_0 + \sqrt{1 - K_t^2} \  \mathbf{x}_T
\end{equation}
where $0 \leq K_t \leq 1$, $K_T = 0$, and $K_0 = 1$. (This may not be a `true' rotation if $\mathbf{x}_0$ and $\mathbf{x}_T$ have unequal norms, which depends on training data normalization.) When $K_t = \alpha_t$, this equation explains almost all trajectory variance: 97.51\% for Stable Diffusion and 98.93\% for CelebA diffusion. 

\section{Theoretical analysis of sampling trajectories}
\label{sec:theory}
\begin{wrapfigure}[18]{r}{0.6\columnwidth}
\begin{center}
\vskip -0.25in
\centerline{\includegraphics[width=0.6\columnwidth]{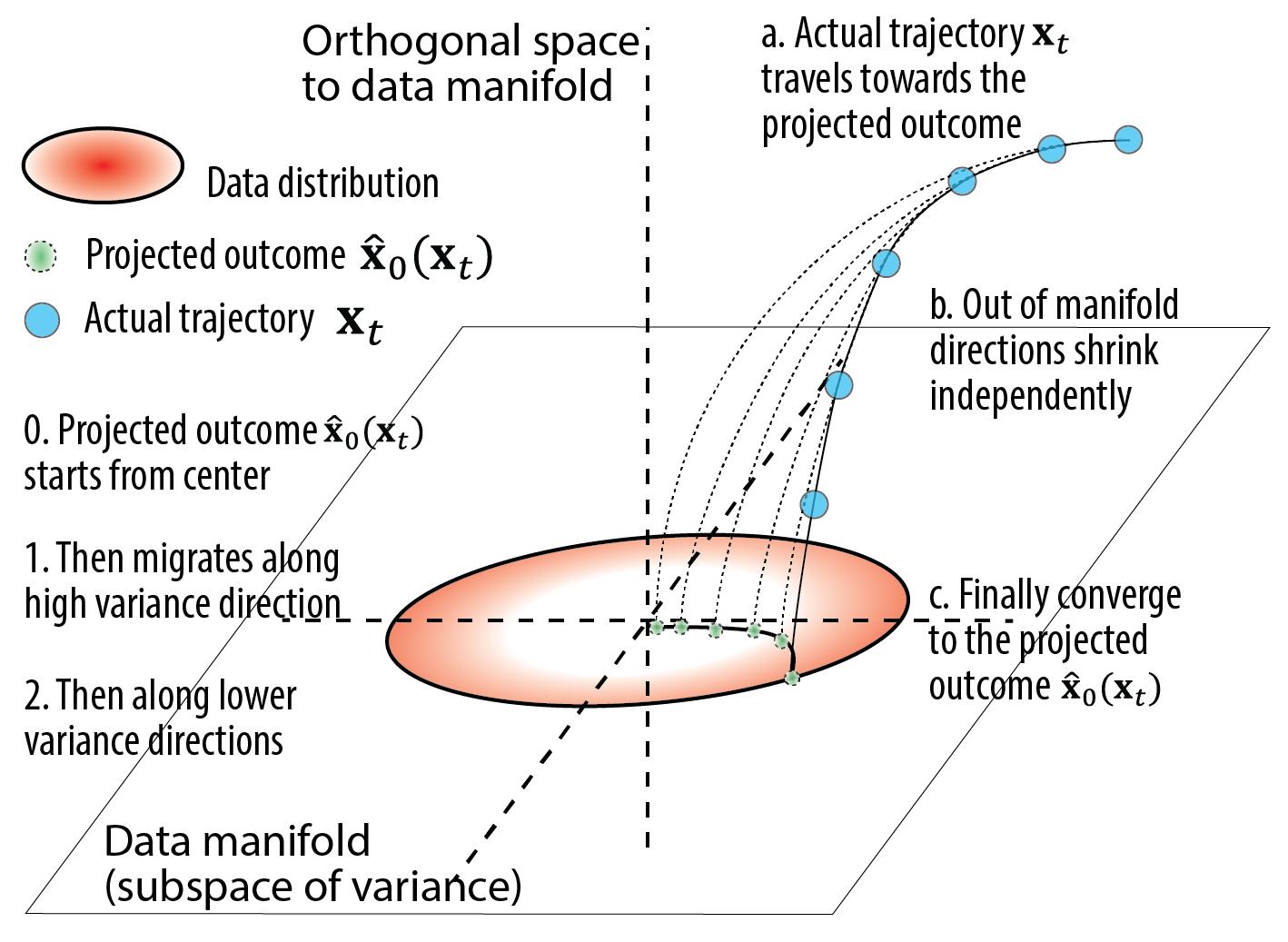}}
\vskip -0.1in
\caption{\textbf{Geometry of single mode reverse diffusion}.}
\label{fig:conceptual_pics}
\end{center}
\end{wrapfigure}
In this section, we will develop a simple but quantitative theory to explain our previous observations. For reasons of analytical tractability, we will start by studying reverse diffusion assuming that the training set is well-described by a multivariate Gaussian distribution, and that the score function is exact. Though simple, the Gaussian model allows us to explore the implications of low-dimensional manifolds and features with different variances. Further, the Gaussian mode assumption is appropriate early in reverse diffusion since the VP-SDE maps all distributions to Gaussians. It may also be reasonable for describing the short time scale dynamics of arbitrary distributions, since arbitrary distributions can be well-approximated by Gaussian mixtures, and the score functions of high-dimensional Gaussian mixtures could be locally dominated by the contribution of a single mode as long as the modes are well-separated. 

\subsection{Exact solution to a Gaussian score model}

\begin{figure*}[ht!]
\vspace{-6pt}
\begin{center}
\centerline{\includegraphics[width=\columnwidth]{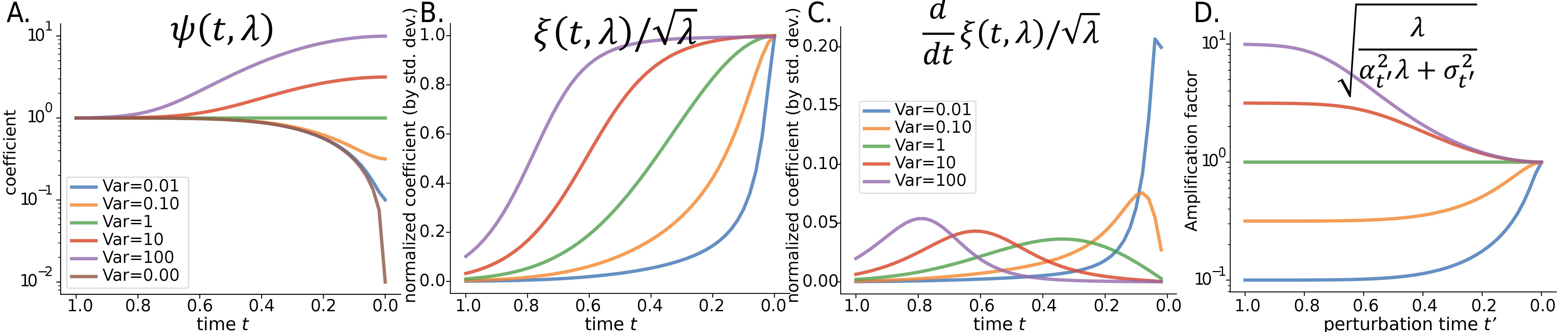}}
\vspace{-6pt}
\caption{\textbf{Analytical solution to diffusion dynamics in Gaussian case} \textbf{A.} $\psi(t,\lambda)$ governs the dynamic of state $\mathbf{x}_t$ along each each principal axis $\mathbf{u}_k$ \textbf{B.} $\xi(t,\lambda)$ governs the dynamics of endpoint estimate $\hat {\mathbf{x}}_0(\mathbf{x}_t)$ along each PC, normalized by the standard deviation $\sqrt{\lambda_k}$. \textbf{C.} Time derivative of $\xi(t,\lambda)/\sqrt{\lambda}$, highlighting the `critical period' when the feature develops. \textbf{D.} $\sqrt{\lambda/(\sigma_{t'}^2+\lambda\alpha_{t'}^2)}$, which quantify the amplification effect of a perturbation along PC $\mathbf{u}_k$ at time $t'$ (Eq.\ref{eq:y_perturb_formula}). We used the $\alpha_t$ schedule from \texttt{ddpm-CIFAR-10}.}
\label{fig:analytical_curve}
\end{center}
\vspace{-25pt}
\end{figure*}

Let $\mathbf{x}_t \in \mathbb{R}^D$, and the mean and covariance of the mode be $\vec{\mu}$ and $\vec{\Sigma}$. Assuming $\vec{\Sigma}$ has rank $r \leq D$, it has a compact singular value decomposition (SVD) $\vec{\Sigma} = \vec{U} \vec{\Lambda} \vec{U}^T$, where $\vec{U} = [\vec{u}_1, ..., \vec{u}_r]$ is a $D \times r$ semi-orthogonal matrix. The columns of $\vec{U}$ are the principal axes along which the mode varies, and their span comprises the `image manifold'. The score function $\vec{s}(\mathbf{x}, t) = \nabla_{\mathbf{x}} p(\mathbf{x}, t)$ at time $t > 0$ is the score of a Gaussian $\mathcal N(\alpha_t\vec{\mu},\sigma_t^2 \vec{I}+\alpha_t^2 \vec{\Sigma})$, so the probability flow ODE is
\begin{equation} \label{eq:flow_onemode}
\dot{\mathbf{x}}= -\beta(t)\mathbf{x}-\beta(t) \vec{s}(\mathbf{x}, t)   = -\beta(t)\mathbf{x}-\beta(t) (\sigma_{t}^2 \vec{I}+\alpha_{t}^2\vec{\Sigma})^{-1}(\alpha_{t}\vec{\mu}-\mathbf{x}) \ ,
\end{equation}
which is exactly solvable (Appendix \ref{apd:deriv_gaussian_model}). The solution is a sum of on- and off-manifold components:
\begin{equation}  \label{eq:xt_solu_psi_def}
\begin{split}
\mathbf{x}_t = \alpha_t \vec{\mu} + \frac{\sigma_t}{\sigma_T} \ \vec{y}^{\perp}_T + \sum_{k=1}^r \psi(t, \lambda_k) c_k(T) \vec{u}_k  \hspace{0.5in}
\psi(t, \lambda_k)= \sqrt{  \frac{\sigma_t^2 + \lambda_k \alpha_t^2}{\sigma_T^2 + \lambda_k \alpha_T^2} }  \\
\vec{y}^{\perp}_T=\ (\vec{I}-\vec{U}^T\vec{U})(\mathbf{x}_T-\alpha_T \vec{\mu})
\hspace{0.5in}
c_k(T) = \ \vec{u}_k^T(\mathbf{x}_T-\alpha_T \vec{\mu}) \ .
\end{split}
\end{equation}  
There are three terms: 1) the scaling up of the distribution mean; 2) the scaling down of the off-manifold component $\vec{y}^{\perp}_T$, proportional to the noise scale $\psi(t,0)=\frac{\sigma_t}{\sigma_T}$; and 3) the on-manifold movement along each eigenvector governed by $\psi(t,\lambda_k)$ (visualized in Fig. \ref{fig:analytical_curve}A). The initial condition $\mathbf{x}_T$ can be decomposed into contributions along each principal direction, an off-manifold contribution, and a $\alpha_T \vec{\mu}$ contribution. Below, we will explore how this exact solution recapitulates our observations.

But first, we note that it explicitly connects the initial noise pattern $\mathbf{x}_T$ to the final sample $\mathbf{x}_0$:
\begin{equation}
\mathbf{x}_0= \vec{\mu}+\sum_{k=1}^r \psi(0,\lambda_k)\vec{u}_k\vec{u}_k^T(\mathbf{x}_T-\alpha_T \vec{\mu}) \ .
\end{equation} 
This is reminiscent of \textit{linear filtering} adapted to the data distribution. The final location of $\mathbf{x}_0$ along each feature axis $\vec{u}_k$ is determined by the projection of the initial noise pattern onto that feature, amplified by the standard deviation $\psi(0,\lambda_k)\approx\sqrt{\lambda_k}$. Thus, it is the subtle alignment between the noise pattern and image manifold features that determine what is generated.

\subsection{Theoretical support for primary claims}
\paragraph{State trajectory.} Throughout the generation process $t=T\to0$, all $\psi(t,\lambda)$ moves from $1$ to $\sqrt{\lambda}$. So, the off-manifold component $\mathbf y^\perp$ decays to $0$ towards the end; while the on-manifold component along $\mathbf u_k$ is scaled up by $\psi(0,\lambda_k)\approx \sqrt{\lambda_k}$ which is the standard deviation along $\mathbf u_k$. From Fig. \ref{fig:analytical_curve} A, we can see the state moves along high variance dimensions first, while the low variance and off-manifold dimensions decay late until the end. This explains when visualizing the state itself, we see the well-formed image unveiled from noise till the end.

\textbf{2D rotations.} Our solution for $\mathbf{x}_t$ implies (see Appendix \ref{apd:deriv_rotation} for the derivation and more discussion)
\begin{equation*} \label{eq:rotation_formula}
\begin{split}
\mathbf{x}_t \approx \ & \alpha_t \mathbf{x}_0 + \sqrt{1 - \alpha_t^2} \ \mathbf{x}_T + \sum_{k=1}^r \left\{ \sqrt{\sigma_t^2 + \lambda_k \alpha_t^2} - \alpha_t \sqrt{\lambda_k} - \sigma_t \right\} c_k(T) \vec{u}_k \ ,
\end{split}
\end{equation*}
i.e. $\mathbf{x}_t$ dynamics tend to look like a rotation within the 2D plane formed by $\mathbf{x}_0$ and $\mathbf{x}_T$ (Fig. \ref{fig:initial_obs}B) up to on-manifold correction terms. The correction terms tend to be small; assuming $r \ll D$, and that the typical overlap between the initial noise and any given eigendirection is roughly $1/\sqrt{D}$, 
\begin{equation}
\left\Vert \mathbf{x}_t - \alpha_t \mathbf{x}_0 - \sqrt{1 - \alpha_t^2} \ \mathbf{x}_T \right\Vert^2_2 \leq \left( 1 - \frac{\sqrt{2}}{2} \right)^2 \frac{r}{D} \ll 1 \ .
\end{equation}
Interestingly, the rotation only depends on $\alpha_t$ and $\sigma_t$, and not on any properties of the mode. We empirically find, however, that it \textit{does} depend on the sampler (Section \ref{sec:effectSampler}). Generally, for any VP-SDE, the dynamics of the scaled state $\mathbf{x}_t/\alpha_t$ can be written as (see Appendix \ref{apd:deriv_rotation}) 
\begin{equation}\label{eq:scaled_state_converg}
    \frac{d}{dt} \left( \frac{\mathbf{x}_t}{\alpha_t} \right)=-\frac{\beta_t}{\sigma_t^2} \left[ \hat{\mathbf{x}}_0(\mathbf{x}_t)- \frac{\mathbf{x}_t}{\alpha_t} \right] \ .
\end{equation}
This ODE is generally difficult to solve; however, assuming $\hat{\mathbf{x}}_0(\mathbf{x}_t)$ changes more slowly than $\mathbf{x}_t$, we find that the trajectory of $\mathbf{x}_t$ can be understood as constantly `rotating' towards the endpoint estimate (Fig.\ref{fig:conceptual_pics}, dashed curves).

\textbf{Feature emergence order.} The endpoint estimate $\hat{\mathbf{x}}_0$ can be written as
\begin{equation}\label{eq:xhat_explicit}
    \hat{\mathbf{x}}_0(\mathbf{x}_t) =\  \vec{\mu} + \sum_{k=1}^r 
\xi(t,\lambda_k) c_k(T) \vec{u}_k \hspace{0.5in}
\xi(t,\lambda):=\frac{\alpha_t\lambda}{\sqrt{(\alpha_t^2\lambda + \sigma_t^2)(\alpha_T^2\lambda + \sigma_T^2)}} \ .
\end{equation}
This equation implies that $\hat{\mathbf{x}}_0$ \textit{always remains on the image manifold}, which explains why the endpoint estimates of well-trained models look like images, rather than images contaminated with noise. Since $\xi(T, \lambda) \approx 0$, $\hat{\mathbf{x}}_0$ is initially similar to the distribution average (e.g. the generic face for CelebA \cite{langlois1990face}). Note that we can exploit $\hat{\mathbf{x}}_0$ being on-manifold to \textit{infer the structure of the image manifold} (Sec. \ref{sec:exp_SD}).

The sigmoidal behavior of the $\xi(t, \lambda)$ function indicates that a given eigendirection is reflected in the endpoint estimate around when $\sigma_t=\alpha_t\sqrt{\lambda}$, i.e. when the noise variance matches the scaled signal variance (Fig. \ref{fig:analytical_curve}C). Moreover, since this happens when $\alpha _t \approx 1/\sqrt{1 + \lambda}$, image manifold features appear in order of descending variance: first the highest variance features, then the next highest, and so on. Natural images have more power and variance in low frequencies than high frequencies \cite{ruderman1994statNatImage}. For face images, features such as gender, head orientation, and skin color, have higher variance than subtle features such as glasses, facial, and hair texture \cite{wang2021aGANGeom}. 
Thus, the combination of natural image statistics and the diffusion process explain why features such as the layout of a scene and the `semantic' features of faces are specified first in the endpoint estimate, or why generation is outline-first, details later.

\paragraph{Effect of perturbations.} Finally we examined the effect of perturbation and feature commitment. Suppose at time $t'\in (0,T)$ the off-manifold directions are perturbed by $\delta \mathbf{y}^{\perp}$, and the on-manifold direction coefficients are perturbed by amounts $\delta c_k$; then their effect on the generated image $\mathbf{x}_0$ is 
\begin{equation} \label{eq:y_perturb_formula}
\Delta \mathbf{x}_0 =\sum_{k=1}^r \frac{\psi(0,\lambda_k)}{\psi(t',\lambda_k)} \delta c_k \mathbf{u}_k= \sum_{k=1}^r \sqrt{\frac{\lambda_k}{\sigma_{t'}^2+\lambda_k\alpha_{t'}^2}}  \delta c_k \mathbf{u}_k
\end{equation}
Thanks to denoising, the off-manifold perturbation has no effect on the sample, while on-manifold perturbations have \textit{maximal effect at different periods} (Fig. \ref{fig:analytical_curve} D). Perturbations of high variance features ($\lambda>1$) are amplified at the start and then decayed; perturbation along low variance features ($\lambda<1$) has a reduced effect until the end. This time-dependent `filtering' explains the classic finding \cite{ho2020DDPM} that during the reverse diffusion process when noise is injected at different steps, early perturbation creates variations of layout and semantic features and late perturbation varies details. 

\paragraph{Summary.} 
The examination of the solution to Eq.\ref{eq:forward_special} suggests a conceptual understanding of the generation process, as depicted in Fig. \ref{fig:conceptual_pics}: The endpoint estimate $\hat{\mathbf{x}}_0$ travels on the image manifold, starting from the center of the distribution, moving first along the high variance axes, and then the lower variance axes; concurrently, the state $\mathbf{x}_t$ in the ambient space keeps rotating towards the evolving endpoint estimate.

\subsection{Beyond the Gaussian score function approximation}
\paragraph{Diffusion as retrieval: dynamics of the state with general point cloud.}\label{sec:delta_gmm_theory} 
In practice, diffusion models are trained using a finite set of points $\{\mathbf y_i\},i=1,...,N$. Thus, without augmentation, the training distribution is effectively a collection of delta functions or a mixture of Gaussian with negligible width. $p(\mathbf{x}_0)=\frac{1}{N}\sum_i\delta(\mathbf{x}_0-\mathbf y_i)$. We proved that (see Appendix \ref{apd:deriv_nongaussian}), with the same forward process (Eq.\ref{eq:forward_special}), the endpoint estimate is a weighted average of the `nearest' training data, while the score is locally equivalent to that of an isotropic Gaussian centered at $\hat{\mathbf{x}}_0(\mathbf{x}_t)$ diffused to time $t$
\begin{align}\label{eq:gmm_delta_score}
    \hat{\mathbf{x}}_0(\mathbf{x}_t)=&\sum_i w_i(\mathbf x_t,t)\mathbf{y}_i\;\;,\;\;
    w_i(\mathbf x_t,t):=\mbox{softmax}\big(\big\{-\frac{\alpha_t^2}{2\sigma_t^2}\|\mathbf{y}_i - \frac{\mathbf x_t}{\alpha_t}\|^2\big\}\big)_i \\
    \mathbf{s}(\mathbf{x},t)=&\frac{-\mathbf{x}_t+\alpha_t\hat{\mathbf{x}}_0(\mathbf{x}_t)}{\sigma_t^2} \ .
\end{align}
The weights are defined by the softmax of the negative squared distance between $\mathbf x_t/\alpha_t$ and all data points, with the temperature set at ${2\sigma_t^2}/{\alpha_t^2}$. Consistent with the Gaussian case, at the start of generation, the temperature $2\sigma_T^2/\alpha_T^2$ is much higher than the distances, so the estimated outcome corresponds to the mean of all data points. As the generation progresses, $w_i(\mathbf x_t,t)$ focuses on the set of training samples that are closest to $\mathbf x_t/\alpha_t$. Towards the end, the temperature approaches $0$, and the softmax focuses on one training sample---the one that generation converges to.

In summary, we can see when the \textit{score is exact}, the reverse diffusion process is \textit{equivalent to an iterative image retrieval process} for a discrete dataset: the scaled state $\mathbf{x}_t/\alpha_t$ migrates towards the weighted average of a subset of data points, and gradually focuses the weights on the nearest data points until it finally converges to one data point. But a priori, it is unclear if this matches what neural network score approximators learn. 

\section{Validating the normative theory on actual diffusion models}

\label{sec:exp_simple}

In this section, we aim to test the extent to which the Gaussian theory can accurately predict actual reverse diffusion trajectories, and to compare the quality of its predictions to two other score function approximations: the delta function mixture described in the previous section, and the Gaussian mixture. Although real image distributions are certainly not Gaussian, a Gaussian approximation may be a reasonable description of the beginning of reverse diffusion, when the data distribution is sufficiently `blurred'. On the other extreme, a mixture of delta functions centered on the training data is an important point of comparison because it represents the optimal solution to Eq. \ref{eq:denoising_score_obj} if no data augmentation is used (Sec.\ref{sec:delta_gmm_theory}); deviations of the score network from this model suggest that the network does not converge to the `optimal' score, and may hint at how they learn to generalize. 

\begin{figure}[!ht]
\vspace{-20pt}
\begin{center}
\centerline{\includegraphics[width=\columnwidth]{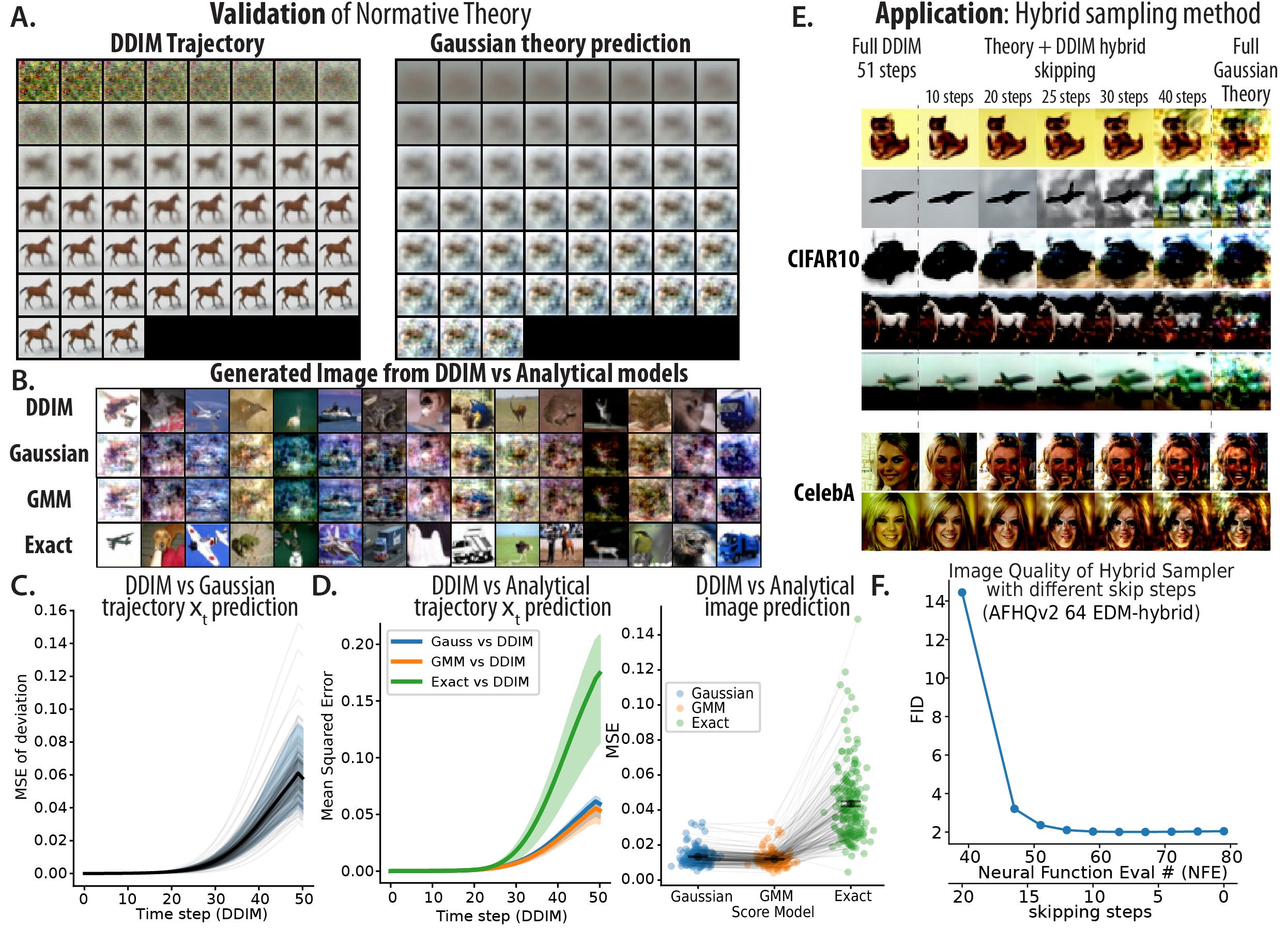}}\vspace{-10pt}

\caption{\textbf{Comparing analytical solution to DDIM sampling for CIFAR-10 diffusion model}. \textbf{A.} $\hat{\mathbf{x}}_0(\mathbf{x}_t)$ of a DDIM trajectory and the Gaussian solution with the same initial condition $\mathbf{x}_T$. \textbf{B.} Samples generated by DDIM and the analytical theories from the same initial condition. \textbf{C.} Mean squared error between the $\mathbf{x}_t$ trajectory of DDIM and Gaussian solution. \textbf{D.} Comparing the state trajectory and final sample of three normative models (Gaussian, GMM, exact) with DDIM. \textbf{E.} Hybrid sampling method combines Gaussian theory prediction with DDIM. \textbf{F.} Image quality of the hybrid method (FID score) as a function of different numbers of skipped steps for EDM model and sampler \cite{karras2022elucidatingDesignSp} (see Appendix \ref{apd:fid_method}).}

\label{fig:CIFAR_theory_valid}
\end{center}\vspace{-25pt}
\end{figure}

\paragraph{Gaussian solution predicts early diffusion trajectory.}
To test the Gaussian approximation, we numerically computed the mean and covariance of training samples in pixel space for models trained on MNIST, CIFAR-10, and CelebA-HQ. Then Eq. \ref{eq:xt_solu_psi_def} was used with these means and covariances to predict the evolution of $\mathbf{x}_t$. We found that the early phase of reverse diffusion is well-predicted by the Gaussian solution (Fig. \ref{fig:CIFAR_theory_valid}A-B). Visually, as the low-frequency information is determined early on, the `layout' of the final image is also well-predicted, while high-frequency details such as edges are less well-predicted. The deviation between the predicted and actual trajectory grows large at around 20 reverse diffusion steps ($t=0.6$, Fig. \ref{fig:CIFAR_theory_valid}C); we interpret this as the moment when the \textit{single mode} assumption breaks down, and the trajectory starts to be guided by a more complicated distribution. (For MNIST and CelebA see Fig. \ref{fig:MNIST_theory_valid_supp},\ref{fig:celebA_theory_valid_supp}. ) We also found that the dynamics along off-manifold directions are well-predicted by the Gaussian solution (Fig. \ref{fig:off-manifold-pred}). 

This result bears interesting implications for score function approximation. It suggests that even for natural images, at high noise scales, $p(\mathbf{x}_t)$ is indistinguishable from a multivariate Gaussian. Therefore, the early phase score function can be effectively approximated by the Gaussian score, which is an \textit{affine function} of $\mathbf{x}$, specifically $\vec{\Sigma}^{-1}(\vec{\mu}-\mathbf{x})$. This raises doubts about the necessity of a nonlinear neural network for this phase (Sec.\ref{sec:app_teleportation}). 

\paragraph{Gaussian solution predicts late trajectory better than the exact score model.}

Taking a step further, we tested two other models on MNIST and CIFAR-10: the 10-mode Gaussian mixture model (\textit{GMM}), where each class is fit by one Gaussian mode; and the \textit{exact} score model (Sec.\ref{sec:delta_gmm_theory}), where a delta mode is defined on each training image. For these models, the score function can be evaluated precisely, but the trajectory has no closed-form solution, so we used an off-the-shelf RK4 ODE solver to integrate it. (For the score of the Gaussian mixture model, see App.\ref{apd:gmm_deriv}.) While all models predict the early trajectory well, surprisingly, we found that the trajectory predicted by the exact score deviates from the actual DDIM trajectory, and does so even earlier than the Gaussian solution (Fig. \ref{fig:CIFAR_theory_valid}D). Visually, both the Gaussian model and GMM predict the generated image better than the exact score model (Fig. \ref{fig:CIFAR_theory_valid}B), with a significantly lower MSE ($p<10^{-30}$). Though the Gaussian model and GMM have comparable predictions, GMM has slightly lower error ($p<10^{-10}$). Thus, we can infer something about the structure of the learned score function through these trajectories. This result implies that the actual score function learned by optimizing Eq. \ref{eq:denoising_score_obj} is different from the exact score---especially late in reverse diffusion---and that it is more similar to the `blurrier' score of a Gaussian or Gaussian mixture, possibly due to the regularizing effect of our neural network function approximator. Similar results were observed for the MNIST model (Fig. \ref{fig:MNIST_gmm_theory_valid}).

\section{Applications: accelerating sampling,  characterizing image manifold}
\label{sec:exp_SD}

\subsection{Accelerating unconditional diffusion by teleportation}\label{sec:app_teleportation}

We can exploit the fact that the Gaussian analytical solution provides a surprisingly good approximation to the early part of the sampling trajectory by using the solution to `teleport' to time $t$. Namely, instead of evaluating the score function approximated by neural network $\vec{\epsilon}_\theta$ and integrating the probability flow ODE, we can use the Gaussian prediction for $\mathbf{x}_t$ instead, where the $\vec{\mu}$ and $\vec{\Sigma}$ used are those of the training set. In principle, this speedup can be combined with any deterministic or stochastic sampler. Here, we showcase its effectiveness with DDIM and Heun's sampler \cite{karras2022elucidatingDesignSp}. %

We tested this hybrid sampler on unconditional diffusion models of MNIST, CIFAR-10, and CelebA-HQ. For the MNIST and CIFAR-10 models, we can easily skip \textit{40\% of the initial steps} with the Gaussian solution without much of a perceptible change in the final sample (Fig.\ref{fig:CIFAR_theory_valid}E). Quantitatively, we found skipping up to 40\% of the initial steps can even slightly decrease the Frechet Inception Distance score, and improve the quality of generated samples (Fig.\ref{fig:CIFAR10_FID_supp}). 
For models of higher resolution data sets like CelebA-HQ, we need to be more careful; skipping more than 20\% of the initial steps will induce some perceptible distortions in the generated images (Fig.\ref{fig:CIFAR_theory_valid}E bottom), which suggests that the Gaussian approximation is less effective for larger images. The reason may have to do with a low-quality covariance matrix estimate, which could arise from the small number of training images compared to the effective dimensionality of the image manifold.

With the more optimized pre-trained diffusion models in EDM and Heun's sampler \cite{karras2022elucidatingDesignSp}, we can still reliably skip 15-30\% neural function evaluation time, while maintaining FID scores competitive with the state-of-the-art level. Specifically, we achieved FID score of $1.934$ on CIFAR10 with 25 NFEs, and FID score of $2.026$ on AFHQv2 64 with 59 NFEs (Fig.\ref{fig:CIFAR_theory_valid}F , full results in Fig.\ref{fig:hybrid_fid_fullresult} in Sec.\ref{apd:EDM_heun_accelr}). This shows that our hybrid acceleration trick is generally effective even when combined with state-of-the-art diffusion models and samplers.

\subsection{Characterizing image manifold by analyzing sampling trajectory}

\begin{wrapfigure}[26]{r}{0.6\columnwidth}
\vspace{-15pt}
\begin{center}
\centerline{\includegraphics[width=0.6\columnwidth]{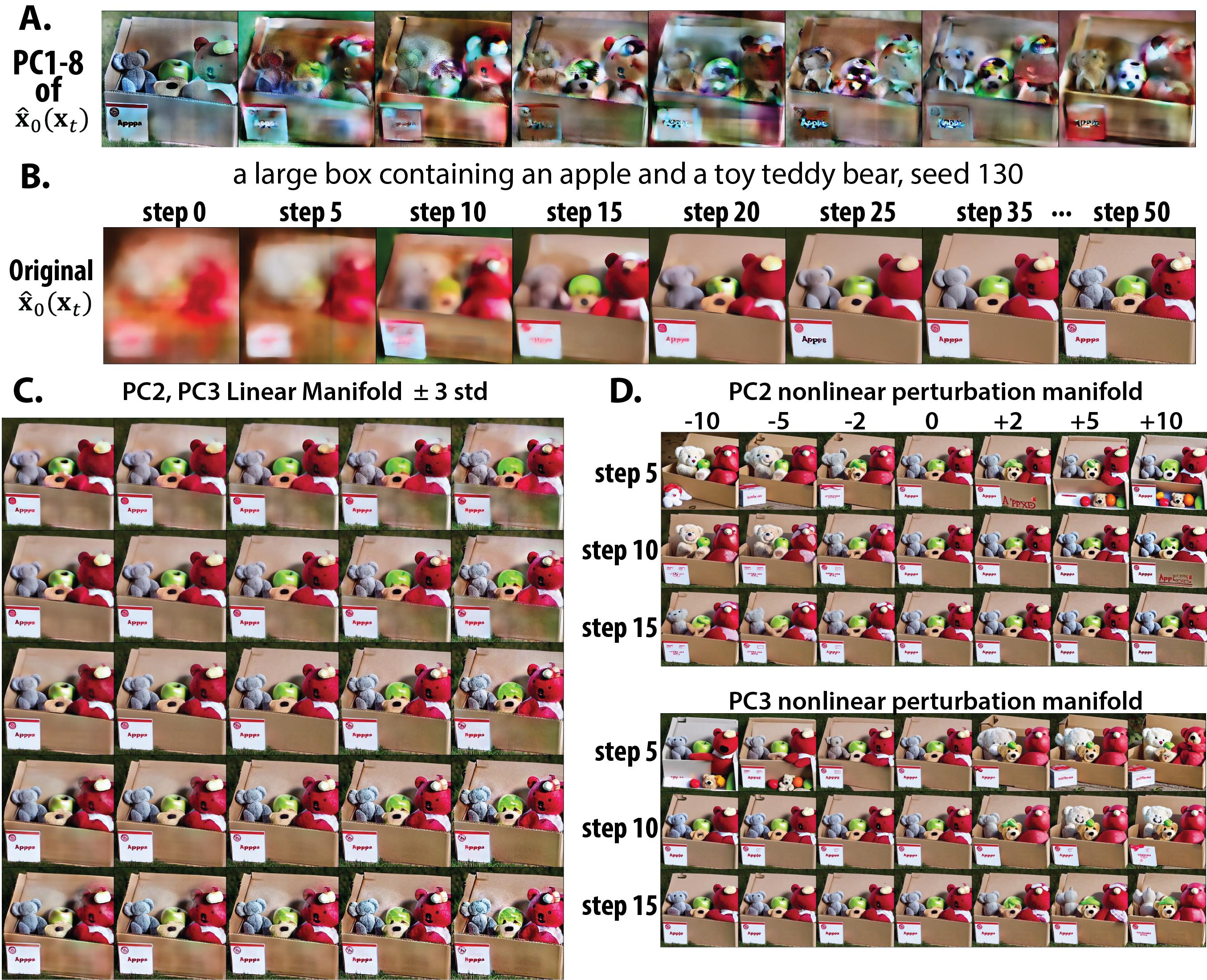}}
\vspace{-6pt}
\caption{\textbf{Stable Diffusion: Local manifold map}. \textbf{A.} PCs of the projected outcome trajectory $\hat{\mathbf{x}}_0(\mathbf{x}_t)$ are on-manifold. 
\textbf{B.} Trajectory of endpoint estimate $G(\hat{\mathbf{x}}_0(\mathbf{x}_t))$ \textbf{C.} Perturbation by PC2 and PC3; notice an apple morphing into a teddy bear. \textbf{D.} Perturbing trajectory along PC2 or PC3 during reverse diffusion. Rows: different perturbation times. Columns: different magnitudes.} 
\label{fig:diffusion_pertub}
\end{center}
\vspace{-10pt}
\end{wrapfigure}

For a large text-to-image conditional model like Stable Diffusion \cite{rombach2022latentdiff}, which given a text prompt $\tau$ samples from the distribution $p(\mathbf{x} | \tau)$, we cannot easily apply our Gaussian theory since we do not have easy access to the conditional distribution. However, we can still leverage the qualitative insight that the endpoint estimate  $\hat{\mathbf{x}}_0$ remains on the image manifold; in particular, $\hat{\mathbf{x}}_0$ trajectories in principle reflect interesting manifold directions. When we visualized the PC directions through the decoder, they appeared to be a clean variation of the target image, i.e. a tangent vector to the image manifold (Fig. \ref{fig:diffusion_pertub}A, \ref{fig:SD_PCA_vis}). Consistent with our theory (Eq.\ref{eq:y_perturb_formula}), we found perturbations in these PC directions more effectively produce nontrivial image variants than perturbations in random directions (Fig. \ref{fig:SD_perturb_PC05},\ref{fig:SD_perturb_RND009},\ref{fig:SD_perturb_summary}).

This gave us an effective way to find local on-manifold directions using a single sampling trajectory. We found two ways to apply these directions: 1) linearly perturb the final state $\mathbf{x}_0$ along these directions; and 2) perturb the state $\mathbf{x}_t$ during reverse diffusion (Eq.\ref{eq:y_perturb_formula}). The first method can visualize the local linear image manifold around the generated image (Fig.\ref{fig:diffusion_pertub}C). But since the image manifold is not linear, traveling too far along PC directions will induce distortion. In contrast, the second method can visualize the local nonlinear manifold, showing that object identity and layout can undergo dramatic changes along these nonlinear axes (Fig.\ref{fig:diffusion_pertub}D). This proved the principle that we can use sampling trajectories to characterize the `image manifold' embedded in the diffusion model. 


\section{Discussion}


To what extent do our main findings---low-dimensional trajectories, outline-first and details-later image generation, and increasing commitment to image elements---hold true for other diffusion model variants and generative models? Simulating reverse SDEs (Eq. \ref{eq:rev_special}) instead of ODEs should not yield many qualitative differences; instead of linear ODEs, one has similarly-behaved OU processes. Antognini and Sohl-Dickstein \cite{antognini2018walk} show OU trajectories are also very low-dimensional. Arguments similar to ones we have made suggest other kinds of models may possess a simple analytic description early in generation. For example, Xu and Liu et al.'s Poisson flow generative models \cite{xu2022poisson,pfgm2023} may feature smeared-out charge distributions.

Our observations provide a phenomenological bridge between diffusion models and GANs. In the GAN literature, the idea of generating images by progressively modeling low-to-high resolution is well-established, e.g. by successful architectures like Progressive Growing GAN and StyleGAN \cite{karras2017PGGAN,karras2020StyleGAN2}. In this paradigm, early layers synthesize the rough layout of the image from noise, while the last few layers add realistic details. 
We showed that when looking at the projected outcome $\hat{\mathbf{x}}_0$, diffusion models have an intriguingly similar generative process. Moreover, the effect of injecting noise at different times is similar to injecting noise into different layers of e.g. StyleGAN. 
In this analogy, the sampling steps of diffusion are equivalent to GAN layers. This connection may facilitate shared techniques to understand both of them. 

The latent manifold geometry of GANs and diffusion models may also be similar. Wang and Ponce \cite{wang2021aGANGeom} found that perturbing large variance directions of a latent space metric has large and interpretable effects on image generation in GANs. If we interpret the covariance matrix of a Gaussian mode as inducing a metric on the latent space of diffusion models, our perturbation-related observations can be cast in a similar light.

Our finding that the early diffusion trajectory is well-predicted by the Gaussian model is somewhat surprising. It calls for more attention to normative analyses of the score function (i.e. given some data, what should score be?). We showed that in the early phase, it has a simple linear structure that does not required advanced function approximation. With a deeper understanding of the score, we can build a better neural architecture that can approximate it more efficiently.

\bibliography{diffbib_new}
\bibliographystyle{unsrtnat}

\newpage
\appendix
\onecolumn
\section{Diffusion models used in numerical experiments} \label{apd:model_details}

\begin{table}[!ht]
\caption{\textbf{Diffusion models used for this paper's numerical experiments}. \\$\dag$: The MNIST diffusion model uses the upsampled $3\times 32\times 32$ RGB pixel space as sample space, while the original MNIST data set consists of $28\times 28$ single channel black and white images. Thus the effective dimensionality of these images is around $784$. }
\label{sample-table}
\begin{center}
\begin{small}
\begin{sc}
\begin{tabular}{lcccr}
\toprule
Data set & Hugging Face model\_id & Dimensionality & Latents? & Conditional? \\
\midrule
MNIST  & \text{dimpo/ddpm-mnist} & 3x32x32=3072 $\dag$ &  $\times$ & $\times$ \\
CIFAR-10  & \text{google/ddpm-cifar10-32} & 3x32x32=3072 &  $\times$ & $\times$ \\
Lsun-Church  & \text{google/ddpm-church-256} & 3x256x256 = 196,608 & $\times$ & $\times$\\
CelebA-HQ  & \text{google/ddpm-celebahq-256} & 3x256x256 = 196,608 & $\times$ & $\times$\\
LAION-2B & runwayml/stable-diffusion-v1-5 &   4x64x64 = 16,384 & $\surd$ & $\surd$ \\
\bottomrule
\end{tabular}
\end{sc}
\end{small}
\end{center}
\vskip -0.1in
\end{table}

\section{Supplementary Results}

\subsection{Visualizing image generation progress for other diffusion models}
\label{sec:vis_image_gen_process}
\begin{figure}[!ht]
\begin{center}
\centerline{\includegraphics[width=0.9\columnwidth]{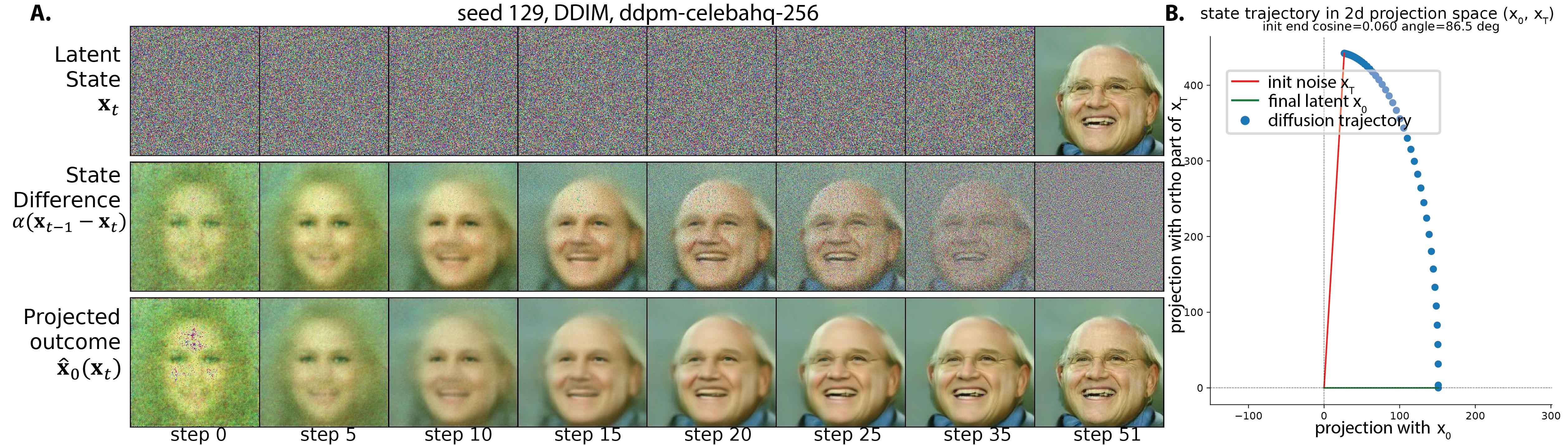}}
\caption{\textbf{Visualizing image generation process for DDPM-CelebA model}. Same layout as Fig.\ref{fig:initial_obs}.}
\label{fig:CelebA_inital_obs}
\end{center}
\vskip -0.3in
\end{figure}
\begin{figure}[!ht]
\begin{center}
\centerline{\includegraphics[width=0.9\columnwidth]{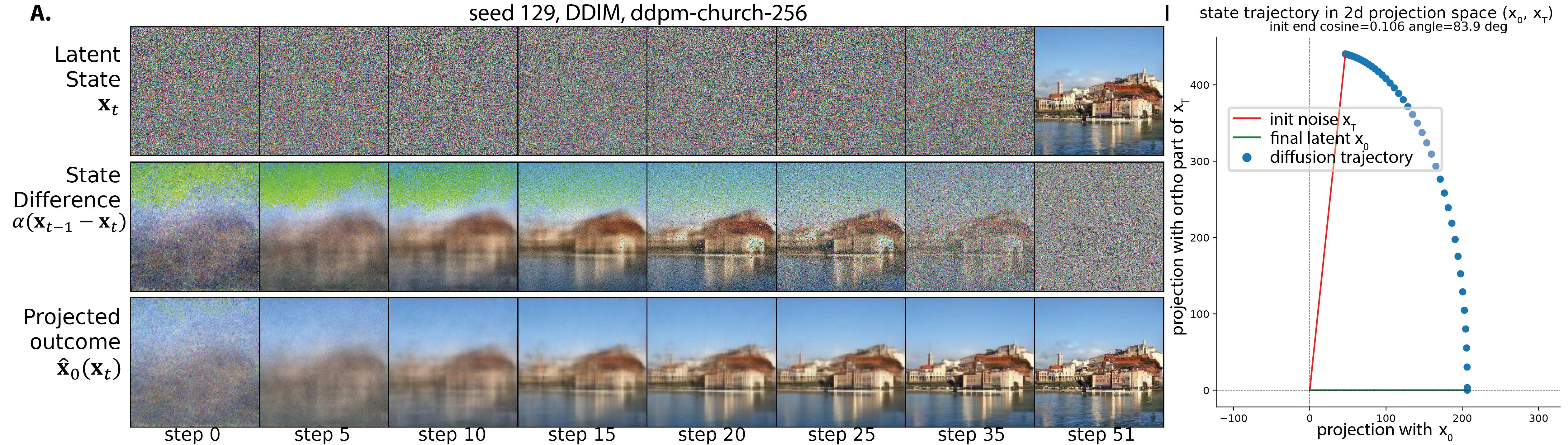}}
\caption{\textbf{Visualizing image generation process for DDPM-Church model}. Same layout as Fig.\ref{fig:initial_obs}.}
\label{fig:Church_inital_obs}
\end{center}
\vskip -0.3in
\end{figure}
\begin{figure}[!ht]
\begin{center}
\centerline{\includegraphics[width=0.9\columnwidth]{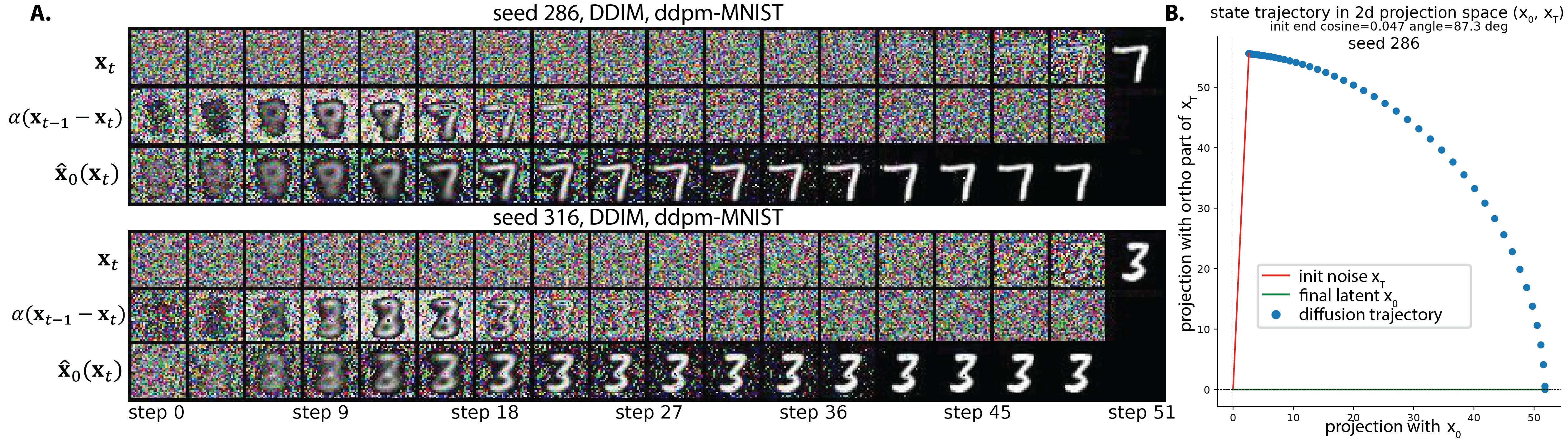}}
\caption{\textbf{Visualizing image generation process for DDPM-MNIST model}. Same layout as Fig.\ref{fig:initial_obs}.}
\label{fig:MNIST_inital_obs}
\end{center}
\vskip -0.3in
\end{figure}
\begin{figure}[!ht]
\begin{center}
\centerline{\includegraphics[width=0.9\columnwidth]{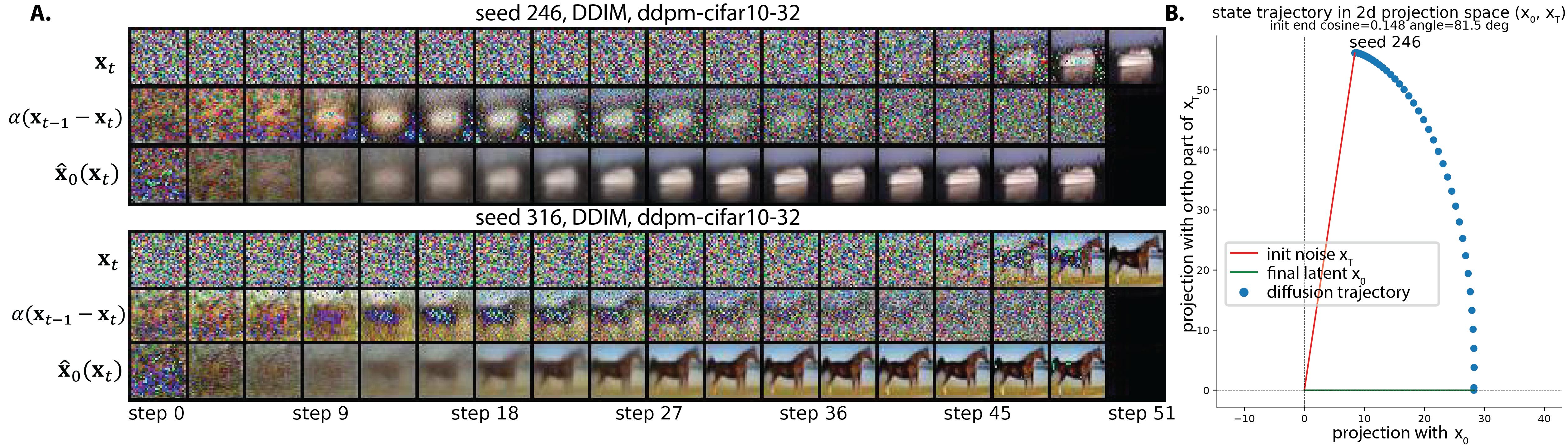}}
\caption{\textbf{Visualizing image generation process for DDPM-CIFAR-10 model}. Same layout as Fig.\ref{fig:initial_obs}.}
\label{fig:CIFAR_inital_obs}
\end{center}
\vskip -0.3in
\end{figure}

\clearpage

\subsection{Latent space trajectory geometry statistics}
\label{sec:traj_geom_stats}
\begin{table}[!ht]\label{tab:traj_geom_stats}
\caption{\textbf{Trajectory geometry statistics for different diffusion models}. Residual variance is here defined to be the squared norm of the error vector divided by the squared norm $\|\vec{x}_t\|^2$. Residual variance is computed for three approximations: 1) projecting a trajectory onto the top 2 PCs, 2) projecting a trajectory onto the plane spanned by $\vec{x}_0$ and $\vec{x}_T$, 3) approximating the trajectory by a rotation $\alpha_t \vec{x}_0+\sqrt{1-\alpha_t^2}\vec{x}_T$. Effective dimensionality is defined as the number of PCs needed to account for 99.9\% of the variance. Shown are the effective dimensionalities of the trajectory $\vec{x}_t$, difference $\vec{x}_{t-1}-\vec{x}_t$, and the U-Net output $\vec{\epsilon}_\theta(\vec{x}_t)$. All sampling used 51 time steps.} 
\begin{center}
\begin{small}
\begin{sc}
\begin{tabular}{cl|lll|llllll}
\toprule
                      &         & \multicolumn{3}{c}{Residual Variance}                                                                                                                                                      & \multicolumn{3}{c}{Dim. for 99.9 Var.} \\
                      & Sampler & \begin{tabular}[c]{@{}l@{}}top 2 PC\\ proj.\end{tabular} & \begin{tabular}[c]{@{}l@{}}$\vec{x}_0,\vec{x}_T$ \\ proj.\end{tabular} & \begin{tabular}[c]{@{}l@{}}$\vec{x}_0,\vec{x}_T$ \\ rotation\end{tabular} & $\vec{x}_t$      & $\Delta \vec{x}_t$      & $\vec{\epsilon}_{\vec{\theta}}(\vec{x}_t)$      \\
                      \midrule
ddpm-mnist            & DDIM    & 0.08\%                                                        & 0.26\%                                                          & 1.70\%                                                   & 2       & 5             & 8        \\
ddpm-cifar10-32       & DDIM    & 0.05\%                                                        & 0.30\%                                                          & 1.01\%                                                   & 2       & 4             & 7        \\
ddpm-church-256       & DDIM    & 0.05\%                                                        & 0.31\%                                                          & 1.09\%                                                   & 2       & 5             & 8        \\
ddpm-celebahq-256     & DDIM    & 0.02\%                                                        & 0.10\%                                                          & 1.07\%                                                   & 2       & 4             & 7        \\
stable-diffusion-v1-5 & PNDM    & 0.46\%                                                        & 0.81\%                                                          & 2.49\%                                                   & 5       & 34            & 28      \\
\bottomrule
\end{tabular}
\end{sc}
\end{small}
\end{center}
\vskip -0.1in
\end{table}

\begin{figure}[!ht]
\vskip 0.2in
\begin{center}
\centerline{\includegraphics[width=0.65\columnwidth]{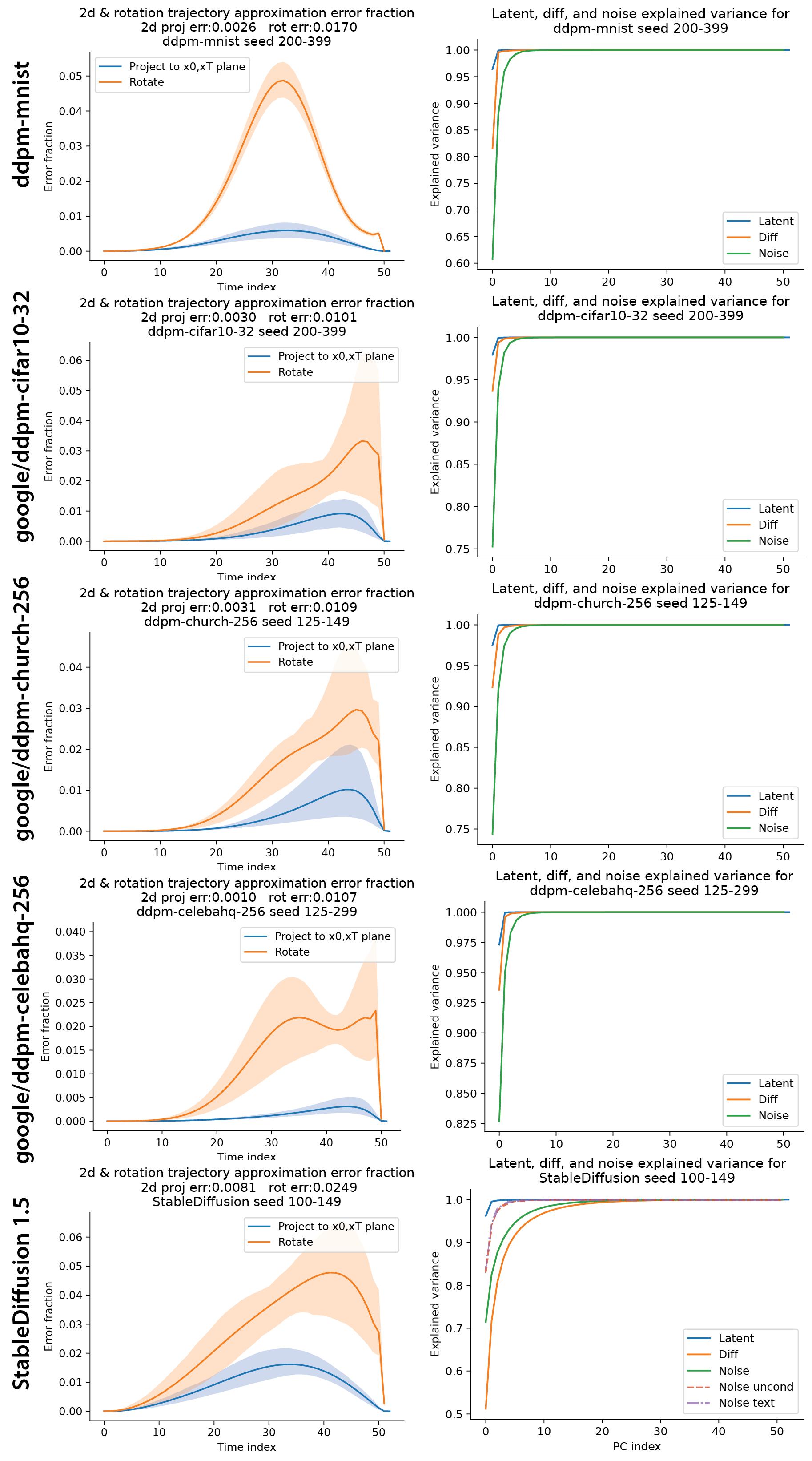}}
\caption{\textbf{Geometry of diffusion sampling trajectories} for 5 different diffusion models. \textbf{Left panel}: the error fraction of the projection onto the 2D plane defined by $\vec{x}_0$ and $\vec{x}_T$, and the error fraction under the rotation approximation. \textbf{Right panel}: the cumulative explained variance of PCs for the trajectory $\vec{x}_t$, state differences $\vec{x}_{t-1}-\vec{x}_{t}$, and the output from the U-Net $\vec{\epsilon}_\theta(\vec{x}_t)$. 2 PCs explained almost all variance for the $\vec{x}_t$ trajectory, while the state difference $\vec{x}_{t-1}-\vec{x}_t$ and U-Net outputs are higher dimensional. For quantification see Tab. \ref{tab:traj_geom_stats}. }
\label{fig:traj_geom_stats}
\end{center}
\vskip -0.4in
\end{figure}

\clearpage

\subsection{Validation of the single mode theory on CIFAR, MNIST, and CelebA}

\begin{figure}[!ht]
\begin{center}
\centerline{\includegraphics[width=0.4\columnwidth]{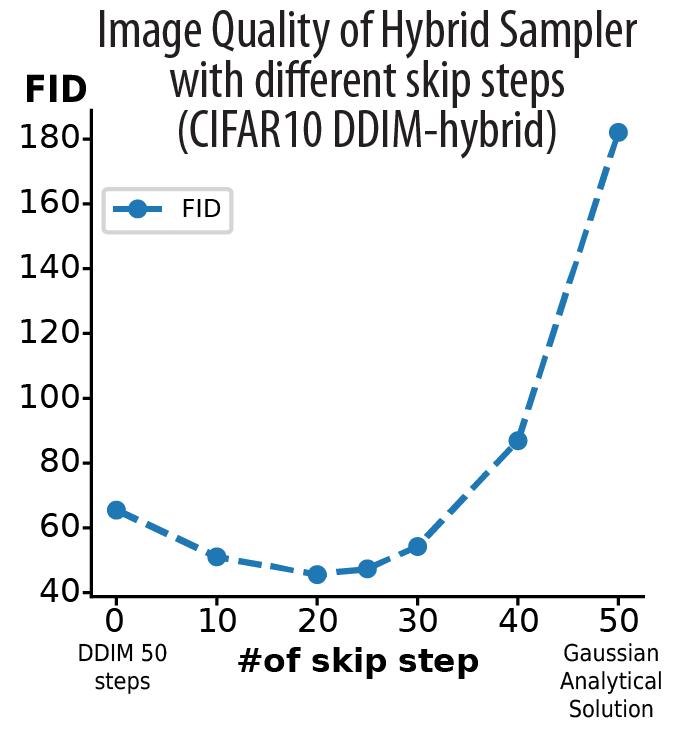}}
\caption{\textbf{Comparing analytical solution and actual diffusion process for the DDPM-CIFAR10 model} Image quality of the hybrid method (FID score) as a function of different numbers of skipped steps for the DDIM sampler. Note that the FID score of the original diffusion model without skipping is also not optimal.}
\label{fig:CIFAR10_FID_supp}
\end{center}
\vskip -0.3in
\end{figure}

\begin{figure}[!ht]
\begin{center}
\centerline{\includegraphics[width=0.9\columnwidth]{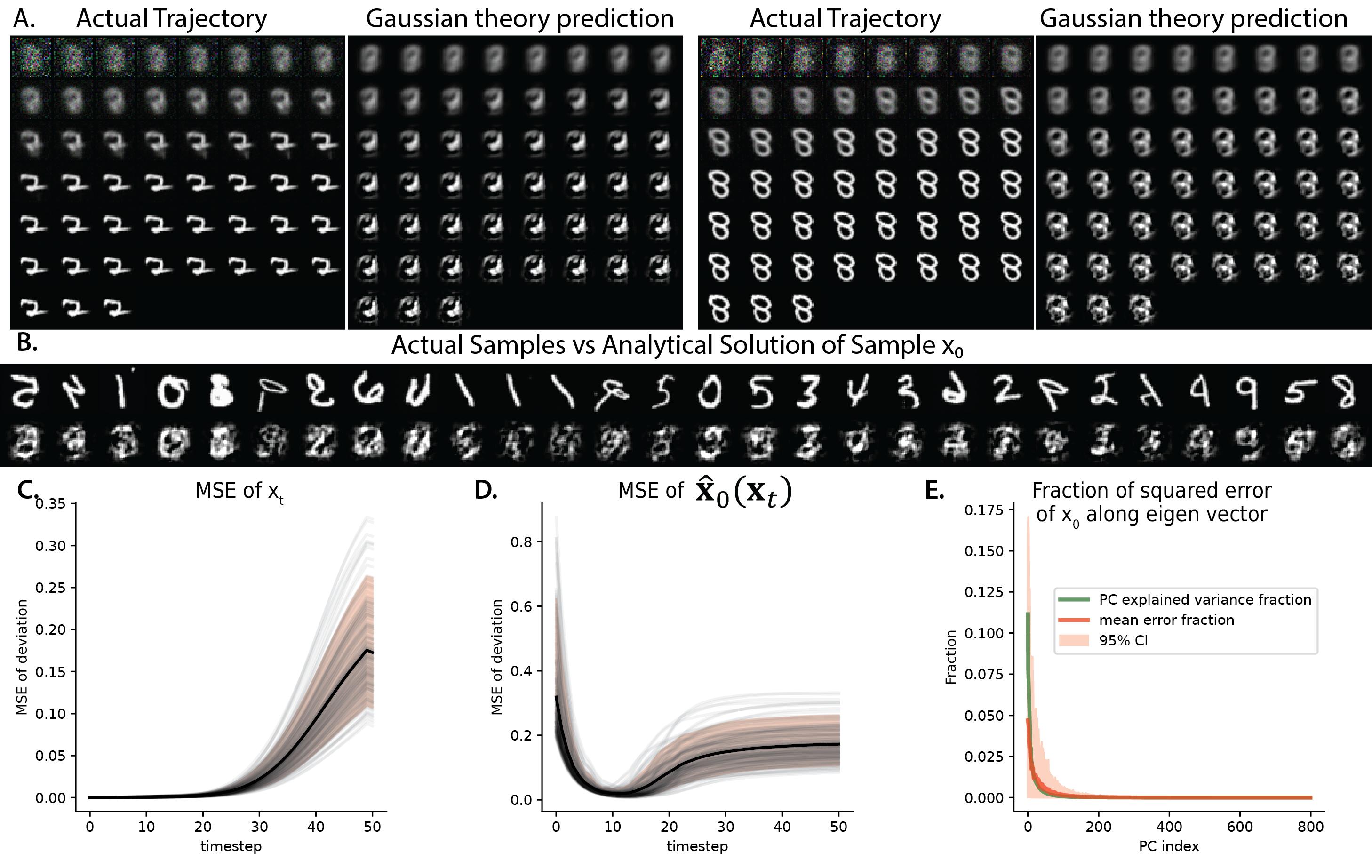}}
\caption{\textbf{Comparing analytical solution and actual diffusion process for the DDPM-MNIST model}. \textbf{A.} True and predicted endpoint estimate $\hat{\vec{x}}_0(\vec{x}_t)$ throughout reverse diffusion. \textbf{B.} Collection of samples of diffusion-generated images and the corresponding images predicted by our analytical theory. \textbf{C.} Mean squared error of the trajectory $\vec{x}_t$. \textbf{D.} MSE of the endpoint estimate during diffusion. \textbf{E.} $\vec{x}_0$ prediction error along each eigendimension.}
\label{fig:MNIST_theory_valid_supp}
\end{center}
\vskip -0.3in
\end{figure}

\begin{figure}[!ht]
\begin{center}
\centerline{\includegraphics[width=0.9\columnwidth]{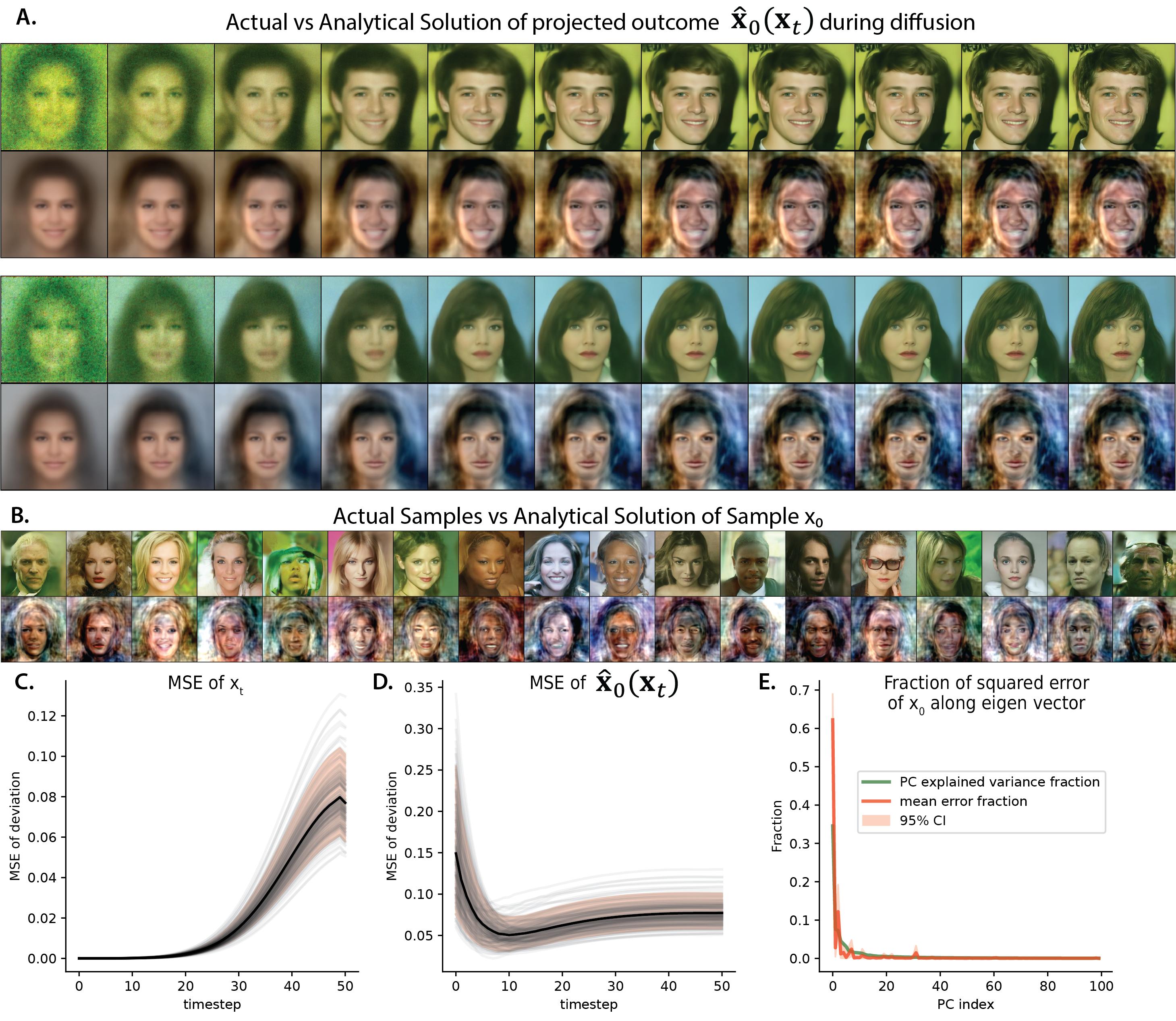}}
\caption{\textbf{Comparing analytical solution and actual diffusion process for the DDPM-CelebA model}. Same layout as Fig.\ref{fig:MNIST_theory_valid_supp}. Note that in \textbf{A.} the general layout and shading around the face is consistent between the theory and actual diffusion trajectory. Note in \textbf{B}, the zoomed-out version of the predicted and actual sample look highly similar.}
\label{fig:celebA_theory_valid_supp}
\end{center}
\vskip -0.3in
\end{figure}
\begin{figure}[!ht]
\begin{center}
\centerline{\includegraphics[width=0.9\columnwidth]{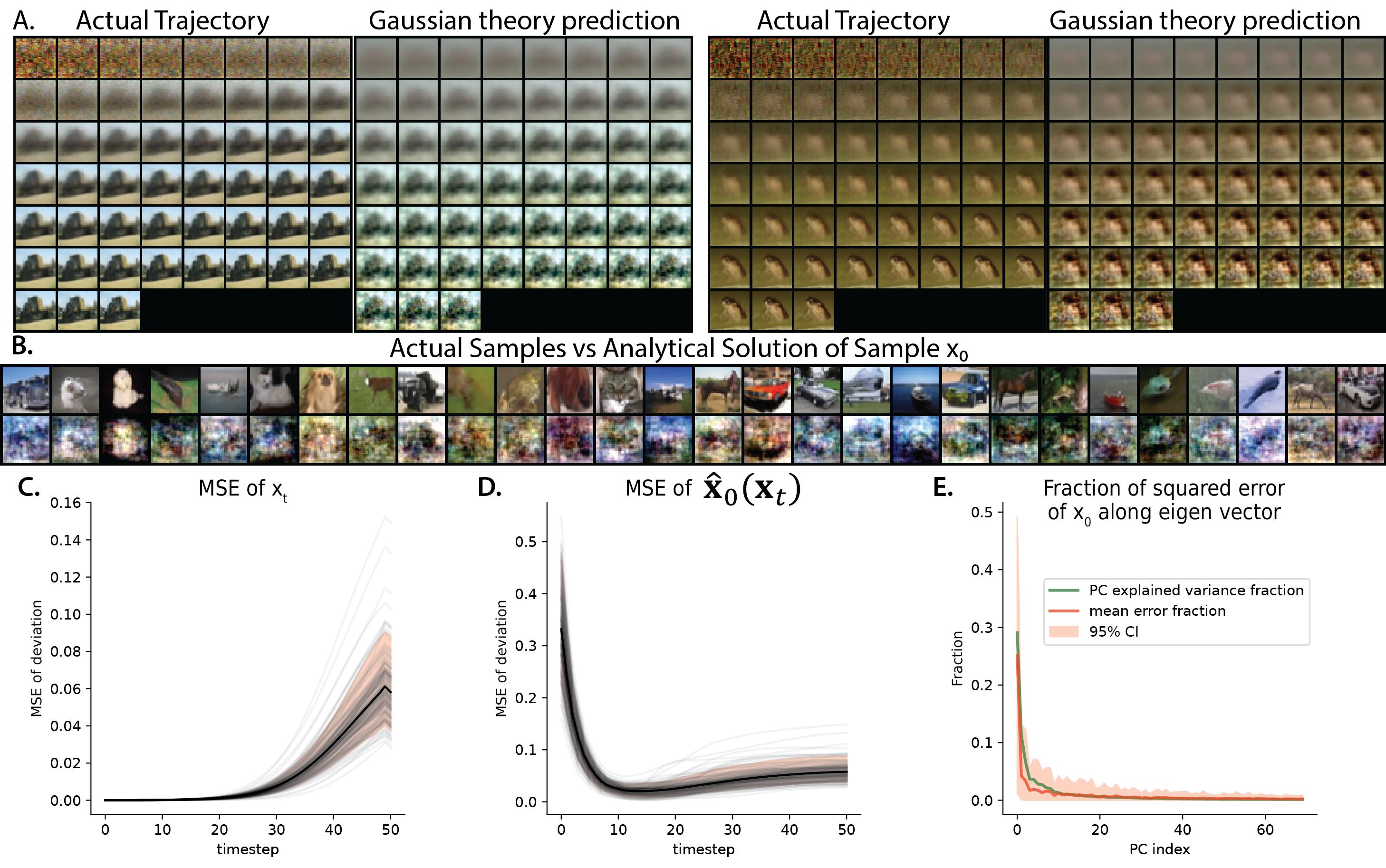}}
\caption{\textbf{Comparing analytical solution and actual diffusion process for the DDPM-CIFAR-10 model}. Same layout as Fig.\ref{fig:MNIST_theory_valid_supp}.}
\label{fig:CIFAR10_theory_valid_supp}
\end{center}
\vskip -0.3in
\end{figure}

\begin{figure}[!ht]
\vskip 0.2in
\begin{center}
\centerline{\includegraphics[width=0.7\columnwidth]{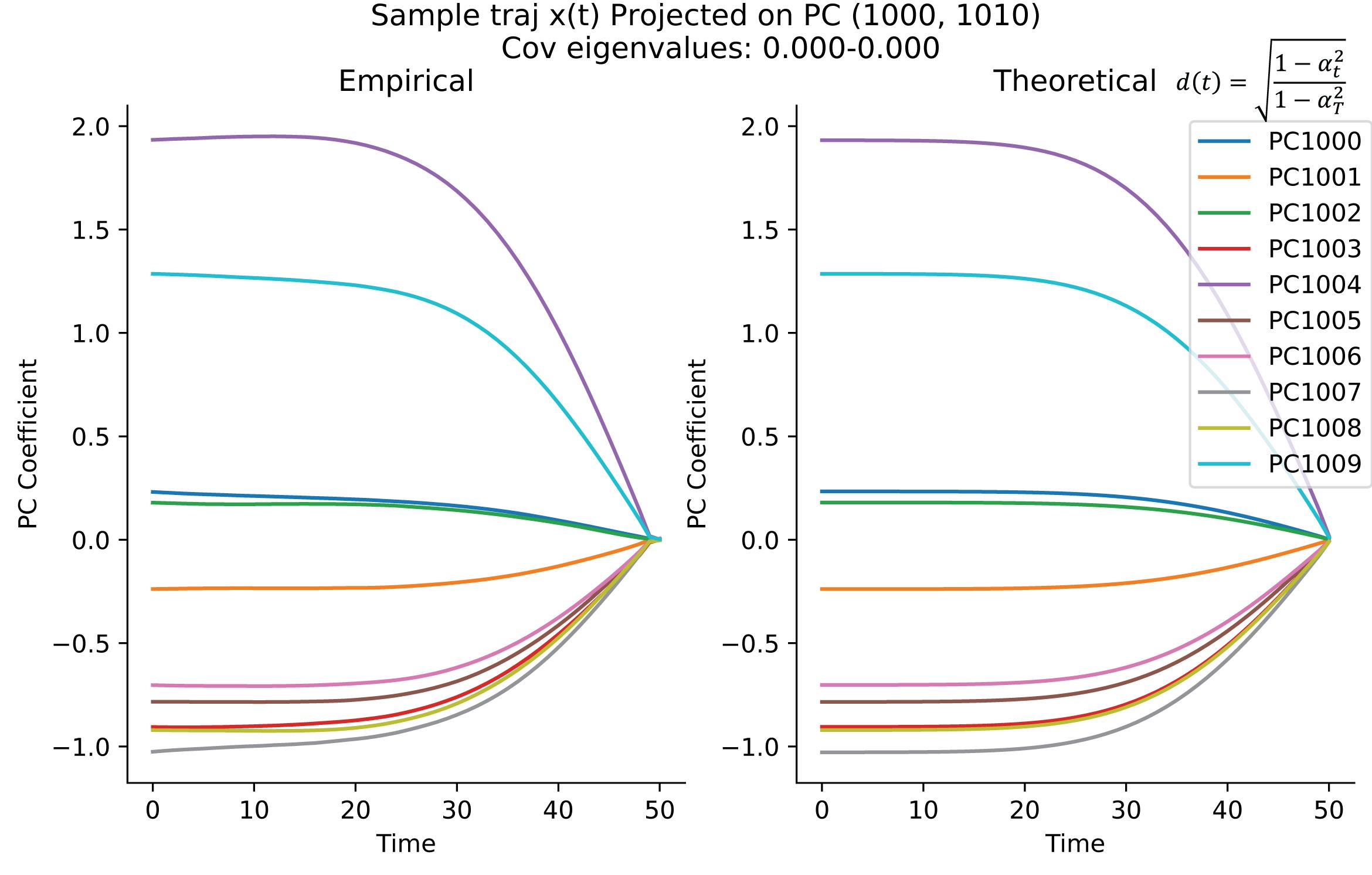}}
\caption{\textbf{Trajectory along off-manifold directions well-predicted by theory}. We computed the PC projection of actual trajectories and compared them to the theoretical prediction given the same initial value; they aligned well. DDPM-MNIST model.}
\label{fig:off-manifold-pred}
\end{center}
\vskip -0.3in
\end{figure}

\clearpage

\subsection{Validation of the single mode and Gaussian mixture theory on an MNIST model}
\begin{figure}[ht]
\vspace{-2pt}
\begin{center}
\centerline{\includegraphics[width=1.0\columnwidth]{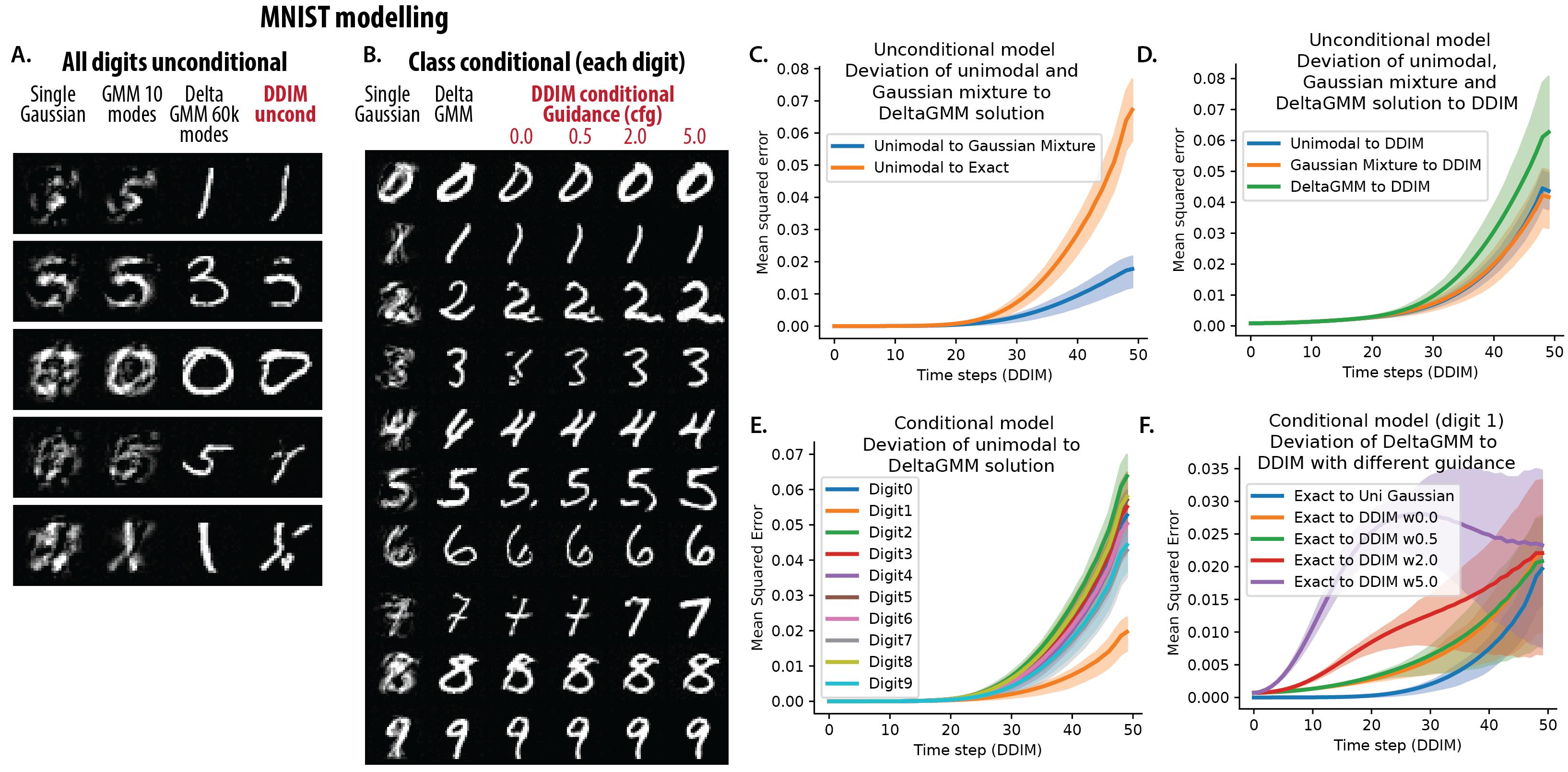}}\vspace{-10pt}
\caption{\textbf{Comparing the predictions of the Gaussian model, a 10-mode Gaussian mixture model, and the exact (delta function mixture) score function to real MNIST reverse diffusion trajectories}. \textbf{A.} True $\hat{\mathbf{x}}_0(\mathbf{x}_t)$ and $\hat{\mathbf{x}}_0(\mathbf{x}_t)$ predicted by various score models, all using the same initial conditions. \textbf{B.} Collections of actual and theory-predicted $\mathbf{x}_0$. \textbf{C.} Mean squared error between the actual trajectory $\mathbf{x}_t$ and trajectory predicted by each model. \textbf{D.} Fraction of squared error as a function of PC, between the actual and predicted $\mathbf{x}_0$.}
\label{fig:MNIST_gmm_theory_valid}
\end{center}
\vskip -0.45in
\end{figure}


\clearpage

\subsection{Acceleration results with EDM model and Heun's sampler}\label{apd:EDM_heun_accelr}

\begin{figure*}[!ht]
\vspace{-2pt}
\centering
\includegraphics[width=1.0\textwidth]{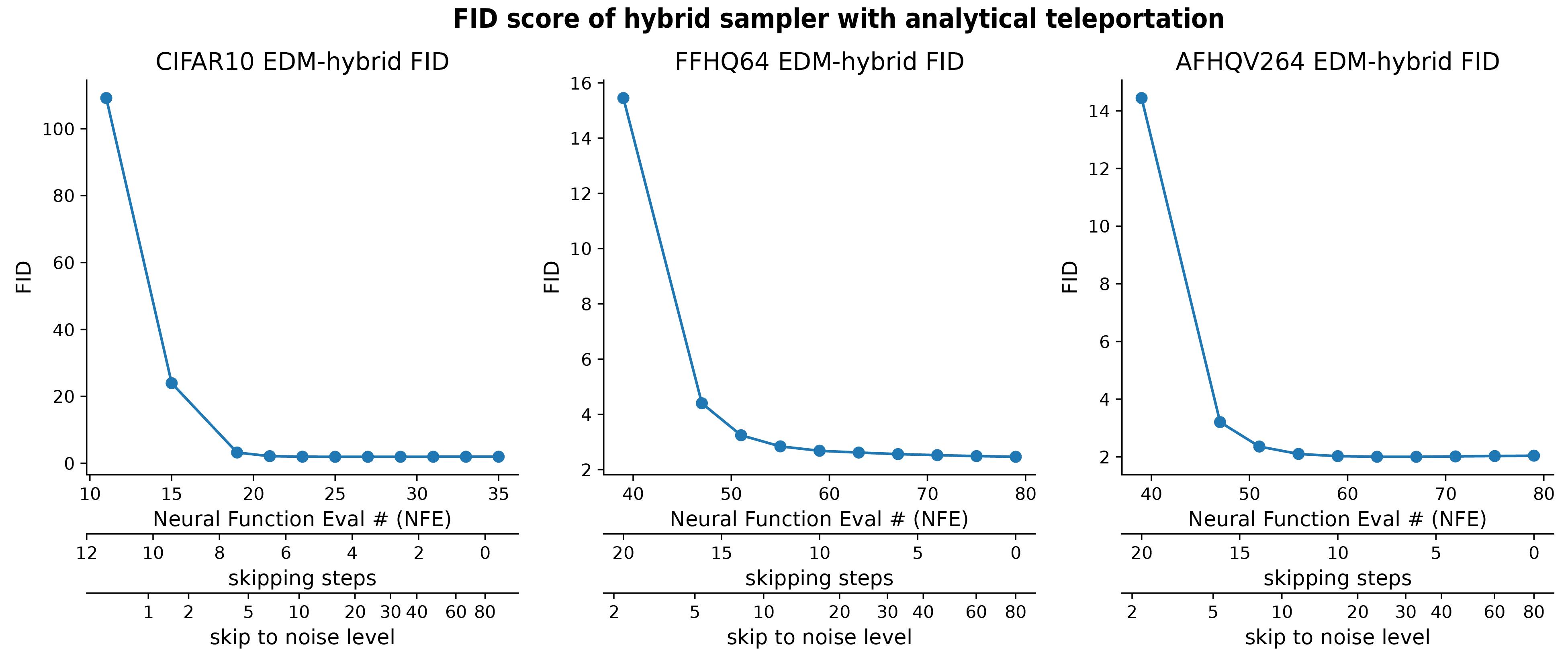}\vspace{-5pt}
\caption{\textbf{Image quality as a function of skipping steps for hybrid sampling approach.} Note the main x-axes are the number of Neural Function Evaluation (NFE); the secondary x-axes are the number of skipping steps from the Heun sampler; the tertiary-axes are the time or noise level $\sigma_{skip}$ at which we evaluate the Gaussian solution.  
See Tab.\ref{tab:cifar10_fid},\ref{tab:ffhq64_fid},\ref{tab:afhqv264_fid} for numbers.}
\vspace{-2pt}
\label{fig:hybrid_fid_fullresult}
\end{figure*}

\begin{table}[!ht]
\centering
\caption{FFHQ64 FID with analytical teleportation}
\label{tab:ffhq64_fid}
\begin{tabular}{rrrr}
\toprule
 Nskip &   NFE &   time/noise scale &    FID \\
\midrule
     0 &    79 &             80.0 &  2.464 \\
     2 &    75 &             60.1 &  2.489 \\
     4 &    71 &             44.6 &  2.523 \\
     6 &    67 &             32.7 &  2.561 \\
     8 &    63 &             23.6 &  2.617 \\
    10 &    59 &             16.8 &  2.681 \\
    12 &    55 &             11.7 &  2.841 \\
    14 &    51 &              8.0 &  3.243 \\
    16 &    47 &              5.4 &  4.402 \\
    20 &    39 &              2.2 & 15.451 \\
\bottomrule
\end{tabular}
\end{table}

\begin{table}[!ht]
\centering
\caption{AFHQV264 FID with analytical teleportation}
\label{tab:afhqv264_fid}
\begin{tabular}{rrrr}
\toprule
 Nskip &   NFE &   time/noise scale &    FID \\
\midrule
     0 &    79 &             80.0 &  2.043 \\
     2 &    75 &             60.1 &  2.029 \\
     4 &    71 &             44.6 &  2.016 \\
     6 &    67 &             32.7 &  2.003 \\
     8 &    63 &             23.6 &  2.005 \\
    10 &    59 &             16.8 &  2.026 \\
    12 &    55 &             11.7 &  2.102 \\
    14 &    51 &              8.0 &  2.359 \\
    16 &    47 &              5.4 &  3.206 \\
    20 &    39 &              2.2 & 14.442 \\
\bottomrule
\end{tabular}
\end{table}

\begin{table}[!ht]
\centering
\caption{CIFAR10 FID with analytical teleportation}
\label{tab:cifar10_fid}
\begin{tabular}{rrrr}
\toprule
 Nskip &   NFE &   time/noise scale &     FID \\
\midrule
     0 &    35 &             80.0 &   1.958 \\
     1 &    33 &             57.6 &   1.955 \\
     2 &    31 &             40.8 &   1.949 \\
     3 &    29 &             28.4 &   1.940 \\
     4 &    27 &             19.4 &   1.932 \\
     5 &    25 &             12.9 &   1.934 \\
     6 &    23 &              8.4 &   1.963 \\
     7 &    21 &              5.3 &   2.123 \\
     8 &    19 &              3.3 &   3.213 \\
    10 &    15 &              1.1 &  23.947 \\
    12 &    11 &              0.3 & 109.178 \\
\bottomrule
\end{tabular}
\end{table}

\clearpage

\subsection{Principal dimensions of Stable Diffusion sampling trajectories}\label{sec:SD_PCA_vis}

\begin{figure}[!ht]
\vskip 0.2in
\begin{center}
\centerline{\includegraphics[width=0.85\columnwidth]{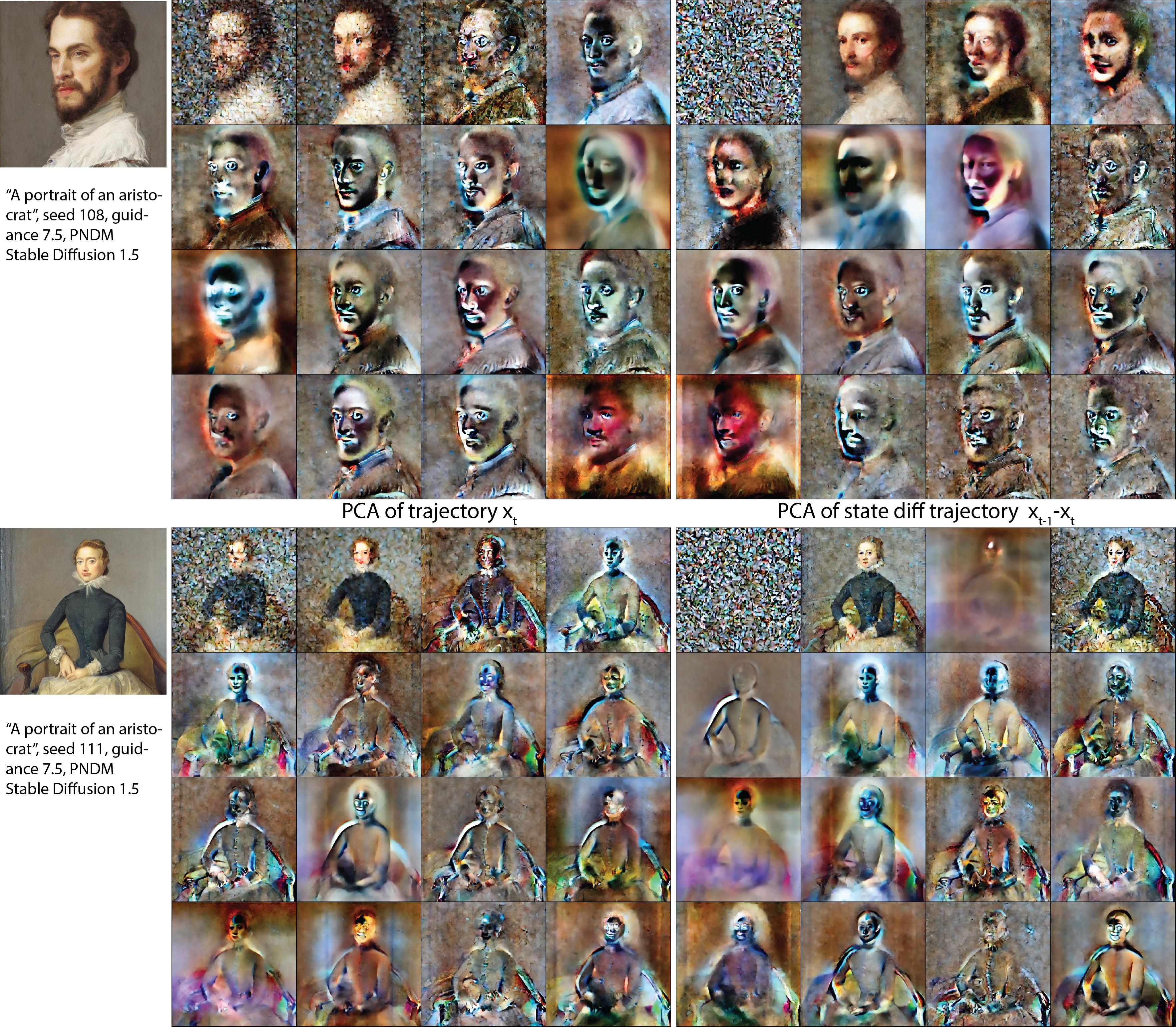}}
\caption{\textbf{Visualizing PCs of Stable Diffusion trajectories}. We computed the principal components of the trajectory $\vec{x}_t$ and the trajectory difference $\vec{x}_{t-1}-\vec{x}_t$, and visualized the scaled version of PC vectors through the decoder. We can see that they represent an interpretable vector space around the actual sample $\vec{x}_0$.}
\label{fig:SD_PCA_vis}
\end{center}
\vskip -0.3in
\end{figure}

\clearpage

\subsection{Additional perturbation experiments}

\begin{figure}[ht!]
\begin{center}
\centerline{\includegraphics[width=\columnwidth]{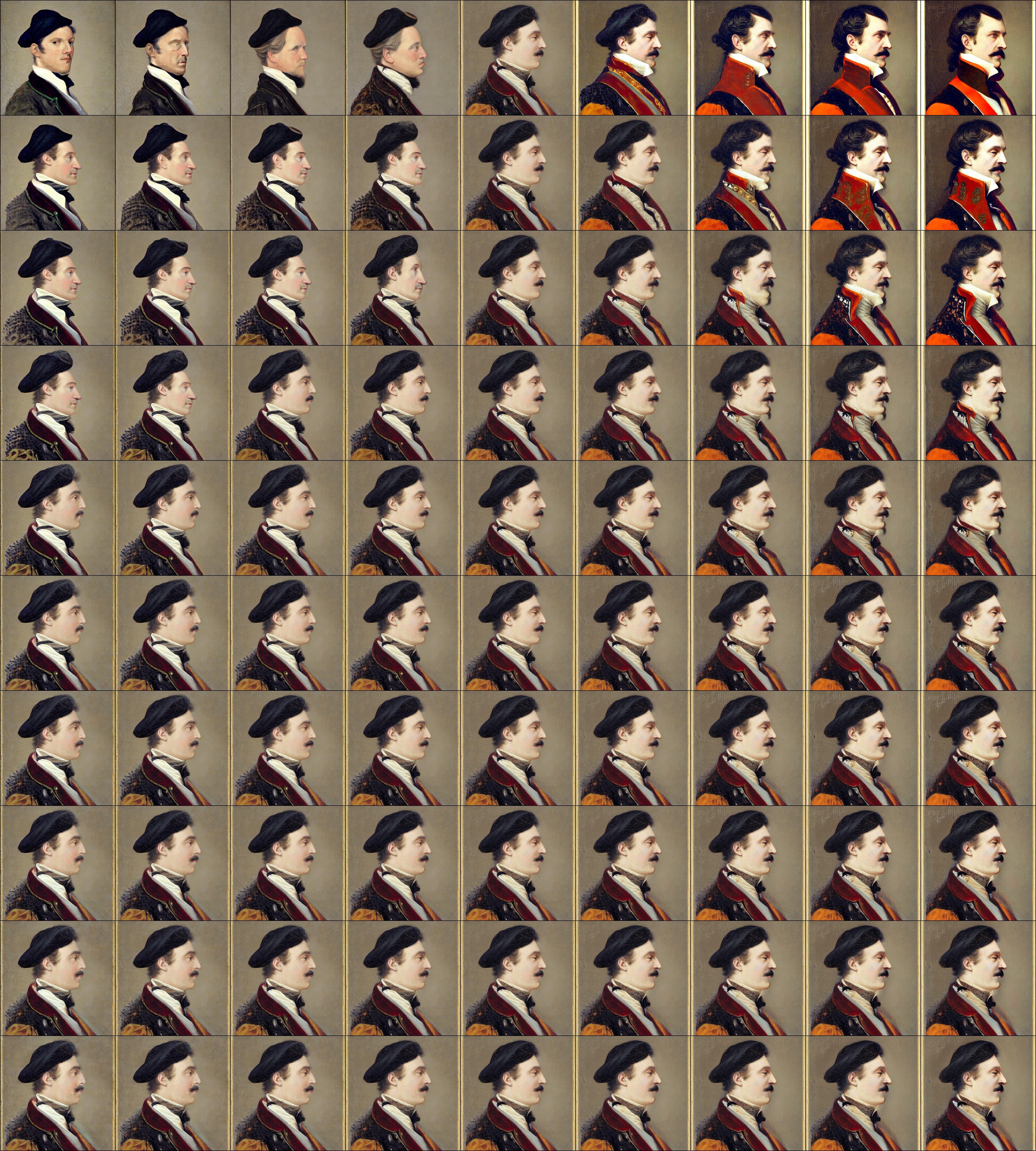}}\label{fig:SD_perturb_PC06}
\caption{\textbf{Example perturbation experiment result.} Different rows represent different perturbation times, from top to bottom $5,10,...50$. Different columns represent different perturbation scales, left to right: negative to positive, $-20,-15,...15,20$. Perturbations are along PC6 of U-Net output. Stable Diffusion, PNDM sampler, seed 100. Prompt: ``a portrait of an aristocrat''.}
\end{center}
\end{figure}

\begin{figure}[ht!]
\begin{center}
\centerline{\includegraphics[width=\columnwidth]{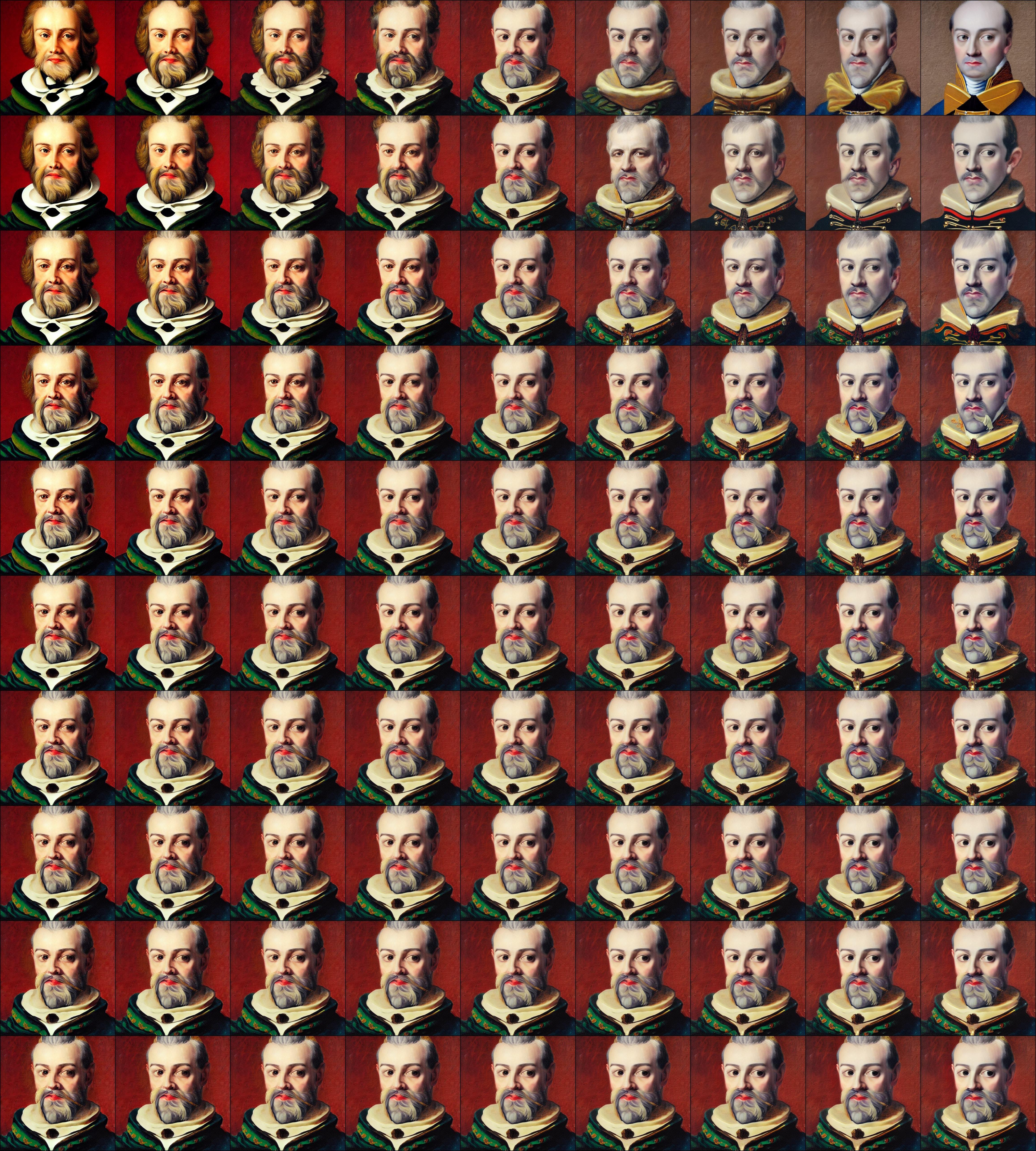}}
\caption{\textbf{Example perturbation experiment result.}  Different rows represent different perturbation times, from top to bottom $5,10,...50$. Different columns represent different perturbation scales, left to right: negative to positive, $-20,-15,...15,20$. Perturbations are along PC5 of U-Net output. Stable Diffusion, PNDM sampler, seed 101. Prompt: ``a portrait of an aristocrat''.}
\label{fig:SD_perturb_PC05}
\end{center}
\end{figure}

\begin{figure}[ht!]
\begin{center}
\centerline{\includegraphics[width=\columnwidth]{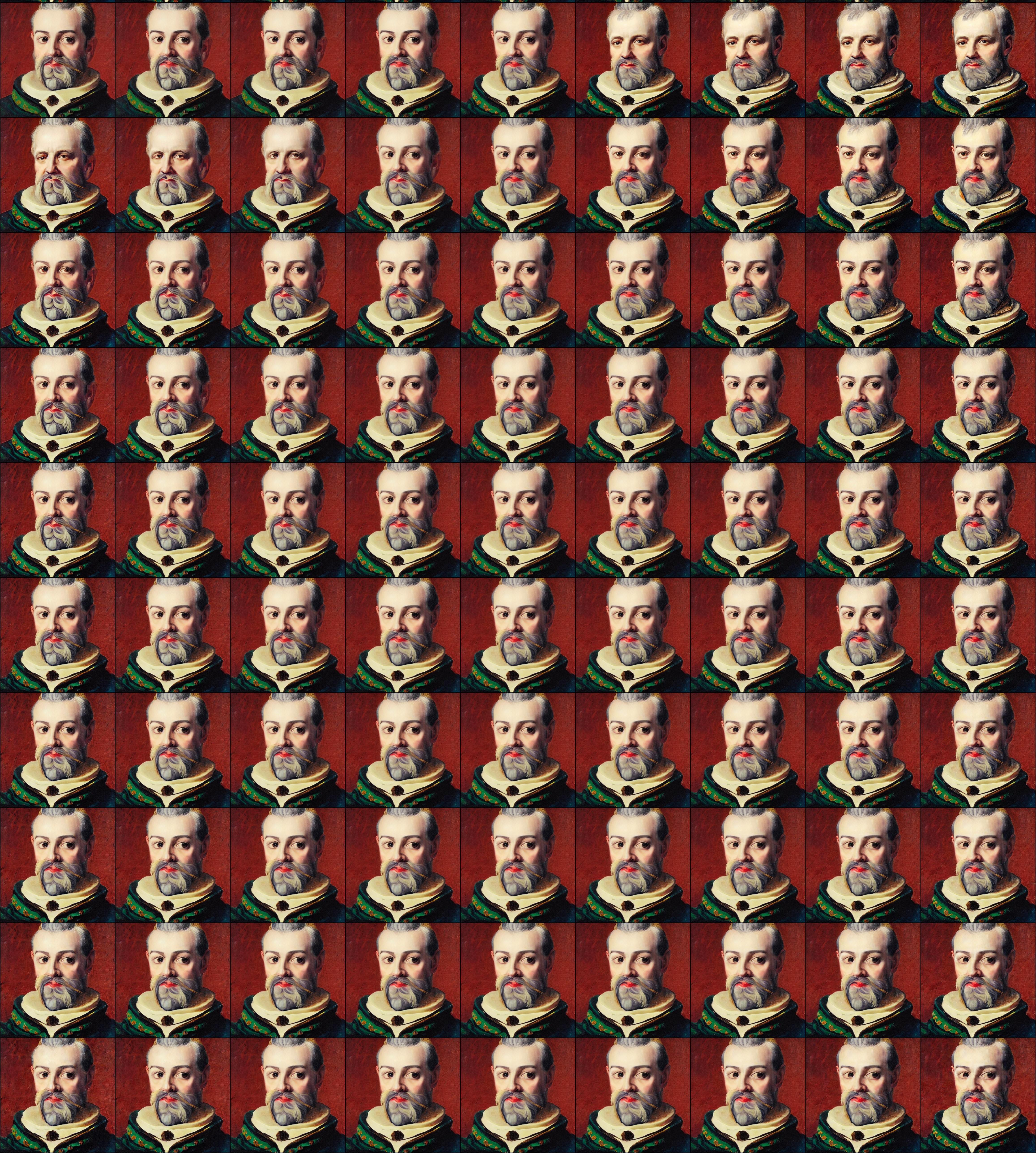}}
\caption{\textbf{Example perturbation experiment result.} Different rows represent different perturbation times, from top to bottom $5,10,...50$. Different columns represent different perturbation scales, left to right: negative to positive, $-20,-15,...15,20$. Perturbations are along a random noise direction, which is less effective than the previous PC perturbation. Stable Diffusion, PNDM sampler, seed 101. Prompt: ``a portrait of an aristocrat''.}
\label{fig:SD_perturb_RND009}
\end{center}
\end{figure}

\begin{figure}[ht!]
\begin{center}
\centerline{\includegraphics[width=\columnwidth]{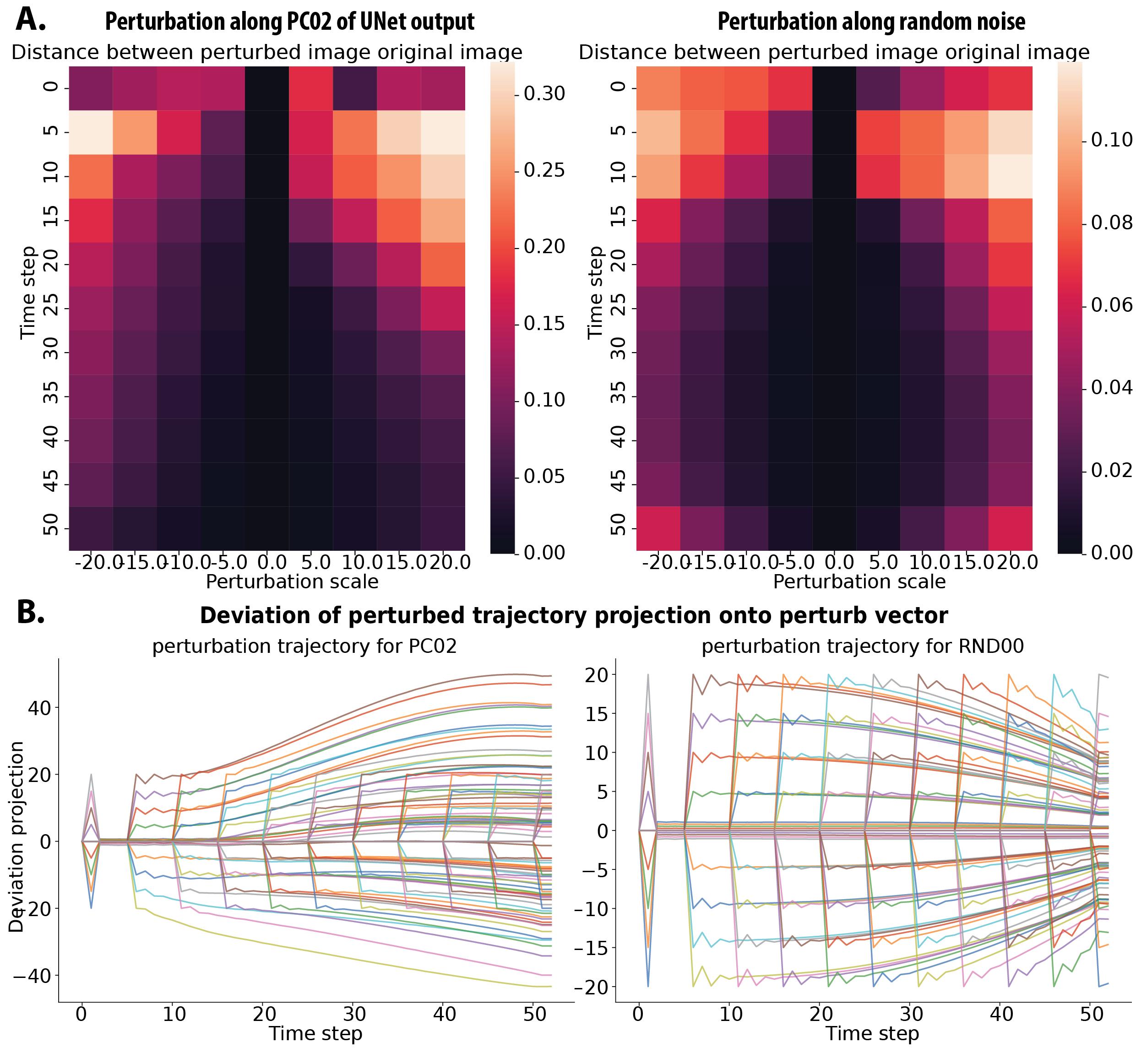}}
\caption{\textbf{Geometry of trajectory perturbation}. \textbf{A.} Quantification of sample deviation due to a perturbation at time $t'$ and strength $K$, heatmap encodes the perceptual image similarity per LPIPS. Left panel: perturbation along PC02, Right panel: perturbation along a random vector. The perturbation along PC02 is much more effective than the random pattern in affecting the sample. \textbf{B.} The underlying geometry, projecting the difference between the perturbed trajectory and the original trajectory onto the unit perturbation vector. As we predicted, the perturbation along the signal manifold gets amplified, while the effect of a random perturbation decays over time. (\ref{eq:y_perturb_formula}) }
\label{fig:SD_perturb_summary}
\end{center}
\end{figure}

\clearpage

\subsection{Impact of design choices on the geometry of diffusion}\label{sec:effectCfgSampler}

The field of diffusion generative modeling has greatly advanced since the DDIM paper \cite{song2020DDIM}. Here, we consider how making slightly different design choices (e.g. using different samplers) affects the geometry and dynamics of sampling trajectories

\subsubsection{Differential equation solver} \label{sec:effectSampler}

\begin{figure*}[ht]
\begin{center}
\centerline{\includegraphics[width=0.9\columnwidth]{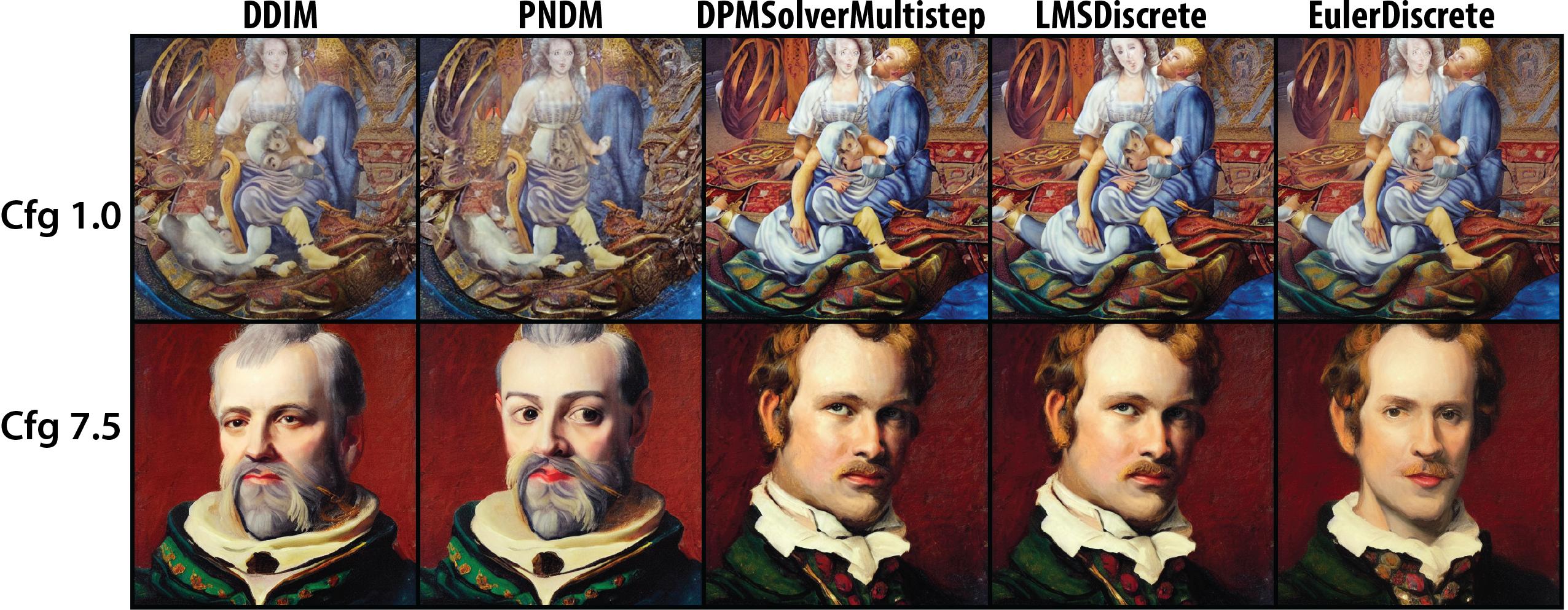}}
\label{fig:guidance_effect_sample}
\caption{\textbf{Effect of classifier-free guidance (cfg) strength and diffusion sampler on the sample}. }
\end{center}
\vskip -0.3in
\end{figure*}

Stable Diffusion by default uses the deterministic PNDM sampler, but one can use other solvers. Because the choice of solver slightly modifies the size and direction of individual time steps, changing the solver is akin to doing an early perturbation on the trajectory. Consistent with this expectation, we observe a variety of effects on image generation with a fixed random seed and different solvers: sometimes there is effectively no difference (e.g. DDIM and PNDM), sometimes there is a slight difference, and sometimes there is a major difference (e.g. PNDM vs LMSDiscrete Solver) (Fig.\ref{fig:guidance_effect_sample}). As for trajectory geometry, usually, DDIM, PNDM, and DPMSolverMultistep create rotation-like 2D trajectories well-predicted by our theory, while LMSDiscrete and EulerDiscrete create more linear 1D trajectories. 



\subsubsection{Classifier-free guidance strength}\label{sec:effectCFGuidance}
\begin{figure*}[ht]
\begin{center}
\centerline{\includegraphics[width=0.9\columnwidth]{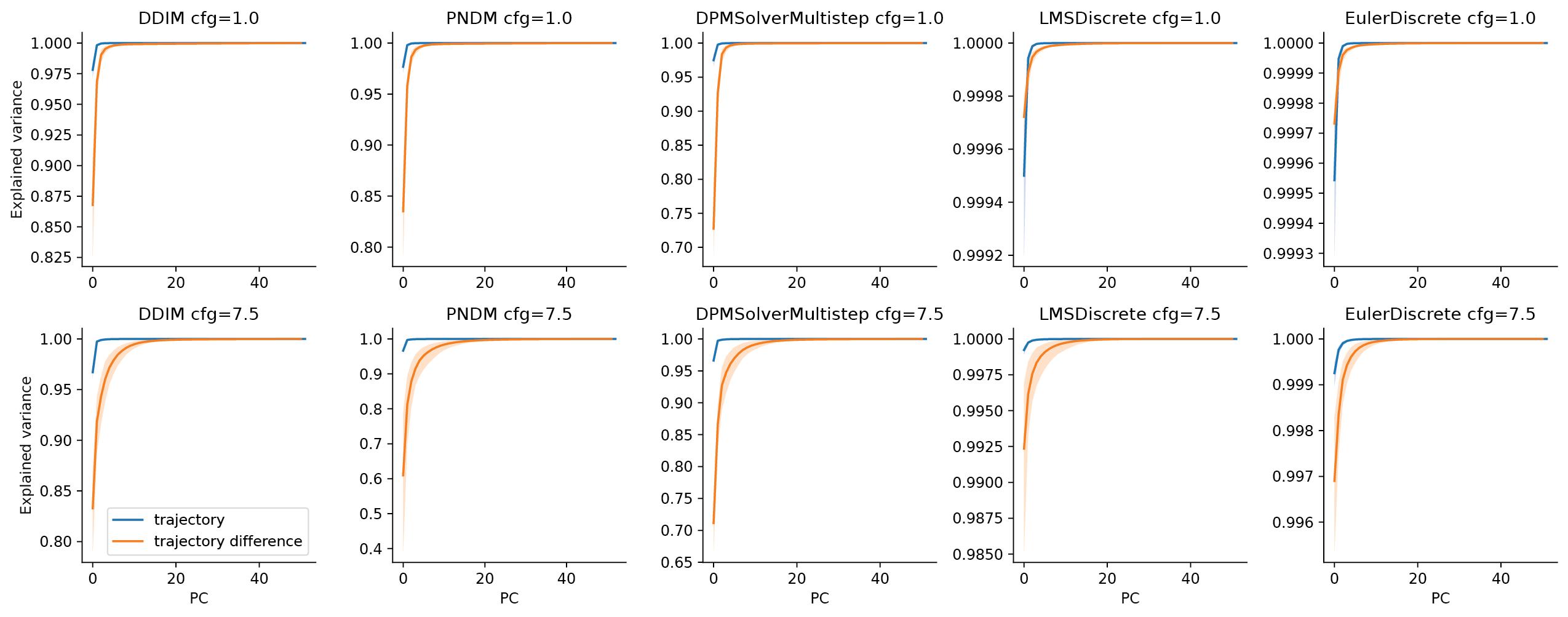}}
\centerline{\includegraphics[width=0.9\columnwidth]{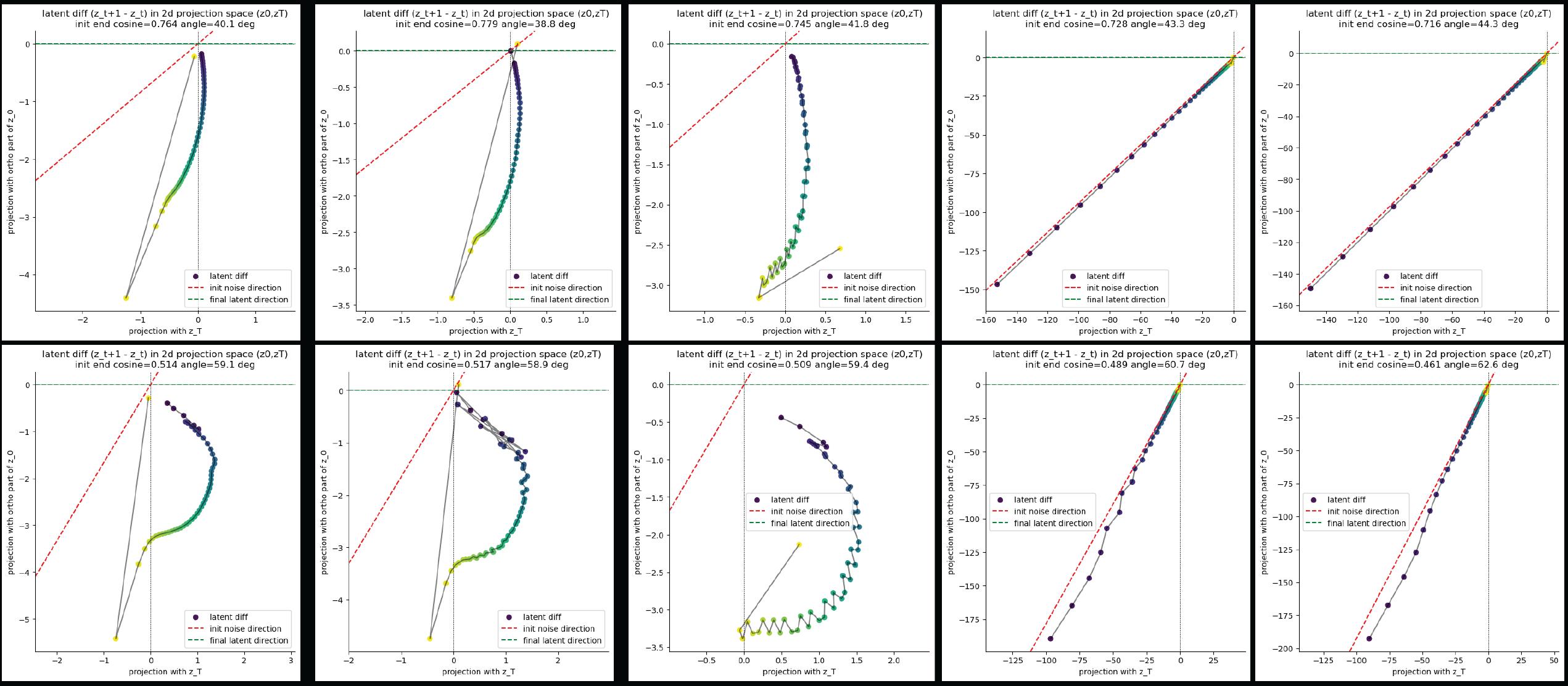}}
\centerline{\includegraphics[width=0.9\columnwidth]{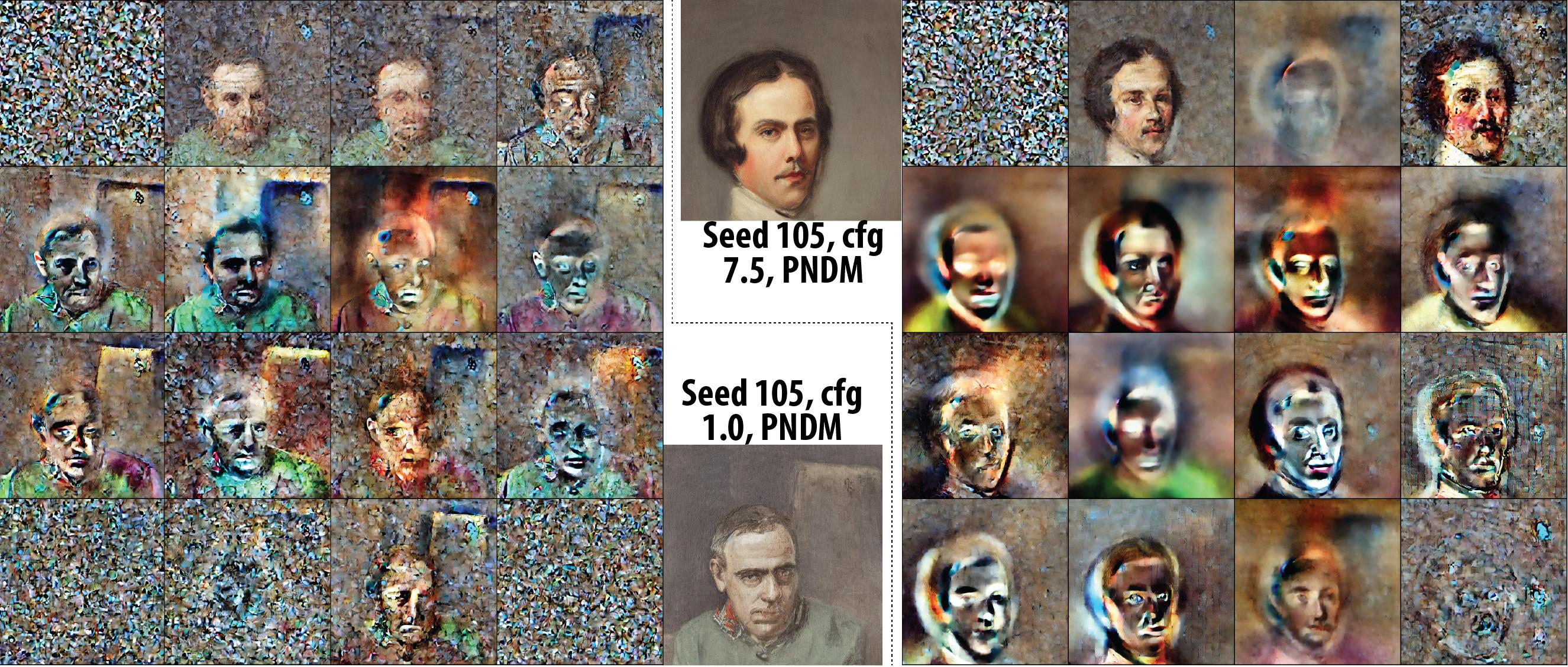}}
\caption{\textbf{Effect of classifier-free guidance strength and sampler on trajectory geometry}. 
\textbf{Top panel}: Dimensionality of trajectory $\mathbf{x}_t$ and state difference $\Delta \mathbf{x}_t$, measured by explained variance of PCs. Higher guidance induces higher dimensionality in $\mathbf{x}_t$ and $\Delta \mathbf{x}_t$.
\textbf{Middle panel}: Trajectory difference $\Delta \mathbf{x}_t$ projected onto the $\mathbf{x}_0,\mathbf{x}_T$ plane. Higher guidance induced oscillation in the search trajectory, esp. for PNDM sampler. 
\textbf{Bottom panel}: Comparing the top 16 PCs of state difference $\Delta \mathbf{x}_t$, for high guidance (7.5) versus low guidance (1.0) trajectory from the same noise seed. The samples are shown in the middle. Trajectories sampled with higher guidance have more `on-manifold' PC dimensions.}
\label{fig:guidance_effect}
\end{center}
\vskip -0.3in
\end{figure*}

Classifier-free guidance \cite{ho2022classifierFreeGuidance} has been widely used in conditional diffusion models as a method to generate samples highly aligned with the conditional signal (e.g. prompt). We examined the effect of the strength of classifier-free guidance on the geometry of trajectories. 
We found that, generally, a higher guidance scale generates trajectories $\vec{x}_t$ and trajectory differences $\Delta \vec{x}_t$ with higher dimensionality (Fig.\ref{fig:guidance_effect} top). Furthermore, when visualizing the top PC vectors, a larger number of interpretable PC dimensions can be found for a higher guidance scale (Fig. \ref{fig:guidance_effect} bottom). 
We also observed that with a smaller guidance value, the trajectory is usually smooth; with a higher guidance value, it induced strong oscillatory movement in the trajectory at the early phase, when combined with certain higher-order schedulers e.g. the default solver PNDM \cite{liu2022PNDM} (Fig. \ref{fig:guidance_effect} middle). 
This effect could be a feature for the sampler to explore the landscape more. It could also be an artifact, which could be fixed by modifying the sampler.

\clearpage

\section{Endpoint estimate trajectory examples}

\begin{figure}[ht!]
\begin{center}
\centerline{\includegraphics[width=0.9\columnwidth]{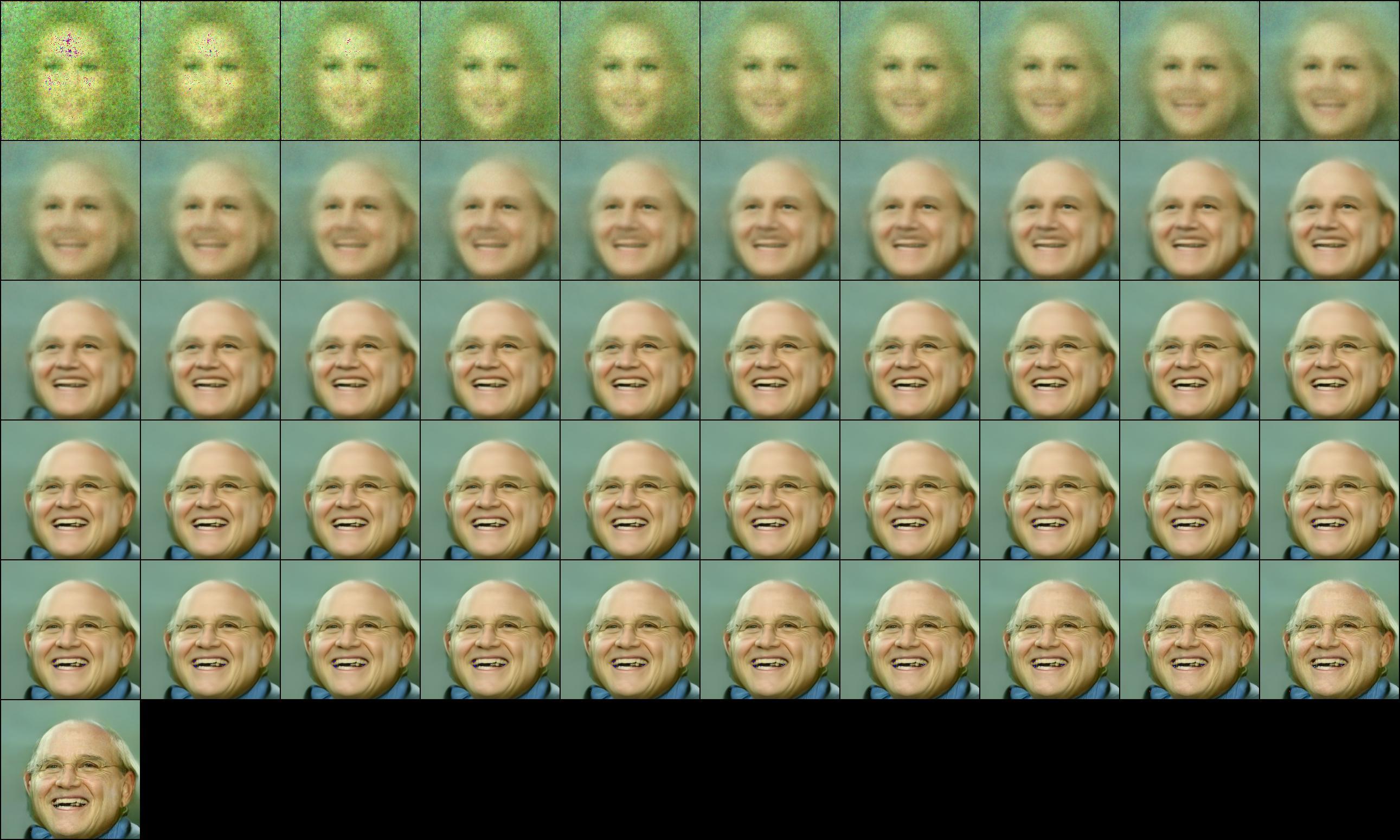}}
\caption{\textbf{Example endpoint estimate trajectory.} CelebA-HQ, DDIM sampler, seed 129.}
\end{center}
\end{figure}

\begin{figure}[ht!]
\begin{center}
\centerline{\includegraphics[width=0.9\columnwidth]{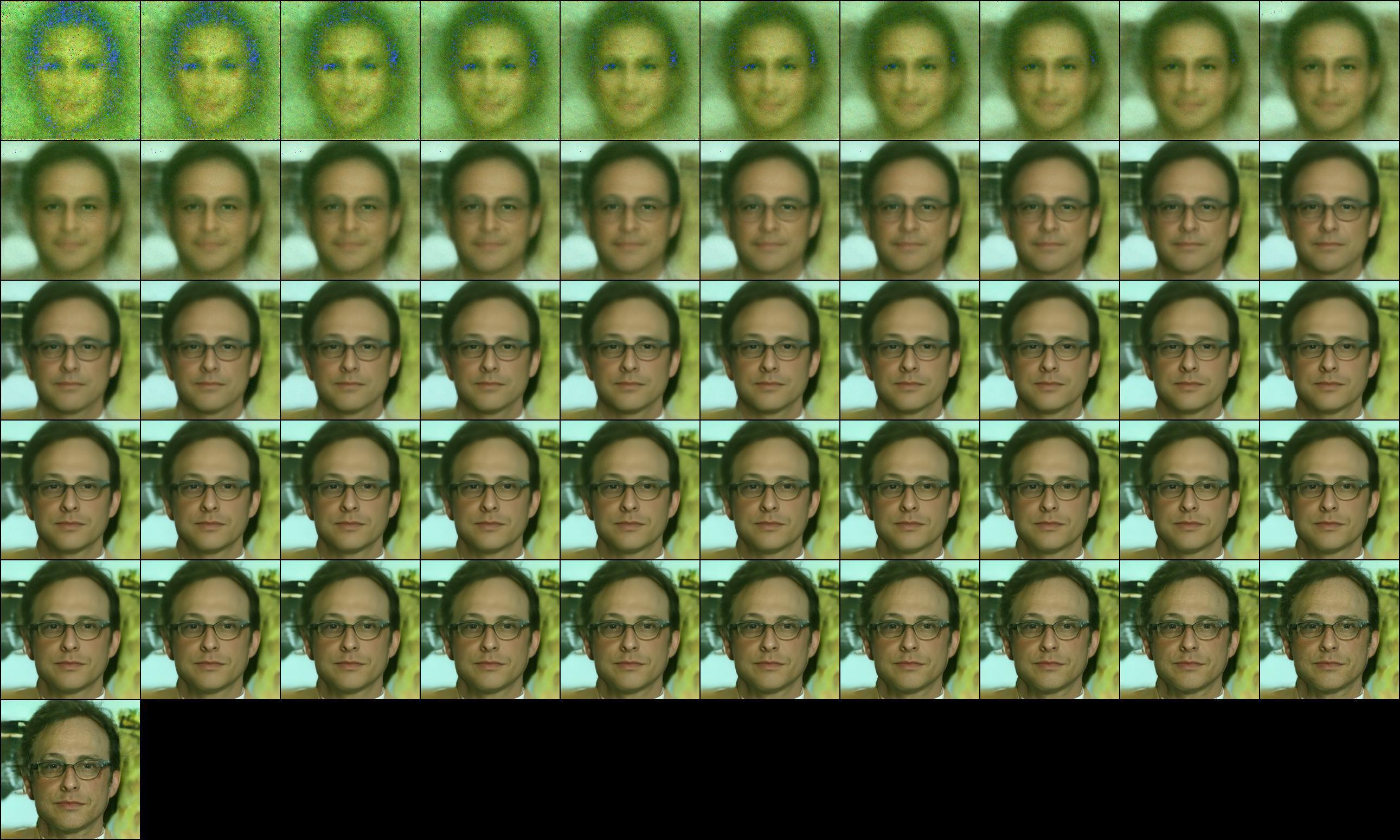}}
\caption{\textbf{Example endpoint estimate trajectory.} CelebA-HQ, DDIM sampler, seed 152.}
\end{center}
\end{figure}

\clearpage

\begin{figure}[ht!]
\begin{center}
\centerline{\includegraphics[width=0.9\columnwidth]{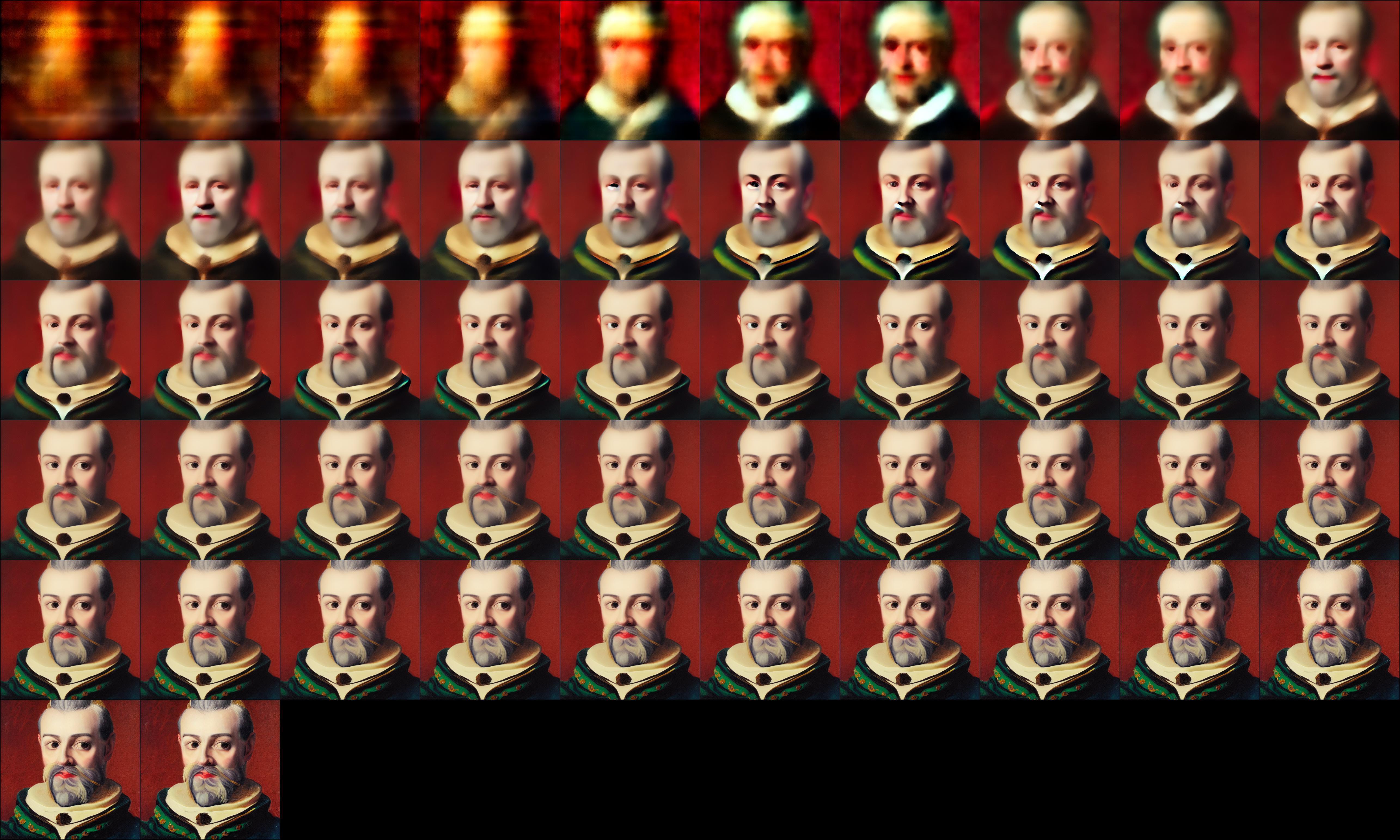}}
\caption{\textbf{Example endpoint estimate trajectory.} Stable Diffusion, PNDM sampler, seed 101. Prompt: ``a portrait of an aristocrat''.}
\end{center}
\end{figure}

\begin{figure}[ht!]
\begin{center}
\centerline{\includegraphics[width=0.9\columnwidth]{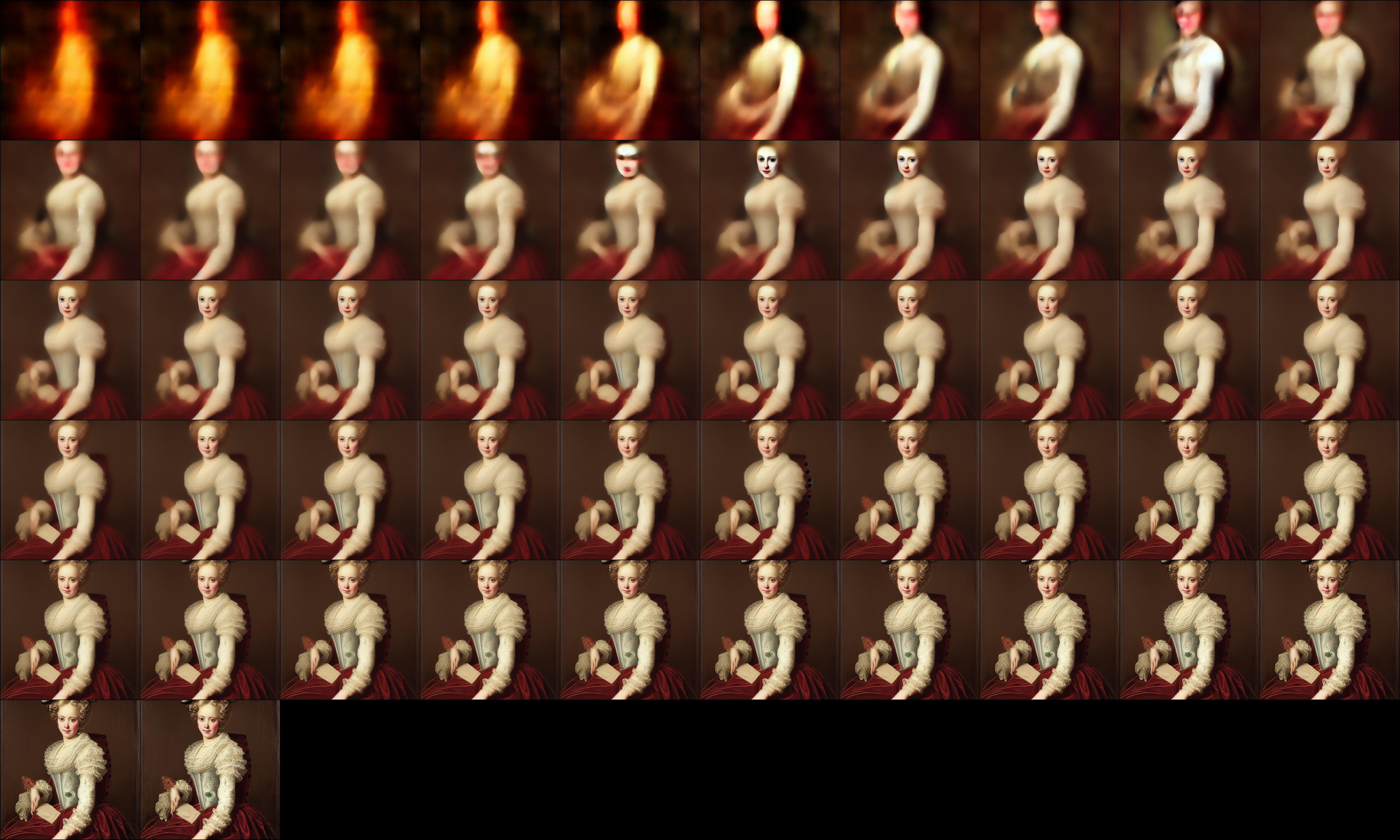}}
\caption{\textbf{Example endpoint estimate trajectory.} Stable Diffusion, PNDM sampler, seed 107. Prompt: ``a portrait of an aristocrat''.}
\end{center}
\end{figure}

\clearpage

\section{Notation correspondence}
\label{apd:notation}
Diffusion models usually have forward processes whose conditional probabilities are
\begin{equation}
p(\mathbf{x}_t | \mathbf{x}_0) = \mathcal{N}( A_t \mathbf{x}_0, B_t \mathbf{I})
\end{equation}
for all $t \in [0, T]$. In the limit of small time steps, the transition probability distribution can be captured by the SDE
\begin{equation}
\dot{\mathbf{x}}_t = - C_t \mathbf{x}_t + D_t \mathbf{\eta}(t)
\end{equation}
where $\mathbf{\eta}(t)$ is a vector of independent Gaussian white noise terms.

Papers discussing these models may use slightly different notation. In the table below, we briefly indicate how various choices of notation correspond to one another. To make comparing discrete and continuous models easier, we assume the time step size is $\Delta t = 1$.

\begin{table}[!ht]
\caption{\textbf{Comparison of notation for diffusion model parameters}.}
\label{notation-table}
\begin{center}
\begin{tiny}
\begin{sc}
\begin{tabular}{llcccc}
\toprule
Paper & Citation & $A_t$ & $B_t$ & $C_t$ & $D_t$ \\
\midrule
DDPM  & \cite{ho2020DDPM} & $\sqrt{\bar{\alpha}_t}$ &  $1 - \bar{\alpha}_t$ & $1 - \sqrt{1 - \beta_t}$ & $\sqrt{\beta_t}$ \\
DDIM  & \cite{song2020DDIM} & $\sqrt{\alpha_t}$ &  $1 - \alpha_t$ & $1 - \sqrt{\alpha_t/\alpha_{t-1}}$ & $\sqrt{1 - \alpha_t/\alpha_{t-1}}$ \\
Stable Diff.  & \cite{rombach2022latentdiff} & $\alpha_t$ &  $\sigma_t^2$ & $1 - \alpha_t/\alpha_{t-1}$ & $\sqrt{\sigma_t^2 - (\alpha_t/\alpha_{t-1})^2 \sigma_{t-1}^2}$ \\
VP SDE & \cite{song2021scorebased} & $\exp\left[ - \frac{1}{2} \int_0^t \beta(s) ds \right]$ & $1 - \exp\left[ - \int_0^t \beta(s) ds \right]$ & $\beta(t)/2$  & $\sqrt{\beta(t)}$ \\
Ours &  & $\alpha_t$ & $\sigma_t^2$ & $\beta(t)$ & $g(t)$ \\
\bottomrule
\end{tabular}
\end{sc}
\end{tiny}
\end{center}
\vskip -0.1in
\end{table}

In the popular huggingface \texttt{diffusers} library implementation of diffusion models, the function \texttt{alphas\_cumprod} corresponds to our $\alpha_t^2$.


\clearpage
\section{Details of the numerical simulation of solution}

To sharpen our intuition about our analytical results (especially Eq. \ref{eq:xt_solu_psi_def} and \ref{eq:xhat_explicit}), we used a common $\alpha_t$ schedule (see Appendix \ref{apd:model_details}) and plotted scaled projection coefficients $\bar{c}_k(t):=c_k(t)/c_k(T)$ and $d(t)$ for different $\lambda_k$, i.e. the variance of the distribution in the direction $\mathbf{u}_k$. The projection coefficients of $\mathbf{x}_t$ along large variance directions increase at first, while the coefficients along the low variance directions remain the same or shrink (Fig. \ref{fig:analytical_curve}A).

The derivative $\dot{\mathbf{x}}_t$, which indicates the direction of particle movement, is initially dominated by contributions from the large variance dimensions. Close to the end, it features comparable but opposite-sign contributions from the high variance dimensions and effectively off-manifold (i.e. noise) directions (Fig. \ref{fig:analytical_curve}B). This explains the observation that state differences $\mathbf{x}_{t-k} - \mathbf{x}_t$ look interpretable at first, and later look like noise-contaminated images (Fig. \ref{fig:initial_obs}A, middle row).

The projection coefficients of the mean-adjusted endpoint estimate $\hat {\mathbf{x}}_0(\mathbf{x}_t) - \vec{\mu}$, when normalized by their standard deviation $\sqrt{\lambda_k}$, look sigmoidal (Fig. \ref{fig:analytical_curve}C). The coefficients converge to their final value earlier along large variance dimensions, and later along small variance dimensions. The time derivatives of these coefficients are bump-like, and peak earlier along high variance dimensions. In plain language, high variance features are added to the endpoint estimate first, and low variance features are added later. This partly explains our earlier observations about the order of feature emergence (Fig. \ref{fig:initial_obs}A, bottom row).

\section{Fr\'{e}chet inception distance score for assessing generated image quality}
\label{apd:fid_method}

The Fr\'{e}chet inception distance (FID) score \cite{FIDpaper} provides one way to assess the quality of images produced by generative models, including GANs and diffusion models. In this paper, we have used it to assess the effect of using our Gaussian solution to skip some number of initial reverse diffusion steps on generated images (e.g. on a model of CIFAR-10; see Fig. \ref{fig:CIFAR_theory_valid}); we found that image quality begins to seriously suffer when the number of skipped steps becomes somewhat larger than $30$, although the exact point varies for models trained on different data sets.

The idea behind the FID score is the following. We would like the distribution of generated images $p(\cdot)$ to be similar to some distribution of images $p_w(\cdot)$ in the world. Hence, one way to assess image quality is via a measure that quantifies the difference between these distributions---or between the distributions of suitably transformed images. The Fr\'{e}chet distance is one such measure, but it is difficult to compute in general.

We will need two facts. First, the Fr\'{e}chet distance can be computed analytically when the distributions being compared are Gaussian. Second, it is possible to apply transformations to images that make their distribution approximately Gaussian, and in particular, deep networks that perform tasks like object recognition well are known to do this. 

The algorithm for implementing the FID score, then, is the following. (1) Transform a set of real and generated images using some nonlinear function. (2) Fit Gaussians to both distributions. (This can be done simply by computing the mean and covariance of each distribution.) (3) Compute the Fr\'{e}chet distance between the two Gaussians. The specific formula used to compare Gaussians $\mathcal{N}(\vec{\mu}_1, \vec{\Sigma}_1)$ and $\mathcal{N}(\vec{\mu}_2, \vec{\Sigma}_2)$ is
\begin{equation}
d = \Vert \vec{\mu}_1 - \vec{\mu}_2 \Vert_2^2 + \text{Tr}\left[ \vec{\Sigma}_1 + \vec{\Sigma}_2 - 2 \left( \vec{\Sigma}_1^{1/2} \vec{\Sigma}_2 \vec{\Sigma}_1^{1/2} \right)^{1/2} \right] \ .
\end{equation}
Lower FID scores are interpreted as indicating that generated images are of higher quality. 

We computed it with the \texttt{fidelity} function from the \texttt{torch-fidelity} library. We also confirmed it with the 
\texttt{fid.py} script from the official repository of \cite{karras2022elucidatingDesignSp} (https://github.com/NVlabs/edm). 
In these implementations, images are transformed using the penultimate layer of the Inception V3 model \cite{inceptionV3} trained on ImageNet for object classification. For each sampling method, we sampled 50,000 images from the same initial states $\mathbf{x}_T$ generated by the random seeds, 1-50,000. The FID score is computed by comparing these 50,000 samples and the 50,000 training set images of CIFAR-10. 




\clearpage

\section{Derivation of exact solution to Gaussian score model} \label{apd:deriv_gaussian_model}

In this section, we derive the analytic solution $\mathbf{x}_t$ for the reverse diffusion trajectory of the Gaussian score model. As in Sec. \ref{sec:theory} of the main text, we will assume throughout that the score function corresponds to a Gaussian image distribution whose mean is $\vec{\mu}$ and covariance matrix is $\vec{\Sigma}$. We will also assume, given that images are usually thought of as residing on a low-dimensional manifold within pixel space, that the rank $r$ of the covariance matrix may be less than the dimensionality $D$ of state space. 

Let $\mathbf{\Sigma}=\mathbf{U}\mathbf{\Lambda} \mathbf{U}^T$ be the eigendecomposition or compact SVD of the covariance matrix, where $\mathbf{U}$ is a $D \times r$ semi-orthogonal matrix whose columns are normalized (i.e. $\mathbf{U}^T\mathbf{U}=\mathbf{I}_r$), and $\mathbf{\Lambda}$ is the $r \times r$ diagonal eigenvalue matrix. Denote the $k$th column of $\mathbf{U}$ by $\mathbf{u}_k$ and the $k$th diagonal element of $\mathbf{\Lambda}$ by $\lambda_k$. 





For a Gaussian score model, the probability flow ODE that reverses a VP-SDE forward process is
\begin{equation}
\dot{\mathbf{x}}=-\beta(t)\mathbf{x}-\frac 12g^2(t) (\sigma_{t}^2 \mathbf{I}+\alpha_{t}^2\mathbf{\Sigma})^{-1}(\alpha_{t}\vec{\mu}-\mathbf{x}) \ .    
\end{equation}
Using the decomposition of $\vec{\Sigma}$ described above, 
\begin{align}
\dot{\mathbf{x}}&=-\beta(t)\mathbf{x}-\frac 12g^2(t) \frac{1}{\sigma_t^2} (\mathbf{I}-\mathbf{U}\tilde{\mathbf{\Lambda}}_t \mathbf{U}^T)(\alpha_{t}\vec{\mu}-\mathbf{x}) 
\end{align}
where $\tilde{\mathbf{\Lambda}}_t$ is defined to be the time-dependent diagonal matrix 
\begin{align}
\tilde{\mathbf{\Lambda}}_t &=\text{diag}\left[ \frac{\alpha_t^2\lambda_k}{\alpha_t^2\lambda_k + \sigma_t^2} \right] \ .
\end{align}
Consider the dynamics of the quantity $\mathbf{x}_t-\alpha_t \vec{\mu}$. Using the relationship between $\beta_t$ and $\alpha_t$, we have
\begin{align}
    \frac{d}{dt} ( \mathbf{x}_t-\alpha_t \vec{\mu} ) &=\dot{\mathbf{x}}_t -\vec{\mu} \dot \alpha_t \\
    &=\dot{\mathbf{x}}_t +\beta_t\alpha_t \vec{\mu} \\
    &=\beta_t(\alpha_t\vec{\mu}-\mathbf{x})-\frac 12g^2(t) \frac{1}{\sigma_t^2} (\mathbf{I}-\mathbf{U}\tilde{\mathbf{\Lambda}}_t \mathbf{U}^T)(\alpha_{t}\vec{\mu}-\mathbf{x})\\
    &= \left[ \frac 12g^2(t) \frac{1}{\sigma_t^2} (\mathbf{I}-\mathbf{U}\tilde{\mathbf{\Lambda}}_t \mathbf{U}^T)-\beta_t \mathbf{I} \right](\mathbf{x} - \alpha_{t}\vec{\mu}) \ .
\end{align}
If we assume that the forward process is a variance-preserving SDE, then $\beta_t=\frac12 g^2(t)$, which implies $\alpha_t^2=1-\sigma_t^2$. Using this, we obtain
\begin{align}
    \frac{d}{dt}(\mathbf{x}_t-\alpha_t \vec{\mu}) 
    &=\beta_t \left[ \frac{1}{\sigma_t^2} (\mathbf{I}-\mathbf{U}\tilde{\mathbf{\Lambda}}_t \mathbf{U}^T)-\mathbf{I} \right](\mathbf{x} - \alpha_{t}\vec{\mu})\\
    &=\beta_t \left[ (\frac{1}{\sigma_t^2} - 1)\mathbf{I}-\frac{1}{\sigma_t^2}\mathbf{U}\tilde{\mathbf{\Lambda}}_t \mathbf{U}^T) \right](\mathbf{x} - \alpha_{t}\vec{\mu}) \ .
\end{align}
Define the variable $\mathbf{y}_t:= \mathbf{x}_t-\alpha_t \vec{\mu} $. We have just shown that its dynamics are fairly `nice', in the sense that the above equation is well-behaved separable linear ODE. As we are about to show, it is exactly solvable.


Write $\mathbf{y}_t$ in terms of the orthonormal columns of $\mathbf{U}$ and a component that lies entirely in the orthogonal space $\mathbf{U}^\perp$:
\begin{equation}
    \mathbf{y}_t = \mathbf{y}^{\perp}(t)+\sum_{k=1}^r c_k(t) \mathbf{u}_k \ , \ \mathbf{y}^{\perp}(t) \in \mathbf{U}^\perp \ .
\end{equation}
The dynamics of the coefficient $c_k(t)$ attached to the eigenvector $\mathbf{u}_k$ are 
\begin{align}
    \dot c_k(t)=\frac{d}{dt}(\mathbf{u}_k^T \mathbf{y}_t)&=\beta_t \left[ (\frac{1}{\sigma_t^2} - 1)-\frac{1}{\sigma_t^2} \frac{\alpha_t^2\lambda_k}{\alpha_t^2\lambda_k + \sigma_t^2} \right](\mathbf{u}_k^T \mathbf{y}_t)\\
    &=\frac{\beta_t}{\sigma_t^2} \left( 1-\sigma_t^2-\frac{\alpha_t^2\lambda_k}{\alpha_t^2\lambda_k + \sigma_t^2} \right)c_k(t)\\
    &=\frac{\beta_t\alpha_t^2}{\sigma_t^2}\left( 1-\frac{\lambda_k}{\alpha_t^2\lambda_k + \sigma_t^2} \right) c_k(t) \ .
\end{align}
Using the constraint that $\alpha_t^2+\sigma_t^2=1$, this becomes
\begin{align}
\dot c_k(t)&=\frac{\beta_t\alpha_t^2(1-\lambda_k)}{\alpha_t^2\lambda_k + \sigma_t^2}c_k(t) \ .
\end{align}
For the orthogonal space component $\mathbf{y}^{\perp}(t)$, it will stay in the orthogonal space $\mathbf{U}^{\perp}$, and more specifically the 1D space spanned by the initial $\mathbf{y}^{\perp}(t)$---so, when going backward in time, its dynamics is simply a downscaling of $\mathbf{y}^{\perp}(T)$.
\begin{align}
    \dot{\mathbf{y}}^{\perp}(t)&=\beta_t \left( \frac{1}{\sigma_t^2}-1 \right)\mathbf{y}^{\perp}(t) =\beta_t\frac{1-\sigma_t^2}{\sigma_t^2}\mathbf{y}^{\perp}(t)
    =\frac{\beta_t\alpha_t^2}{\sigma_t^2}\mathbf{y}^{\perp}(t) \ .
\end{align}
Combining these two results and solving the ODEs in the usual way, we have the trajectory solution
\begin{align}
    \mathbf{y}_t &= d(t) \mathbf{y}^{\perp}(T)+\sum_{k=1}^r c_k(t)\mathbf{u}_k\\
    d(t)&=\exp\left(\int_T^t d\tau \frac{\beta_\tau\alpha_\tau^2}{\sigma_\tau^2}\right)\\
    c_k(t)&=c_k(T)\exp\left(\int_T^t d\tau \frac{\beta_\tau\alpha_\tau^2(1-\lambda_k)}{\alpha_\tau^2\lambda_k + \sigma_\tau^2}\right) \ .
\end{align}
The initial conditions are
\begin{align}
    c_k(T)&=\mathbf{u}_k^T \mathbf{y}_T\\
    \mathbf{y}^{\perp}(T)&=\mathbf{y}_T-\sum_{k=1}^r c_k(T)\mathbf{u}_k \ , \ \mathbf{y}^{\perp}(T)\in \mathbf{U}^\perp \ .
\end{align}
To solve the ODEs, it is helpful to use a particular reparameterization of time. In particular, consider a reparameterization in terms of $\alpha_t$ using the relationship $-\beta_t \alpha_t dt=d\alpha_t$. The integral we must do is
\begin{align}
    \int_T^t d\tau \frac{\beta_\tau\alpha_\tau^2(1-\lambda_k)}{\alpha_\tau^2\lambda_k + \sigma_\tau^2}
    &=\int_T^t d\tau \frac{\beta_\tau\alpha_\tau^2(1-\lambda_k)}{1+\alpha_\tau^2(\lambda_k -1)}\\
    &=\int_{\alpha_T}^{\alpha_t} d\alpha_\tau \frac{\alpha_\tau(\lambda_k-1)}{1+\alpha_\tau^2(\lambda_k -1)}\\
    &=\frac 12 \log(1+\alpha_\tau^2(\lambda_k -1))\Bigr|^{\alpha_t}_{\alpha_T}\\
    &=\frac 12 \log\left(\frac{1+(\lambda_k -1)\alpha_t^2}{1+(\lambda_k -1)\alpha_T^2}\right) \ .
\end{align}
Note that taking $\lambda_k=0$ gives us the solution to dynamics in the directions orthogonal to the manifold. We have 
\begin{align}
    c_k(t)&=c_k(T)\sqrt{\frac{1+(\lambda_k-1)\alpha_t^2}{1+(\lambda_k-1)\alpha_T^2}}\\
    d(t)&=\sqrt{\frac{1-\alpha_t^2}{1-\alpha_T^2}} \ .
\end{align}
The time derivatives of these coefficients are
\begin{align}
\dot{c}_k(t)&=c_k(T)\frac{-(\lambda_k-1)\alpha_t^2\beta_t}{\sqrt{(1+(\lambda_k-1)\alpha_T^2)({1+(\lambda_k-1)\alpha_t^2)}}}\\
\dot{d}(t)&=\frac{\alpha_t^2\beta_t}{\sqrt{(1-\alpha_T^2)(1-\alpha_t^2)}} \ .
\end{align}
Finally, we can write out the explicit solution for the trajectory $\mathbf{x}_t$: 
\begin{align}
    \mathbf{x}_t = \alpha_t \vec{\mu} + d(t) \mathbf{y}^{\perp}(T)+\sum_{k=1}^r c_k(t)\mathbf{u}_k \ .
\end{align}
We can see that there are three terms: 1) $\alpha_t\vec{\mu}$, an increasing term that scales up to the mean $\vec{\mu}$ of the distribution; 2) $d(t) \mathbf{y}^{\perp}(T)$, a decaying term downscaling the residual part of the initial noise vector, which is orthogonal to the data manifold; and 3) the $c_k(t)\mathbf{u}_k$ sum, each term of which has independent dynamics. 


We also now have the analytical solution for the projected outcome:
\begin{align}
    \hat{\mathbf{x}}_0(\mathbf{x}_t)-\vec{\mu}
    &=\frac{1}{\alpha_t}\mathbf{U}\tilde{\mathbf{\Lambda}}_t \mathbf{U}^T(\mathbf{x}_t-\alpha_t \vec{\mu})\\
    &=\sum_{k=1}^rc_k(t) \frac{\alpha_t\lambda_k}{\alpha_t^2\lambda_k + \sigma_t^2} \mathbf{u}_k \notag \\
    &=\sum_{k=1}^rc_k(T) \frac{\alpha_t\lambda_k}{\sqrt{(\alpha_t^2\lambda_k + \sigma_t^2)(\alpha_T^2\lambda_k + \sigma_T^2)}} \mathbf{u}_k \ . \notag
\end{align}
Similarly, we can write out the exact formula for the trajectory's tangent vector $\dot{\mathbf{x}}_t$:
\begin{align}
    \dot{\mathbf{x}}_t =& \dot{\alpha_t} \vec{\mu} + \dot{d(t)} \mathbf{y}^{\perp}(T)+\sum_{k=1}^r \dot{c_k}(t)\mathbf{u}_k\\
    =&-\alpha_t\beta_t\vec{\mu} + \frac{\alpha_t^2\beta_t}{\sqrt{(1-\alpha_T^2)(1-\alpha_t^2)}} \mathbf{y}^{\perp}(T) - \sum_{k=1}^r c_k(T)\frac{(\lambda_k-1)\alpha_t^2\beta_t}{\sqrt{(1+(\lambda_k-1)\alpha_T^2)({1+(\lambda_k-1)\alpha_t^2)}}} \mathbf{u}_k \ . \notag
\end{align}

\clearpage

\section{Derivation of endpoint estimate properties} \label{apd:deriv_estimate}

In this section, we study the endpoint estimate (or projected outcome) $\hat{\mathbf{x}}_0$ in the case of a Gaussian score model. 

\subsection{Projected outcome for a Gaussian image distribution}



By definition, the endpoint estimate $\hat{\mathbf{x}}_0(\mathbf{x}_t)$ is
\begin{align}
    \hat{\mathbf{x}}_0(\mathbf{x}_t) &=\frac{\mathbf{x}_t +\sigma_t^2 \mathbf{s}(\mathbf{x}_t,t)}{\alpha_t} =\frac{\mathbf{x}_t +\sigma_t^2 (\sigma_t^2 \mathbf{I}+\alpha_t^2\mathbf{\Sigma})^{-1}(\alpha_t \vec{\mu} -\mathbf{x}_t)}{\alpha_t} \ .
\end{align}
Let's examine the difference between $\hat{\mathbf{x}}_0(\mathbf{x}_t)$ and the mean $\vec{\mu}$. We find that
\begin{align} \label{eq:xhat_supp_formula}
    \hat{\mathbf{x}}_0(\mathbf{x}_t)-\vec{\mu}
    &=\frac{1}{\alpha_t} \left[ \mathbf{x}_t-\alpha_t \vec{\mu} +\sigma_t^2 (\sigma_t^2 \mathbf{I}+\alpha_t^2\mathbf{\Sigma})^{-1}(\alpha_t \vec{\mu} -\mathbf{x}_t) \right] \\
    &=\frac{\sigma_t^2}{\alpha_t} \left[ \frac{1}{\sigma_t^2}\mathbf{I} - (\sigma_t^2 \mathbf{I}+\alpha_t^2\mathbf{\Sigma})^{-1} \right](\mathbf{x}_t-\alpha_t \vec{\mu}) \ . \notag
\end{align}
What can we learn from this formula? We explore several consequences of it below.


\paragraph{The projected outcome is always exact for a delta function.} If the initial distribution $p(\mathbf{x}_0)$ is a delta function, $\mathbf{\Sigma}\to 0$. Then at any time $t$, $p(\mathbf{x}_t)=\mathcal N(\alpha_t \vec{\mu}, \sigma_t^2 \mathbf{I})$, and the projected outcome is
$\hat{\mathbf{x}}_0(\mathbf{x}_t)-\vec{\mu}\equiv0, \forall \mathbf{x}_t,\forall t$. Hence, the projected outcome is always exact, regardless of the position of $\mathbf{x}_t$ and the time $t$. This can be regarded as one justification for this statistic $\hat{\mathbf{x}}_0(\mathbf{x}_t)$: it is exact and invariant in the isotropic score field created by a point distribution. 

This point may be relevant for understanding the very end of reverse diffusion dynamics, when $\mathbf{x}_t$ is likely to live within a score field created by a single point. At such a time, $\hat{\mathbf{x}}_0(\mathbf{x}_t)$ scarcely changes. 


\paragraph{Projected outcome for an isotropic Gaussian.} When $\mathbf{\Sigma}=\bar\sigma^2 \mathbf{I}$, Eq. \ref{eq:xhat_supp_formula} takes a particularly simple form:
\begin{align}
    \hat{\mathbf{x}}_0(\mathbf{x}_t)-\vec{\mu}
    &=\frac{1}{\alpha_t}\frac{\alpha_t^2\bar\sigma^2}{\sigma_t^2+\alpha_t^2\bar\sigma^2} (\mathbf{x}_t-\alpha_t\vec{\mu}) \ .
\end{align}

\subsection{Low rank $\Sigma$} 


Assume that the covariance matrix $\mathbf{\Sigma}$ has rank $r$ somewhat less than $D$, the dimensionality of state space. This case is of particular interest, since images (as previously mentioned) images are often viewed as residing on low-dimensional manifolds. We can use the Woodbury matrix inversion identity to write
\begin{align}
    (\sigma_t^2\mathbf{I}+\alpha_t^2\mathbf{\Sigma})^{-1}&=(\sigma_t^2\mathbf{I}+\alpha_t^2\mathbf{U}\mathbf{\Lambda} \mathbf{U}^T)^{-1}\\
    &=\frac{1}{\sigma_t^2} \mathbf{I} - \frac{1}{\sigma_t^4}\mathbf{U}(\frac{1}{\alpha_t^2}\mathbf{\Lambda}^{-1}+\frac{1}{\sigma_t^2}\mathbf{U}^T\mathbf{U})^{-1}\mathbf{U}^T\\
   &=\frac{1}{\sigma_t^2} \mathbf{I} - \frac{1}{\sigma_t^4}\mathbf{U}(\frac{1}{\alpha_t^2}\mathbf{\Lambda}^{-1}+\frac{1}{\sigma_t^2}\mathbf{I}_r)^{-1}\mathbf{U}^T \ .
\end{align}
We can reuse the previously defined diagonal matrix
\begin{align}
    \tilde{\mathbf{\Lambda}}_t =\text{diag}\left[\frac{\alpha_t^2\lambda_k}{\alpha_t^2\lambda_k + \sigma_t^2} \right] 
\end{align}
to write $(\sigma_t^2\mathbf{I}+\alpha_t^2\mathbf{\Sigma})^{-1}$ as
\begin{align}\label{eq:cov_inv_formula}
    (\sigma_t^2\mathbf{I}+\alpha_t^2\mathbf{\Sigma})^{-1}&=
    \frac{1}{\sigma_t^2} (\mathbf{I}-\mathbf{U}\tilde{\mathbf{\Lambda}}_t \mathbf{U}^T) \ .
\end{align}
Using this result, we can write the endpoint estimate as
\begin{align}
    \hat{\mathbf{x}}_0(\mathbf{x}_t)-\vec{\mu}
    &=\frac{\sigma_t^2}{\alpha_t}(\frac{1}{\sigma_t^2}\mathbf{I} - (\sigma_t^2\mathbf{I}+\alpha_t^2\mathbf{\Sigma})^{-1})(\mathbf{x}_t-\alpha_t \vec{\mu}) =\frac{1}{\alpha_t}\mathbf{U}\tilde{\mathbf{\Lambda}}_t \mathbf{U}^T(\mathbf{x}_t-\alpha_t \vec{\mu}) \ .
\end{align}
This formula has a series of interesting implications. 

\paragraph{Projected outcome stays on image manifold.} Note that the deviation of the endpoint estimate from the distribution mean $\hat{\mathbf{x}}_0(\mathbf{x}_t)-\vec{\mu}$ always remains in the subspace spanned by the columns of $\mathbf{U}$, i.e. $\hat{\mathbf{x}}_0(\mathbf{x}_t)-\vec{\mu}\in \text{span}(\mathbf{U})$. This an interesting result: if the data distribution is a low-dimensional manifold (e.g. image manifold), the estimate $\hat{\mathbf{x}}_0$ will not deviate from this manifold. Even if the projected outcome does not exactly reflect the true outcome, it will not make errors out of the image manifold, i.e. orthogonal to high variance directions. Visually, such directions would correspond to random noise, and hence perturbations in those directions would be `nonsense' perturbations to images.

\paragraph{Projected outcome starts around the center of the distribution.} 
At the start of the reverse diffusion, $t\approx T$, $\alpha_t\approx 0$, and $\sigma_t^2\gg\alpha_t^2$. This means $\tilde{\mathbf{\Lambda}}_t\approx \mathbf{0}$ and $\hat{\mathbf{x}}_0(\mathbf{x}_t)-\vec{\mu} \approx \mathbf{0}$. Hence, the projected outcome $\hat{\mathbf{x}}_0$ initially corresponds to the center of the distribution. For multi-class unconditional generation, this center will be relatively class-ambiguous. For the unconditional generation of faces, we observed that $\hat{\mathbf{x}}_0(\mathbf{x}_T)$ points to a generic face, which is close to the `average face'.

\paragraph{Projected outcome ends in the real outcome.}
At the end of reverse diffusion, $t=0$, $\sigma_t=0,$ and $\alpha_t=1$. This means $\hat{\mathbf{x}}_0(\mathbf{x}_0)=\mathbf{x}_0$, i.e. the projected outcome corresponds to the real result of reverse diffusion.

\paragraph{Features emerge in descending order of their variance.}
Consider the projection of this vector $\hat{\mathbf{x}}_0(\mathbf{x}_t)-\vec{\mu}$ on an eigenvector $\mathbf{u}_k$ of $\mathbf{\Sigma}$:
\begin{align}
    \mathbf{u}_k^T(\hat{\mathbf{x}}_0(\mathbf{x}_t)-\vec{\mu})=\frac{1}{\alpha_t}\frac{\alpha_t^2\lambda_k}{\alpha_t^2\lambda_k + \sigma_t^2}\mathbf{u}_k^T(\mathbf{x}_t-\alpha_t\vec{\mu}) \ .
\end{align}
When $\sigma_t^2 \gg \alpha_t^2\lambda_k$, i.e. when the noise scale is much larger than the signal scale, we have
\begin{equation}
    \mathbf{u}_k^T(\hat{\mathbf{x}}_0(\mathbf{x}_t)-\vec{\mu})\ll \frac{\mathbf{u}_k^T(\mathbf{x}_t-\alpha_t\vec{\mu})}{\alpha_t} \ ,
\end{equation}
so the projected outcome is approximately $\mathbf{0}$. At the other extreme, when $\sigma_t^2\ll \alpha_t^2\lambda_k$, the signal variance is much bigger than the noise variance, and we have
\begin{equation}
    \mathbf{u}_k^T(\hat{\mathbf{x}}_0(\mathbf{x}_t)-\vec{\mu})\approx \frac{\mathbf{u}_k^T(\mathbf{x}_t-\alpha_t\vec{\mu})}{\alpha_t} \ .
\end{equation}
To interpret this, note that if we regard $\mathbf{u}_k$ as a feature direction, then $\lambda_k$ is the variance along this feature direction. The above equation tells us that this feature stays around the mean value of the distribution when the noise variance is much larger than the scaled variance of this feature. In plain language, when the signal scale along a certain dimension is less than the noise scale, the projected outcome along that dimension will remain undetermined.

Empirically, people have found that natural image space has spectra close to $1/f$ \cite{ruderman1994statNatImage}, which means more image variance exists in low-frequency features than in high-frequency features. Thus, we expect that in the generating process, the projected outcome $\hat{\mathbf{x}}_0(\mathbf{x}_t)$ will specify the low-frequency layout first, and then gradually add the high-frequency details. 



\paragraph{Invariance of projected outcome to off-manifold perturbations.} Consider a perturbation of $\mathbf{x}_t$ at time $t$ by $\delta \mathbf{x}$. The projected outcome will change by an amount
\begin{align}
    \hat{\mathbf{x}}_0(\mathbf{x}_t+\delta \mathbf{x})-\hat{\mathbf{x}}_0(\mathbf{x}_t)
    &=\frac{1}{\alpha_t}\mathbf{U}\tilde{\mathbf{\Lambda}}_t \mathbf{U}^T \delta \mathbf{x} \ .
\end{align}
Notice that if $\mathbf{x}_t$ is perturbed in a direction orthogonal to the manifold spanned by the columns of $\mathbf{U}$, the projected outcome $\hat{\mathbf{x}}_0$ will not change. Thus, perturbation in a random non-signal direction will not change the projected outcome of reverse diffusion (and it will also, as we will see, not affect the real result either). In contrast, if the perturbation is aligned with the image manifold, the perturbation \textit{will} affect the projected outcome.

\paragraph{Effect of a perturbation decreases over time.}
As we can see from the formula above, the effect of the same perturbation changes over time. After $\alpha_t^2\lambda_k\gg\sigma_t^2$, the term $\frac{\alpha_t^2\lambda_k}{\alpha_t^2\lambda_k+\sigma_t^2}$ is approximately $1$, so $\frac{1}{\alpha_t}\frac{\alpha_t^2\lambda_k}{\alpha_t^2\lambda_k+\sigma_t^2}$ will decrease over time. We predict on this basis that after certain features emerge, perturbations along those dimensions will have a limited effect.


\clearpage

\section{Derivation of rotational dynamics} \label{apd:deriv_rotation}

In this section, we derive various results quantifying how reverse diffusion trajectories are rotation-like. In particular, under certain assumptions, we will show that the dynamics of the state $\vec{x}_t$ looks like a rotation within a 2D plane spanned by $\vec{x}_0$ (the reverse diffusion endpoint) and $\vec{x}_T$ (the initial noise). We will derive the formula by assuming that the training set consists of a single Gaussian mode, but will explain why this assumption may not be strictly necessary.



\subsection{Derivation of rotation formula and correction terms}

Assume that reverse diffusion begins at time $T$ with $\alpha_T \approx 0$, and ends at time $t = 0$ with $\alpha_0 = 1$. Using our exact solution for $\vec{x}_t$ (Eq. \ref{eq:xt_solu_psi_def}), at some intermediate time $t$ we have that
\begin{equation} \label{eq:x_base}
\vec{x}_t = \alpha_t \vec{\mu} + \sqrt{\frac{1 - \alpha_t^2}{1 - \alpha_T^2}} \vec{y}^{\perp}(T) + \sum_{k=1}^r 
 \sqrt{  \frac{1 + (\lambda_k - 1) \alpha_t^2}{1 + (\lambda_k - 1) \alpha_T^2} }  c_k(T) \vec{u}_k  \ .
\end{equation}
It is also true, by substituting $t = 0$ and $t = T$, that
\begin{equation}
\begin{split}
\vec{\mu} &=  \vec{x}_0 -  \sum_{k=1}^r 
 \sqrt{  \frac{\lambda_k}{1 + (\lambda_k - 1) \alpha_T^2} }  c_k(T) \vec{u}_k  \\
\vec{y}^{\perp}(T)  &= \vec{x}_T - \alpha_T \vec{\mu} - \sum_{k=1}^r   c_k(T) \vec{u}_k  \ .
 \end{split}
\end{equation}
Using these two equations, we can rewrite Eq. \ref{eq:x_base} as
\begin{equation}
\begin{split}
\vec{x}_t &= \alpha_t \vec{\mu} + \sqrt{\frac{1 - \alpha_t^2}{1 - \alpha_T^2}} \left[  \vec{x}_T - \alpha_T \vec{\mu} - \sum_{k=1}^r   c_k(T) \vec{u}_k \right] + \sum_{k=1}^r 
 \sqrt{  \frac{1 + (\lambda_k - 1) \alpha_t^2}{1 + (\lambda_k - 1) \alpha_T^2} }  c_k(T) \vec{u}_k  \\
 &= \left[ \alpha_t - \alpha_T \sqrt{\frac{1 - \alpha_t^2}{1 - \alpha_T^2}} \right] \vec{\mu} + \sqrt{\frac{1 - \alpha_t^2}{1 - \alpha_T^2}} \vec{x}_T + \sum_{k=1}^r 
 \left\{ \sqrt{  \frac{1 + (\lambda_k - 1) \alpha_t^2}{1 + (\lambda_k - 1) \alpha_T^2} } - \sqrt{\frac{1 - \alpha_t^2}{1 - \alpha_T^2}}  \right\}  c_k(T) \vec{u}_k  \\
 &= \left[ \alpha_t - \alpha_T \sqrt{\frac{1 - \alpha_t^2}{1 - \alpha_T^2}} \right] \vec{x}_0 + \sqrt{\frac{1 - \alpha_t^2}{1 - \alpha_T^2}} \vec{x}_T + \vec{R}_t
\end{split}
\end{equation}
where the remainder term $\vec{R}_t$ is equal to
\begin{equation*}
\vec{R}_t =  \sum_{k=1}^r 
 \left\{ \sqrt{  \frac{1 + (\lambda_k - 1) \alpha_t^2}{1 + (\lambda_k - 1) \alpha_T^2} } - \sqrt{\frac{1 - \alpha_t^2}{1 - \alpha_T^2}} - \left[ \alpha_t - \alpha_T \sqrt{\frac{1 - \alpha_t^2}{1 - \alpha_T^2}} \right] \sqrt{  \frac{\lambda_k}{1 + (\lambda_k - 1) \alpha_T^2} } \right\}  c_k(T) \vec{u}_k  \ .
\end{equation*}
The expression simplifies somewhat if we take $\alpha_T \approx 0$. Doing so, we obtain the equation seen in the main text:
\begin{equation*} 
\begin{split}
\vec{x}_t \approx \ & \alpha_t \vec{x}_0 + \sqrt{1 - \alpha_t^2} \ \vec{x}_T + \sum_{k=1}^r \left\{ \sqrt{\sigma_t^2 + \lambda_k \alpha_t^2} - \alpha_t \sqrt{\lambda_k} - \sigma_t \right\} c_k(T) \vec{u}_k \ .
\end{split}
\end{equation*}
Let's examine the correction terms more closely. Define the function
\begin{equation}
\begin{split}
J(\alpha_t; \lambda) &:= \sqrt{\sigma_t^2 + \lambda \alpha_t^2} - \alpha_t \sqrt{\lambda} - \sigma_t \\
&= \sqrt{1 + (\lambda - 1) \alpha_t^2} - \alpha_t \sqrt{\lambda} - \sqrt{1 - \alpha_t^2} \ .
\end{split}
\end{equation}
Note that we can rewrite $J$ as
\begin{equation}
\begin{split}
J(\alpha_t; \lambda) &=  \frac{\left( \sqrt{\sigma_t^2 + \lambda \alpha_t^2} - \alpha_t \sqrt{\lambda} - \sigma_t \right) \left( \sqrt{\sigma_t^2 + \lambda \alpha_t^2} + \alpha_t \sqrt{\lambda} + \sigma_t \right)}{ \sqrt{\sigma_t^2 + \lambda_k \alpha_t^2} + \alpha_t \sqrt{\lambda} + \sigma_t } \\
&= - 2 \frac{ \sigma_t \alpha_t \sqrt{\lambda}}{ \sqrt{\sigma_t^2 + \lambda \alpha_t^2} + \alpha_t \sqrt{\lambda} + \sigma_t}  \ .
\end{split}
\end{equation}
From this, it is immediately clear that the correction term is not completely arbitrary. First, $J$ is always negative. Second, its time course is bowl-shaped: it begins close to zero (since $\alpha_T \approx 0$), becomes more negative, then ends close to zero (since $\sigma_0 \approx 0$). It achieves its most negative value roughly when $\sigma_t$ and $\alpha_t \sqrt{\lambda}$ are comparable, i.e. when
\begin{equation}
\alpha_t \approx \sqrt{\frac{1}{\lambda + 1}} \ ,
\end{equation}
in which case
\begin{equation}
J(\alpha_t; \lambda) \approx - 2 \left( 1 - \frac{\sqrt{2}}{2} \right) \sqrt{\frac{\lambda}{\lambda + 1}} \geq - 2 \left( 1 - \frac{\sqrt{2}}{2} \right) \approx - 0.586 \ .
\end{equation}
Notice that $|J| < 1$, regardless of $\lambda$. 

Although we derived this formula by assuming a single Gaussian mode, its form does not actually depend on any properties of the mode. This suggests that, as long as the score function landscape looks \textit{locally} Gaussian, the formula may still be applicable. For example, suppose the learned image distribution is a Gaussian mixture. Even though the mean and the covariance of the nearest mode---the one which we expect to dominate the score function---may regularly change throughout reverse diffusion, even in a discontinuous way, the rotation equation should stay the same.

\subsection{Low-rank image distribution sufficient for small correction terms}

Suppose that the rank $r$ of the covariance matrix $\vec{\Sigma}$ is much less than $D$, the dimensionality of state space. The error in the rotation formula is
\begin{equation}
\left\Vert \vec{x}_t - \alpha_t \vec{x}_0 - \sqrt{1 - \alpha_t^2} \ \vec{x}_T \right\Vert^2_2 = \sum_{k=1}^r J(\alpha_t; \lambda_k)^2 c_k(T)^2  \ .
\end{equation}
Recall that $c_k(T)$ is the coefficient of the original noise seed $\vec{x}_T \sim \mathcal{N}(\vec{0}, \vec{I})$ along the direction $\vec{u}_k$. Assuming $D$ is large, the norm of the noise seed is approximately $1$. Since there is a priori no relationship between $\vec{x}_T$ and $\vec{u}_k$, we expect that $\vec{x}_T \cdot \vec{u}_k \approx 1/\sqrt{D}$. (Suppose we express $\vec{x}_T$ in terms of a set of $D$ orthonormal basis vectors. Given that its norm is $1$, and that it has no special relationship with any basis vector, the overlap between $\vec{x}_T$ and each vector must be about $1/\sqrt{D}$.) The error becomes
\begin{equation*}
\left\Vert \vec{x}_t - \alpha_t \vec{x}_0 - \sqrt{1 - \alpha_t^2} \ \vec{x}_T \right\Vert^2_2 \approx \sum_{k=1}^r J(\alpha_t; \lambda_k)^2 \frac{1}{D} \leq  \sum_{k=1}^r 4 \left( 1 - \frac{\sqrt{2}}{2}  \right)^2 \frac{1}{D}  = 4 \left( 1 - \frac{\sqrt{2}}{2}  \right)^2 \frac{r}{D}
\end{equation*}
where we have used the bound from the previous subsection. Since $r \ll D$, this error is small.

It is worth noting, however, that the $r \ll D$ assumption is not \textit{necessary} for the rotation formula correction terms to be negligible. Another case in which this is true is when the image distribution is isotropic, i.e. $\lambda_k = \lambda$ for all $k$. Then the error is
\begin{equation*}
\left\Vert \vec{x}_t - \alpha_t \vec{x}_0 - \sqrt{1 - \alpha_t^2} \ \vec{x}_T \right\Vert^2_2 \approx  4 \left( 1 - \frac{\sqrt{2}}{2}  \right)^2 \frac{\lambda}{\lambda + 1} \frac{r}{D} < \frac{\lambda}{\lambda + 1} \ .
\end{equation*}
This is somewhat smaller than the typical scale of $\vec{x}_t$, since $\Vert \vec{x}_t \Vert_2^2$ remains roughly between $1$ and $\lambda$.


\subsection{Can the rotation formula be used to predict the trajectory endpoint?}

Naively, since any two vectors are mathematically sufficient to define a plane, the rotation plane should be completely determined from the first two steps---or if not the first two steps, one might naively expect the first \textit{several} steps to be sufficient. In particular, since
\begin{equation}
\vec{x}_t \approx \alpha_t \vec{x}_0 + \sqrt{1 - \alpha_t^2} \vec{x}_T + \vec{R}_t \ ,
\end{equation}
where $\vec{R}_t$ is the correction term we derived earlier, we can approximate $\vec{x}_0$ as
\begin{equation}
\vec{x}_0 \approx  \frac{\vec{x}_t - \sqrt{1 - \alpha_t^2} \ \vec{x}_T}{\alpha_t} \ .
\end{equation}
The problem is that the correction term along each feature direction is `large' until that feature has been `committed to'. Concretely, the correction term along direction $\vec{u}_k$ is proportional to
\begin{equation}
\frac{J(\alpha_t; \lambda_k)}{\alpha_t} =  - 2 \frac{ \sigma_t \sqrt{\lambda}}{ \sqrt{\sigma_t^2 + \lambda \alpha_t^2} + \alpha_t \sqrt{\lambda} + \sigma_t} \ .
\end{equation}
This function has saturating behavior, and remains high (in the sense of being $\sim \sqrt{\lambda_k}$) until around the time when $\alpha_t \sqrt{\lambda_k} \approx \sqrt{1 - \alpha_t^2}$. But consistent with our other results (see Figure 3 and the associated formulas), this is roughly around the time of `feature commitment', or more specifically when the sigmoid-shaped coefficients of $\hat{\vec{x}}_0$ begin to transition to their final value. So the rotation formula only becomes useful for determining the endpoint after enough time has passed that one is sufficiently close to the endpoint.

Using multiple $\vec{x}_t$ does not help reduce the error in applying the rotation formula, since the error equation above is monotonic in time. In other words, averaging over multiple recent $\vec{x}_t$ is \textit{strictly worse} than just using the rotation formula and the most recent (i.e. greatest number of reverse diffusion time steps) $\vec{x}_t$.

As a final note: since we have the form of the correction terms, why not use that as additional information? We \textit{could} do this, but this only works for the Gaussian case, where we already have access to the full solution! And knowing these terms at all times is also roughly equivalent to knowing the full trajectory. So in summary, viewing reverse diffusion trajectories as 2D `rotations' is a useful geometric picture, but it is less quantitatively useful than the full analytical solution to the Gaussian model, e.g. for accelerating sampling.

\subsection{Rotation-like dynamics beyond the Gaussian model}

There is an alternative source of evidence that reverse diffusion dynamics are rotation-like, even in a more general non-Gaussian setting. Using the form of the projected outcome (Eq. \ref{eq:xhat_supp_formula}), we can write the probability flow ODE (Eq. \ref{eq:rev_flow}) as
\begin{align}
    \dot{\mathbf{x}}&=-\beta(t)\mathbf{x}-\frac 12g^2(t) \nabla_{\mathbf{x}} \log p(\mathbf{x},t) =-\beta(t)\mathbf{x}-\frac 12\frac{g^2(t)}{\sigma_t^2} \left[ \alpha_t \hat{\mathbf{x}}_0(\mathbf{x})- \mathbf{x} \right] \ .
\end{align}
Notice that 
\begin{align}
    \alpha(t)=\exp\bigg(-\int_0^t\beta(\tau)d\tau\bigg) \hspace{0.3in} \frac{d}{dt}\alpha(t)=-\beta(t)\alpha(t) \ ,
\end{align}
which allows us to write
\begin{equation}
\begin{split}
    \frac{d}{dt}\left( \frac{\mathbf{x}_t}{\alpha_t} \right)&=\frac{\dot{\mathbf{x}}_t}{\alpha_t}-\frac{\dot{\alpha}_t \mathbf{x}_t}{\alpha_t^2} =\frac{\dot{\mathbf{x}}_t}{\alpha_t}+\frac{\beta_t \mathbf{x}_t}{\alpha_t} \ .
\end{split}
\end{equation}
Equivalently,
\begin{align} \label{eq:attract_ODE}
    \frac{d}{dt} \left( \frac{\mathbf{x}_t}{\alpha_t} \right) &=-\frac 12\frac{g^2(t)}{\sigma_t^2}\left( \hat{\mathbf{x}}_0(\mathbf{x}_t)- \frac{\mathbf{x}_t}{\alpha_t} \right) = -\frac{\beta_t}{\sigma_t^2}\left( \hat{\mathbf{x}}_0(\mathbf{x}_t)- \frac{\mathbf{x}_t}{\alpha_t} \right)  \ .
\end{align}
From this, we can see that the quantity $\mathbf{x}_t/\alpha_t$, i.e. the state scaled by the signal scale, is isotropically attracted towards the moving target $\hat{\mathbf{x}}_0(\mathbf{x}_t)$ at a rate determined by $\frac 12\frac{g^2(t)}{\sigma_t^2}$. 

Suppose that the endpoint estimates $\hat{\mathbf{x}}_0$ change slowly compared to the state $\mathbf{x}_t$; we will show that this gives us rotation-like dynamics. 

First, note that we can evaluate the integral
\begin{equation}
\begin{split}
\int_{T}^t \frac{\beta_t}{\sigma_t^2} \ dt = \int_T^t \frac{\beta_t}{1 - \alpha_t^2} \ dt
\end{split}
\end{equation}
using a change of variables $\alpha := \alpha_t$ with $d\alpha = - \beta \alpha \ dt$. The integral becomes
\begin{equation}
\begin{split}
- \int_{\alpha_T}^{\alpha_t} \frac{d \alpha}{\alpha (1 - \alpha^2)} = \left. \log \frac{\sqrt{1 - \alpha ^2}}{\alpha }  \right|_{\alpha_T}^{\alpha_t} \ .
\end{split}
\end{equation}
Using this integral, we can find that the solution to Eq. \ref{eq:attract_ODE} under the assumption that $\hat{\mathbf{x}}_0$ remains constant is
\begin{equation} \label{eq:slow_ODE_sln}
\begin{split}
\log\left( \frac{\frac{\mathbf{x}_t}{\alpha_t} - \hat{\mathbf{x}}_0}{\frac{\mathbf{x}_T}{\alpha_T} - \hat{\mathbf{x}}_0} \right) &= \log\left( \frac{\sqrt{1 - \alpha_t^2}}{\alpha_t} \frac{\alpha_T}{\sqrt{1 - \alpha_T^2}} \right) \\
\implies \ \frac{\mathbf{x}_t - \alpha_t \hat{\mathbf{x}}_0}{\sqrt{1 - \alpha_t^2}} &= \frac{\mathbf{x}_T - \alpha_T \hat{\mathbf{x}}_0}{\sqrt{1 - \alpha_T^2}} \ .
\end{split}
\end{equation}
Interestingly, this indicates that there is a conserved quantity
\begin{equation}
\frac{\mathbf{x}_t - \alpha_t \hat{\mathbf{x}}_0}{\sqrt{1 - \alpha_t^2}} = \text{const.}
\end{equation}
along the reverse diffusion trajectory, under this approximation. Since $\alpha_T \approx 0$,  $\mathbf{x}_T \approx \text{const}.$, i.e. the value of the constant roughly matches the initial noise seed $\mathbf{x}_T$. Given any $\hat{\mathbf{x}}_0$ the solution to the ODE at time $t$ can be written as 
\begin{equation}\label{eq:fictative_traj}
    \mathbf{x}_t = \alpha_t \hat{\mathbf{x}}_0 + \sqrt{1 - \alpha_t^2} \ \text{const.} \approx \alpha_t \hat{\mathbf{x}}_0 + \sqrt{1 - \alpha_t^2} \ \mathbf{x}_T \ .
\end{equation}
In words: through a rotation, $\mathbf{x}_t$ interpolates between $\hat{ \mathbf{x}}_0$ and $\text{const}$. This solution paints the picture that the state is constantly rotating towards the estimated outcome $\hat{\mathbf{x}}_0$, with Eq. \ref{eq:fictative_traj} describing that hypothetical trajectory's shape. But as the target $\hat{\mathbf{x}}_0$ is moving, the actual trajectory will be similar to Eq. \ref{eq:fictative_traj} only on short time scales, and not on longer time scales. This idea is visualized by the circular dashed curves in Fig. \ref{fig:conceptual_pics}. 

Beyond the constant $\hat{\mathbf{x}}_0$ approximation, there will be some correction terms to the above rotation formula, whose precise form depends on the (generally not Gaussian) score function.

\clearpage


\section{Derivation of properties of non-Gaussian score models} \label{apd:deriv_nongaussian} \label{apd:gmm_deriv}
In this appendix, we will derive some properties of non-Gaussian score models. In particular, we will derive the score function and endpoint estimate for a Gaussian mixture model, and show that both have a simple form and intuitive interpretation.

\subsection{Score function of general Gaussian mixture}
Let 
\begin{equation}
q(\mathbf{x}) = \sum_i \pi_i \mathcal N(\mathbf{x};\vec{\mu}_i,\vec{\Sigma}_i)
\end{equation}
be a Gaussian mixture distribution, where the $\pi_i$ are mixture weights, $\vec{\mu}_i$ is the $i$-th mean, and $\vec{\Sigma}_i$ is the $i$-th covariance matrix. The score function for this distribution is
\begin{equation}
\begin{split}
  \nabla_{\mathbf{x}} \log q(\mathbf{x}) = &\frac{\sum_i \pi_i \nabla_{\mathbf{x}} \mathcal N(\mathbf{x};\vec{\mu}_i,\vec{\Sigma}_i)}{q(\mathbf{x})}\\
    =&\sum_i -\Sigma_i^{-1}(\mathbf{x}-\vec{\mu}_i)\frac{\pi_i \mathcal N(\mathbf{x};\vec{\mu}_i,\vec{\Sigma}_i)}{q(\mathbf{x})}\\
    =&\sum_i \frac{\pi_i \mathcal N(\mathbf{x};\vec{\mu}_i,\vec{\Sigma}_i)}{q(\mathbf{x})}\nabla_\mathbf{x} \log \mathcal N(\mathbf{x};\vec{\mu}_i,\vec{\Sigma}_i) \\
    =&\sum_i w_i(\mathbf{x})\nabla_\mathbf{x} \log \mathcal N(\mathbf{x};\vec{\mu}_i,\vec{\Sigma}_i) 
\end{split}
\end{equation}
where we have defined the mixing weights
\begin{align}
    w_i(\mathbf{x}):=\frac{\pi_i \mathcal N(\mathbf{x};\vec{\mu}_i,\vec{\Sigma}_i)}{q(\mathbf{x})}  \ .
\end{align}
Thus, the score of the Gaussian mixture is a weighted mixture of the score fields of each of the individual Gaussians. 

In the context of diffusion, we are interested in the \textit{time-dependent} score function. Given a Gaussian mixture initial condition, the end result of the VP-SDE forward process will also be a Gaussian mixture:
\begin{align}
    p_t(\mathbf{x}) = \sum_i \pi_i \mathcal N(\mathbf{x};\alpha_t \vec{\mu}_i,\sigma_t^2 \vec{I}+\alpha_t^2 \vec{\Sigma}_i) \ .
\end{align}
The corresponding time-dependent score is
\begin{equation}
\begin{split}
\mathbf{s}(\mathbf{x},t)&=\nabla_{\mathbf{x}} \log p_t(\mathbf{x})\\
=&\sum_i -(\sigma_t^2 \vec{I}+\alpha_t^2 \vec{\Sigma}_i)^{-1}(\mathbf{x}-\alpha_t \vec{\mu}_i)\frac{\pi_i \mathcal N(\mathbf{x};\alpha_t \vec{\mu}_i,\sigma_t^2 \vec{I}+\alpha_t^2 \vec{\Sigma}_i)}{p_t(\mathbf{x})}\\
=&\sum_i -(\sigma_t^2 \vec{I}+\alpha_t^2 \vec{\Sigma}_i)^{-1}(\mathbf{x}-\alpha_t \vec{\mu}_i)w_i(\mathbf{x}, t) \ .
\end{split}
\end{equation}
Note that we have a formula for $(\sigma_t^2 \vec{I}+\alpha_t^2 \vec{\Sigma}_i)^{-1}$ (derived using the Woodbury matrix inversion identity; see Eq.\ref{eq:cov_inv_formula}) in terms of the (compact) SVD of $\vec{\Sigma}_i$. We can use it to write
\begin{align}\label{eq:efficient_gmm_score}
\mathbf{s}(\mathbf{x},t)=\frac{1}{\sigma_t^2} \sum_i -(\mathbf{I}-\mathbf{U}_i\tilde{\mathbf{\Lambda}_i}_t \mathbf{U}_i^T)(\mathbf{x}-\alpha_t \vec{\mu}_i) w_i(\mathbf{x}, t)
\end{align}
where
\begin{equation}
\vec{\Sigma}_i=\mathbf{U}_i\mathbf{\Lambda}_i\mathbf{U}_i^T \hspace{1in} \tilde{\mathbf{\Lambda}}_t :=\text{diag}\left[\frac{\alpha_t^2\lambda_k}{\alpha_t^2\lambda_k + \sigma_t^2} \right] \ .
\end{equation}
This representation of the score function is numerically convenient, since (once the SVDs of each covariance matrix have been obtained), it can be evaluated using a relatively small number of matrix multiplications, which are cheaper than the covariance matrix inversions that a naive implementation of the Gaussian mixture score function would require.


We used this formula (Eq.\ref{eq:efficient_gmm_score}) and an off-the-shelf ODE solver to simulate the reverse diffusion trajectory of a 10-mode Gaussian mixture score model (Fig. \ref{fig:CIFAR_theory_valid}). 

\subsection{Score function of Gaussian mixture with identical and isotropic covariance}
Assume each Gaussian mode has covariance $\vec{\Sigma}_i=\sigma^2 \vec{I}$ and that every mixture weight is the same (i.e. $\pi_i=\pi_j,\forall i,j$). Then the score for this kind of specific Gaussian mixture is 
\begin{equation}\label{eq:gmm_isotrop}
\begin{split}
\nabla_{\mathbf{x}} \log q(\mathbf{x}) = &\frac{\sum_i \pi_i \nabla_{\mathbf{x}} \mathcal N(\mathbf{x};\vec{\mu}_i,\sigma^2 \vec{I})}{\sum_i \pi_i \mathcal N(\mathbf{x};\vec{\mu}_i,\sigma^2 \vec{I})}\\
    =&\frac{\sum_i -\frac{1}{\sigma^2}(\mathbf x-\vec{\mu}_i) \exp \big(-\frac{1}{2\sigma^2}\|\mathbf x - \vec{\mu}_i\|_2^2\big)}{\sum_i\exp \big(-\frac{1}{2\sigma^2}\|\mathbf x - \vec{\mu}_i\|_2^2\big)}\\
    =&\frac{1}{\sigma^2}\sum_i w_i(\mathbf{x})(\vec{\mu}_i - \mathbf x) \ ,
\end{split}
\end{equation}
where the weight $w_i(\mathbf{x})$ is a softmax of the negative squared distance to all the means, with $\sigma^2$ functioning as a temperature parameter:
\begin{align}\label{eq:gmm_w_softmax}
    w_i(\mathbf{x}) &= Softmax(\big\{-\frac{1}{2\sigma^2}\|\mathbf x - \vec{\mu}_i\|_2^2\big\}) =\frac{\exp \big(-\frac{1}{2\sigma^2}\|\mathbf x - \vec{\mu}_i\|_2^2\big)}{\sum_i\exp \big(-\frac{1}{2\sigma^2}\|\mathbf x - \vec{\mu}_i\|_2^2\big)} \ .
\end{align}
Since the weights $w_i(\mathbf{x})$ sum to $1$, we can also write the score function in the suggestive form
\begin{align}
\nabla_{\mathbf{x}} \log q(\mathbf{x}) = \frac{(\sum_i w_i(\mathbf{x}) \vec{\mu}_i )-\mathbf{x}}{\sigma^2} \ .
\end{align}
This has a form analogous to the score of a single Gaussian mode---but instead of $\mathbf{x}$ being `attracted' towards a single mean $\vec{\mu}$, it is attracted towards a weighted combination of all of the means, with modes closer to the state $\mathbf{x}$ being more highly weighted.

The time-dependent score function of this model is
\begin{equation}\label{eq:gmm_isotrop_rewrite}
\nabla_{\mathbf{x}} \log q(\mathbf{x}) = \frac{(\sum_i w_i(\mathbf{x}, t) \alpha_t \vec{\mu}_i )-\mathbf{x}}{\sigma_t^2 + \alpha_t^2 \sigma^2} 
\end{equation}
where
\begin{align}
    w_i(\mathbf{x}, t) &= Softmax(\big\{-\frac{1}{2 ( \sigma_t^2 + \alpha_t^2 \sigma^2 ) }\|\mathbf x - \alpha_t \vec{\mu}_i\|_2^2\big\})  \ .
\end{align}

\subsection{Endpoint estimate of Gaussian mixture with identical and isotropic covariance}

The endpoint estimate of the Gaussian mixture whose modes have identical isotropic covariances is
\begin{equation}
\begin{split}
    \hat{\mathbf x}_0(\mathbf x_t)&=\frac{\mathbf{x}_t+\sigma_t^2 \nabla \log p(\mathbf x_t)}{\alpha_t} = \frac{1}{\alpha_t} \left[ \frac{\alpha_t^2 \sigma^2}{\sigma_t^2 + \alpha_t^2 \sigma^2} \mathbf{x}_t + \frac{1}{\sigma_t^2 + \alpha_t^2 \sigma^2} \sum_i w_i(\mathbf{x}_t, t) \alpha_t \vec{\mu}_i \right] \ .
\end{split}
\end{equation}

\subsection{Score and endpoint estimate of exact (delta mixture) score model}

A particularly interesting special case of the Gaussian mixture model is the delta mixture model used in the main text, whose components are vanishing width Gaussians centered on the training images. In particular, consider a data set $\{\mathbf{y}_i\}$ with $i=1,...,N$, so that the starting distribution is 
\begin{equation}
p(\mathbf{x})=\frac{1}{N} \sum_i\delta (\mathbf{x}-\mathbf{y}_i) \ .
\end{equation} 
At time $t$, the marginal distribution will be a Gaussian mixture 
\begin{align}
    p_t(\mathbf{x}_t)=\frac{1}{N} \sum_i \mathcal{N}(\mathbf{x}_t;\alpha_t\mathbf{y}_i,\sigma_t^2 \mathbf{I}) \ .
\end{align}
Then using the Eq. \ref{eq:gmm_isotrop} above we have 
\begin{equation}
\begin{split}
    s(\mathbf x_t,t)=\nabla \log p_t(\mathbf x_t) &= \frac{1}{\sigma_t^2} \sum_i w_i(\mathbf x_t)(\alpha_t\mathbf{y}_i - \mathbf x_t)\\
    &= \frac{1}{\sigma_t^2} \left[- \mathbf x_t + \alpha_t \sum_i w_i(\mathbf x_t)\mathbf{y}_i \right]\\
    &= \frac{1}{\sigma_t^2} \left[- \mathbf x_t + \alpha_t \sum_i Softmax(\big\{-\frac{1}{2\sigma_t^2}\|\alpha_t\mathbf{y}_i - \mathbf x_t\|^2\big\}) \mathbf{y}_i \right] \ .
\end{split}
\end{equation}
The endpoint estimate of the distribution is 
\begin{equation}
\begin{split}
    \hat{\mathbf x}_0(\mathbf x_t)&=\frac{\mathbf{x}_t+\sigma_t^2 \nabla \log p(\mathbf x_t)}{\alpha_t} \\
    &=\sum_i Softmax(\big\{-\frac{1}{2\sigma_t^2}\|\alpha_t\mathbf{y}_i - \mathbf x_t\|^2\big\}) \mathbf{y}_i\\
    &=\sum_i w_i(\mathbf x_t)\mathbf{y}_i \ .
\end{split}
\end{equation}
Thus, the endpoint estimate is a weighted average of training data, with the softmax of negative squared distance as weights and $\sigma_t^2$ as a temperature parameter. 

\end{document}